%% file: april_aigc.tex
\PassOptionsToPackage{table}{xcolor}
\PassOptionsToPackage{pagebackref,breaklinks=true,colorlinks,citecolor=blue,urlcolor=blue,linkcolor=blue,bookmarks=false}{hyperref}
\PassOptionsToPackage{noabbrev,nameinlink,capitalize}{cleveref}

\documentclass[onecolumn,numbers]{april_aigc}


\usepackage{xurl}
\usepackage{amsfonts}
\usepackage{amssymb}
\usepackage{amsmath}
\usepackage{nicefrac}
\usepackage{microtype}
\usepackage{xspace}
\usepackage{fix-cm}

\usepackage{booktabs}
\usepackage{wrapfig}
\usepackage{multicol}
\usepackage{multirow}
\usepackage{makecell}
\usepackage{tabularx}
\usepackage{longtable}
\usepackage{adjustbox}
\usepackage{array}
\usepackage{colortbl}
\usepackage{pifont}
\usepackage{enumitem}
\usepackage[normalem]{ulem}
\usepackage[most]{tcolorbox}

\usepackage{xcolor}
\definecolor{linkcolor}{named}{aprilblue}
\definecolor{urlcolor}{RGB}{255,105,180}
\definecolor{citecolor}{RGB}{66,168,235}
\definecolor{lightgray}{rgb}{0.8, 0.8, 0.8}
\definecolor{darkgreen}{rgb}{0.00, 0.81, 0.78}

\definecolor{gray_tab}{RGB}{220, 220, 220}
\definecolor{blue_tab}{RGB}{227, 240, 251}
\definecolor{oran_tab}{RGB}{252, 242, 237}
\definecolor{whit_tab}{RGB}{255, 255, 255}
\definecolor{green_code}{RGB}{55, 126, 34}
\definecolor{codeblue}{rgb}{0.25,0.5,0.5}
\definecolor{codekw}{rgb}{0.85,0.18,0.50}
\definecolor{sgink}{HTML}{1C2B33}
\definecolor{sgline}{HTML}{5D6B78}
\definecolor{sgpanel}{HTML}{F5F7FA}
\definecolor{sgpanelblue}{HTML}{EAF2FF}
\definecolor{sgpanelgreen}{HTML}{ECF8F2}
\definecolor{sgpanelorange}{HTML}{FFF3E6}
\definecolor{sgblue}{HTML}{0064E0}
\definecolor{sgcyan}{HTML}{0088A9}
\definecolor{sggreen}{HTML}{168A54}
\definecolor{sgorange}{HTML}{D66A00}
\definecolor{sgred}{HTML}{B8322A}
\definecolor{sgpurple}{HTML}{6B5FB5}

\newcommand{\tb}{\textcolor[RGB]{192, 0, 0}}

\newcolumntype{L}[1]{>{\raggedright\arraybackslash}p{#1}}
\newcolumntype{Y}{>{\raggedright\arraybackslash}X}

\usepackage{algorithm}
\usepackage{algorithmic}
\usepackage{listings}
\usepackage{etoolbox}
\usepackage{tikz}
\usetikzlibrary{arrows.meta,positioning,fit,calc}

\makeatletter
\AfterEndEnvironment{algorithm}{\let\@algcomment\relax}
\AtEndEnvironment{algorithm}{\kern2pt\hrule\relax\vskip3pt\@algcomment}
\let\@algcomment\relax
\newcommand\algcomment[1]{\def\@algcomment{\footnotesize#1}}
\renewcommand\fs@ruled{\def\@fs@cfont{\bfseries}\let\@fs@capt\floatc@ruled
  \def\@fs@pre{\hrule height.8pt depth0pt \kern2pt}%
  \def\@fs@post{}%
  \def\@fs@mid{\kern2pt\hrule\kern2pt}%
  \let\@fs@iftopcapt\iftrue}
\makeatother

\def\onedot{.\xspace}
\def\eg{\textit{e.g}\onedot}
\def\Eg{\textit{E.g}\onedot}

\def\etc{\textit{etc}\onedot}

\setlist[itemize]{leftmargin=1.4em,itemsep=0.15em,topsep=0.3em}
\setlist[enumerate]{leftmargin=1.6em,itemsep=0.15em,topsep=0.3em}
\AtEndPreamble{
    \crefname{section}{Sec.}{Secs.}
    \Crefname{section}{Section}{Sections}
    \crefname{table}{Tab.}{Tabs.}
    \Crefname{table}{Table}{Tables}
    \crefname{equation}{Eq.}{Eqs.}
    \Crefname{equation}{Equation}{Equations}
    \crefname{figure}{Fig.}{Figs.}
    \Crefname{figure}{Figure}{Figures}
}
\hypersetup{colorlinks=true,linkcolor=linkcolor,urlcolor=urlcolor,citecolor=citecolor}

\DeclareCaptionFormat{custom}{{\color{aprilblue}\sffamily\textbf{#1 #2}} #3}
\titleformat*{\section}{\color{aprilblue}\Large\sffamily\bfseries}
\titleformat*{\subsection}{\color{aprilblue}\large\sffamily\bfseries}
\titleformat*{\subsubsection}{\color{aprilblue}\normalsize\sffamily\bfseries}

\usepackage{fancyhdr}
\newif\ifshowlogo
\showlogotrue
\newcommand{\insertlogo}{%
  \ifshowlogo
    \IfFileExists{assets/april_logo1.png}%
    {\includegraphics[height=0.68cm]{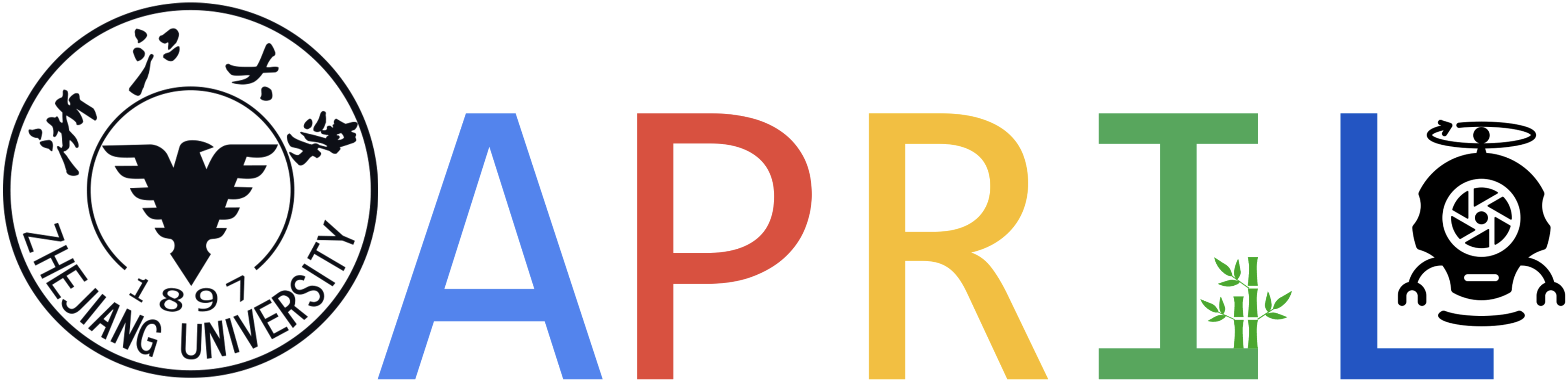}}%
    {}%
  \fi
}
\newif\ifshowtoc
\showtocfalse

\renewcommand{\title}[1]{\def\titlelist{{\fontsize{20pt}{28pt}\selectfont\sffamily\bfseries #1}}}
\title{From Seeing to Acting: Smart Glasses as First-Person Intelligence Platforms}

\author[1,\star\raisebox{-0.2em}{\includegraphics[height=0.85em]{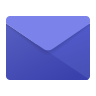}}]{Jiangning Zhang}
\author[1,\star]{Haojun Chen}
\author[1]{Yong Liu}
\affiliation[1]{Zhejiang University, APRIL Lab}

\abstract{
Smart glasses are evolving from capture and display accessories into first-person intelligence platforms that connect human perception, persistent context, and digital or physical action. Their on-body viewpoint aligns with the wearer's vision, audition, motion, and hand-object interaction, yet must operate under tight energy, thermal, privacy, and feedback constraints. Despite rapid advances in augmented reality, egocentric vision, multimodal models, human-computer interaction, and embodied intelligence, the literature remains fragmented across isolated devices, tasks, and benchmarks. 
\textit{The central challenge is not whether a model can recognize, answer, remember, or act in isolation, but whether a complete system can sustain a reliable, temporally valid, correctable, and governable perception-state-interaction-action loop. }
This survey is \textit{the \textbf{first} to systematically study smart glasses and develops a unified framework for investigating this loop}. We formalize smart glasses through first-person data flow and constrained task utility, consolidate devices into eight verifiable hardware capability axes, organize the literature around seven interdependent foundational capabilities, and introduce an L0-L5 framework spanning capture, reactive perception, contextual assistance, persistent state, governed action, and embodied coupling. Across nine application scenes, we connect tasks to datasets, systems, products, stakeholders, failure consequences, and evidence gaps. We further present a nine-dimensional deployment framework, a claim-conditioned evaluation protocol, and an evidence ladder from controlled measurement to longitudinal field validation and audit. 
The resulting framework turns smart glasses into comparable, deployable, and reproducibly evaluated research objects, and outlines a roadmap toward trustworthy first-person intelligence, providing guidance for future research and applications in this rapidly evolving field. 
}

\coverdate{August 2026}
\covercorrespondence{\email{186368@zju.edu.cn}}
\coverproject{https://github.com/zhangzjn/awesome-smart-glasses}
\metadata[Keywords]{Smart Glasses, First-Person Intelligence, Egocentric Perception, Ego-View Interaction, Multimodal Interaction, Physical AI, Embodied Intelligence, Wearable Agents}

\begin{document}

\maketitle
\thispagestyle{plain}

\ifshowtoc
    \clearpage
    \setcounter{tocdepth}{2}
    \tableofcontents
    \vspace{1cm}
    \clearpage
\fi

\input{sec/01_introduction}

\input{sec/02_background}
\input{sec/03_capabilities}
\input{sec/04_applications}

\input{sec/05_design_framework}
\input{sec/06_conclusion}


\bibliography{april_aigc}

\end{document}

%% file: sec/01_introduction.tex
\providecommand{\sgpoint}[2]{\par\noindent\textbf{#1.} #2\par\vspace{0.18em}}
\providecommand{\sgsubheading}[1]{\par\medskip\noindent\textbf{#1}\par\vspace{0.1em}}

\section{Introduction}
\label{sec:introduction}

The emergence of smart glasses marks a shift from computing devices that users periodically consult to systems that can continuously share the user's viewpoint. Desktop computing requires deliberate operation, smartphones divert visual attention and occupy the hands, and immersive headsets provide high-bandwidth spatial interfaces at the cost of occlusion, weight, and limited wearing duration. Smart glasses instead place sensing, feedback, and interaction close to the everyday center of human vision, audition, and action. Depending on their hardware profiles, they can observe subsets of what the wearer sees, hears, says, attends to, and manipulates, while returning assistance through open-ear audio, a Head-up Display (HUD), or spatial Augmented Reality (AR) cues. 
\textbf{\textit{Their significance therefore lies not merely in miniaturizing cameras, displays, or conversational assistants, but in enabling a closed loop that connects first-person evidence, evolving contextual state, user intent, and digital or physical consequences.}} 
Early enterprise systems such as Google Glass Enterprise Edition~2 demonstrated the utility of hands-free documentation and workflow guidance~\cite{sgprod2026_google_google_glass_enterprise_edition_2}, while Ray-Ban Stories brought first-person capture into a familiar everyday-eyewear form factor~\cite{sgprod2026_meta_ray_ban_ray_ban_stories}. In parallel, Ego4D established long-form daily activity, episodic memory, and interaction as learnable first-person problems~\cite{Ego4D}, Ego-Exo4D aligned skilled behavior across first- and third-person views~\cite{Ego-Exo4D}, HoloAssist moved egocentric analysis toward interactive procedural assistance~\cite{Holoassist}, and Project Aria demonstrated the scientific value of synchronized RGB, audio, eye-tracking, inertial, and pose-related streams with calibrated data access~\cite{ProjectAria}. Recent camera- and audio-first smart glasses, camera-and-display systems, and research platforms expand the available hardware pathways for situated, increasingly agentic services. 
Representative examples include Ray-Ban Meta Gen~1~\cite{RayBanMeta}, Xiaomi AI Glasses~\cite{sgprod2026_xiaomi_xiaomi_ai_glasses}, Ray-Ban Meta Gen~2, Meta Ray-Ban Display, and Aria Gen~2~\cite{sgprod2026_meta_ray_ban_ray_ban_meta_gen_2,sgprod2026_meta_ray_ban_meta_ray_ban_display,sgprod2026_meta_reality_labs_research_aria_gen_2}.
\textit{We therefore view smart glasses as \textbf{hardware-constrained first-person intelligence platforms}: wearable systems that transform temporally aligned observations and user intent into feedback, persistent state, and, when authorized, digital or embodied action.}

This convergence has produced a rapidly growing but fragmented research landscape. 
\textbf{\textit{i)}} AR research has developed mature foundations for registration, display, gaze-supported interaction, and spatial user interfaces~\cite{GazePointAR,Visimark}, but frequently assumes richer optics, compute, or wearing conditions than lightweight everyday glasses. 
\textbf{\textit{ii)}} Egocentric vision has advanced action, hand-object, gaze, and fine-grained interaction understanding~\cite{EK-100,PredictingGaze,ModelingFGHandObjectDynamics,Annexe}, yet remains dominated by offline datasets and localized metrics. 
\textbf{\textit{iii)}} General-purpose multimodal models have substantially advanced visual-language understanding and instruction following~\cite{Gemini25,Qwen35,Roma,Vinci}. Multimodal benchmarks and systems for wearable settings now further study wearable question answering, situated awareness, streaming interaction, personal memory, and proactive assistance~\cite{WearVQA,Superglasses,SAW-Bench,EgoSAT,EgoMemReason,EgoStream,IPIBench}, but often abstract away camera placement, sensor dropout, network variability, output bandwidth, thermal throttling, and model or service drift. 
\textbf{\textit{iv)}} HCI and privacy research exposes interaction breakdowns, accessibility needs, social acceptability, and wearer-bystander tensions~\cite{FingerGlass,HCDesignAndFabrication,Conversational,CameraGlassesPrivacy,MindTheGap},
\textbf{\textit{v)}} whereas embodied-intelligence research increasingly exploits first-person human demonstrations~\cite{EgoScale,EgoVerse,Humanego} for imitation, Vision-Language-Action (VLA) learning, and robot-data synthesis~\cite{Egomimic,Egozero,EgoVLA,EgoEngine,Ego2Robot}.
Each line of work is necessary, but none alone establishes a deployable smart-glasses system. Three mismatches remain especially consequential. 
\textbf{\textit{First}}, a list of sensors or product functions does not determine which intelligent claims the hardware can actually support. 
\textbf{\textit{Second}}, component-level accuracy does not establish end-to-end utility when evidence can become stale, feedback can arrive outside its useful time window, and errors can propagate into memory or action. 
\textbf{\textit{Third}}, persistence and action authority qualitatively change the responsibility boundary: an incorrect answer may be retried, whereas a false memory, unauthorized purchase, unsafe instruction, or failed robot handoff can affect the wearer and other stakeholders long after the initiating observation. 
The central thesis of this survey is therefore that \textit{\textbf{Smart-glasses capability is a claim conditioned on hardware, temporal horizon, state persistence, action authority, operating environment, participant structure, and system version, as well as supporting evidence, rather than an intrinsic property of a device or model.}} A systematic account of the field must consequently answer four coupled questions: 
\textbf{\textit{1)}} what a glasses-based system can observe and communicate, 
\textbf{\textit{2)}} which composable mechanisms support first-person intelligence, 
\textbf{\textit{3)}} where those mechanisms create value and risk in real activities, and 
\textbf{\textit{4)}} how much evidence is sufficient to justify a capability or deployment claim.

\noindent\textbf{Scope.}\label{sec:intro-scope}
Rather than delimiting smart glasses through vendor terminology, consumer categories, or the presence of a display, this survey adopts a first-person data-flow and system-function perspective. Our primary analytical class comprises lightweight eyewear or glasses-mounted systems that participate in a wearer-aligned sensing, feedback, or interaction loop and provide at least one verifiable capability among \textbf{\textit{i)}} first-person visual or audio sensing, \textbf{\textit{ii)}} wearable audio or near-eye feedback, \textbf{\textit{iii)}} hands-free interaction, \textbf{\textit{iv)}} real-time AI assistance, and \textbf{\textit{v)}} spatial or contextual computing. We synthesize evidence from academic studies, datasets, benchmarks, research prototypes, developer platforms, and publicly documented commercial products, while distinguishing nominal feature availability from independently demonstrated system behavior. 
Neighboring classes are used as boundary references rather than merged into the same analytical category. Immersive Mixed-Reality (MR) headsets such as Meta Quest~3 and Apple Vision Pro provide upper-bound references for spatial interaction~\cite{sgprod2026_meta_meta_quest_3,apple_vision_pro}, Point-of-View (POV) and action cameras provide capture baselines~\cite{sgprod2026_gopro_gopro_hero13_black,sgprod2026_insta360_insta360_ace_pro_2}, dedicated egocentric data rigs inform embodied-data acquisition, mobile and web agents provide methods for planning and tool use, and non-eyewear wearables provide complementary physiological or interaction signals. These systems become relevant when their methods transfer to the wearer-aligned loop, but their results are not treated as direct evidence of eyewear-constrained deployability. Accordingly, the survey does not produce a universal product ranking. It instead asks which research claims are supportable for a specified task, hardware route, runtime, operating condition, and body of evidence. 
\textit{To the best of our knowledge, this is the first survey to jointly formalize smart glasses as claim-conditioned closed-loop systems and connect route-aware hardware profiles, capability levels, application-specific responsibility, and deployment evidence within a unified framework.}

\noindent\textbf{Contributions.}
Our main contributions are fourfold:
\begin{itemize}[leftmargin=1.4em,itemsep=0.18em,topsep=0.25em]

    \item \textbf{A formal and hardware-grounded problem definition.}
    We define smart glasses through a first-person observation stream, a closed-loop mapping from observation and intent to feedback, persistent state, and optional action, and a constrained utility objective that makes latency, energy, thermals, privacy, and social cost explicit. We further consolidate heterogeneous devices into eight verifiable device/platform capability axes and route-aware product profiles, converting marketing categories into evidence-bearing experimental substrates.

    \item \textbf{A compositional capability framework with explicit evidential boundaries.}
    We synthesize seven interdependent capabilities: first-person perception, multimodal context, persistent spatial state, auditable personal memory, situated agentic action, embodied data interfaces, and cross-cutting deployment constraints. Building upon these building blocks, we present an L0-L5 framework that covers capture, reactive perception, contextual assistance, persistent state, governed action, and embodied coupling. The framework treats levels as task- and evidence-conditioned claims, distinguishes prerequisites from demonstrated capability, and makes clear that L5 crosses the embodiment boundary rather than simply extending the wearer-facing L0-L4 axis.

    \item \textbf{An application-centered evidence map.}
    We reorganize the literature into nine application scenes and connect each scene to its required capability loop, representative datasets and benchmarks, research systems, product entry points, affected stakeholders, failure consequences, and missing validation evidence. This structure separates transferable component evidence from direct smart-glasses evidence and clarifies why identical model functions require different thresholds in daily assistance, accessibility, industry, healthcare, education, mobility, social collaboration, spatial intelligence, and embodied intelligence.

    \item \textbf{A deployment and evaluation blueprint.}
    We formulate nine coupled design dimensions covering hardware, runtime, perception and inference, memory, feedback, external action, reliability, governance, and reproducibility. On this basis, we provide a claim-conditioned evaluation protocol, a deployment checklist, and an iterative evidence ladder that connects documentation, laboratory measurement, benchmark testing, device-stream replay, fault injection, end-to-end studies, longitudinal deployment, and privacy or security audit. We finally derive eight open challenges and six roadmap directions for building trustworthy first-person embodied-intelligence systems.

\end{itemize}

\begin{figure}[t]
\centering
\includegraphics[width=\linewidth]{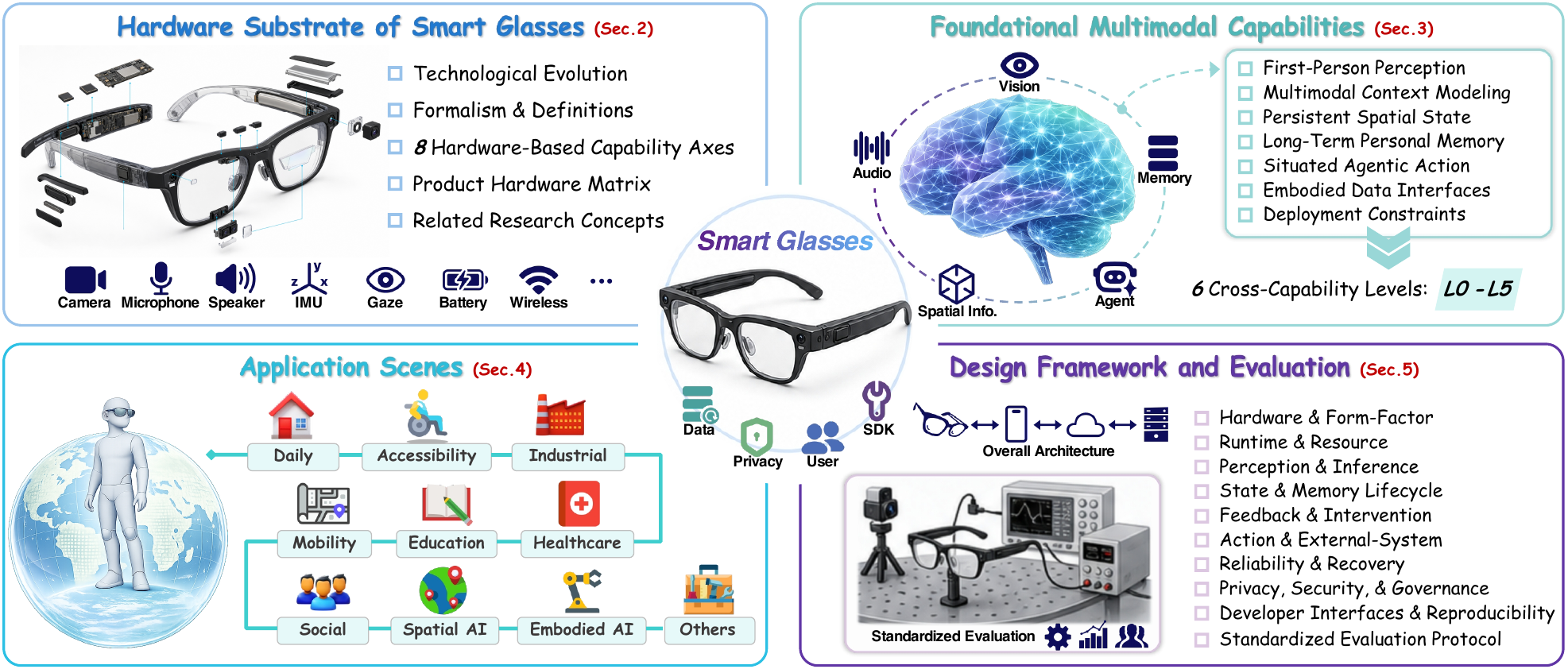}
\caption{\textbf{Organization and analytical loop of this survey.}
The survey connects the hardware substrate and first-person data-flow formulation (\cref{sec:background}), foundational capabilities and L0-L5 evidential levels (\cref{sec:capabilities}), application scenes and responsibility structures (\cref{sec:applications}), and deployment-oriented design and standardized evaluation (\cref{sec:design}). Evaluation evidence and observed failures feed back into system design and delimit the capability claims that can be defended.}
\label{fig:framework}
\vspace{-1em}
\end{figure}
\FloatBarrier

\noindent\textbf{Survey pipeline.}
As illustrated in \cref{fig:framework}, the survey is organized as a closed analytical loop from hardware substrate to capability, application, design, and evidence. \cref{sec:background} traces the evolution of smart glasses, formalizes their data flow and constrained objective, introduces the eight device/platform capability axes and representative product profiles, and clarifies boundaries with neighboring device classes. \cref{sec:capabilities} then abstracts the seven foundational capabilities and presents the L0-L5 cross-capability framework. \cref{sec:applications} reorganizes the literature around nine real-world scenes, with particular attention to the stronger state, stakeholder, and validation requirements of social collaboration, spatial intelligence, and embodied intelligence. \cref{sec:design} translates these requirements into a nine-dimensional deployment framework, standardized evaluation protocol, and design checklist. Finally, \cref{sec:conclusion} summarizes the principal system-level challenges and outlines a roadmap spanning reproducible platforms, privacy-aware longitudinal data, auditable memory, inclusive proactivity, interoperable action ecosystems, and robot-validated transfer. The loop is intentionally bidirectional: evaluation evidence and failure analysis feed back into hardware selection, runtime placement, state policy, interaction design, and the scope of future capability claims. 
We keep track of the latest works and products at \textbf{\href{https://github.com/zhangzjn/awesome-smart-glasses}{this project}}.

%% file: sec/02_background.tex
\begin{figure}[t]
\centering
\includegraphics[width=\linewidth]{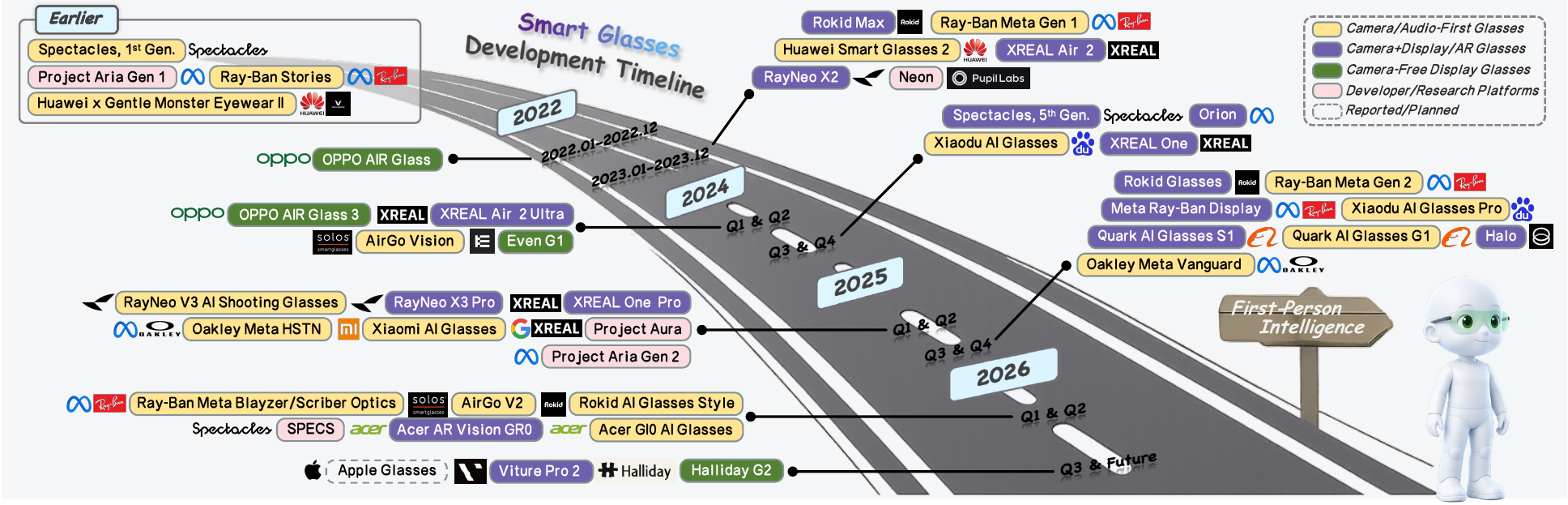}
\caption{\textbf{Evolution of mainstream smart-glasses products and platforms.} The timeline summarizes representative hardware entries and associated manufacturers over time. Fill colors denote four product profiles
in \cref{sec:background-evolution}, while dashed outlines denote reported entries that are not yet publicly available.}
\label{fig:timeline_glasses}
\vspace{-0.75em}
\end{figure}
\FloatBarrier

\section{Background}
\label{sec:background}

To establish the core concepts and provide a rigorous foundation for the remainder of this survey, we begin by tracing the evolution of smart glasses, from early explorations of the form factor, through the development of data and model infrastructure, to the recent diversification of product profiles in \cref{sec:background-evolution}. We then formalize smart glasses from a data-flow perspective and define their system functions and deployment constraints in \cref{sec:formulation}. Building on this formulation, we consolidate the hardware stack into eight verifiable capability axes in \cref{sec:hardware-axis} and introduce a typical product matrix in \cref{sec:product-matrix} to characterize the capabilities that different device profiles can support according to publicly available evidence. Finally, in \cref{sec:related-fields}, we position neighboring device classes, including immersive headsets and non-eyewear smart wearables, as boundary references for clarifying the scope of this survey.

\subsection{Evolution of Smart Glasses}
\label{sec:background-evolution}

\sgpoint{2013-2020: The era of early exploration}{
Early smart glasses primarily demonstrated the value of the eyewear form factor for first-person capture, notifications, remote expertise, procedural guidance, and low-friction information access. Typically, Google Glass Enterprise Edition~\cite{sgprod2026_google_google_glass_enterprise_edition_2} adapted the glass platform for industrial workflows through a lightweight head-up display, hands-free input, and task-specific software for manufacturing, logistics, and field service. Vuzix M100 and Vuzix M400~\cite{sgprod2026_vuzix_vuzix_m400} emphasized the coordination of camera sensing and visual display for enterprise settings, while Spectacles First Generation featured a built-in camera to allow users to capture first-person video~\cite{Spectacles}. Aria Gen 1 glasses released in 2020 was a research-grade sensing platform, designed to collect multimodal data for the development of future AR systems and egocentric AI research~\cite{ProjectAria}. The central limitation of this period was not the failure of any single product route, but the absence of an end-to-end closed loop integrating real-time perception and understanding, sustained user feedback, bystander privacy, mature developer ecosystems, reproducible evaluation, \etc
}

\sgpoint{2021-2024: Data and model infrastructure}{
Egocentric datasets, research-grade sensing platforms, and multimodal models collectively established much of the shared infrastructure required for intelligent eyewear. \Eg, Ego4D established a large-scale, geographically diverse foundation for long-form egocentric representation learning and multi-task evaluation~\cite{Ego4D}, EPIC-KITCHENS-100 provided dense and reproducible annotations for fine-grained procedural activity understanding in a controlled but realistic environment~\cite{EK-100}, HOI4D connected egocentric video analysis with geometric perception and explicit interaction structure, providing egocentric RGB-D sequences with 3D point-cloud information, hand/object masks, interaction labels, and motion annotations~\cite{HOI4D}, and HoloAssist framed procedural assistance and error recovery as interactive multimodal reasoning problems~\cite{Holoassist}. Together, these efforts shifted the central research question from whether glasses can capture first-person observations to whether they can infer task progress, contextual states, user intent, and actionable opportunities for assistance.
}

\sgpoint{2025-2026: Product-profile diversification}{
Recent products increasingly diverge into \textbf{four parallel routes}:
\textbf{\textit{i)}} camera/audio-first consumer glasses;
\textbf{\textit{ii)}} camera-and-display and AI-enabled AR consumer glasses;
\textbf{\textit{iii)}} lightweight camera-free HUD glasses; and
\textbf{\textit{iv)}} developer and research-oriented platforms, spanning developer ecosystems, advanced sensing, gaze tracking, \etc For example, typical Ray-Ban Meta Gen 2~\cite{sgprod2026_meta_ray_ban_ray_ban_meta_gen_2}, Quark AI Glasses G1~\cite{quark_ai_glasses_g1}, and Rokid AI Glasses Style~\cite{sgprod2026_rokid_rokid_ai_glasses_style} combine camera sensing, open-ear audio, and voice interaction, and emphasize everyday entry points. RayNeo X3 Pro~\cite{sgprod2026_tcl_rayneo_rayneo_x3_pro}, XREAL One Pro~\cite{XREAL_One_Pro}, and VITURE Pro 2~\cite{ViturePro2} represent the route combining visual sensing, near-eye displays, and spatial computing to support augmented information, virtual screens, navigation, and context-aware interaction. Halliday G2 provides discreet, glanceable information through a lightweight head-up display~\cite{Halliday_G2}, while Aria Gen 2 highlights the foundational value of research-grade synchronized sensing and raw data~\cite{sgprod2026_meta_reality_labs_research_aria_gen_2}. The resulting coexistence of diverse product profiles is summarized in \cref{fig:timeline_glasses}, which places representative entries on a common temporal axis. This diversification suggests that smart glasses are not advancing along a single linear axis; instead, they are redistributing hardware budgets across first-person sensing, low-burden feedback, spatial state, open interfaces, deployment credibility, \etc Rather than converging on a single canonical architecture, these routes reflect distinct trade-offs among sensing capability, display modality, on-device intelligence, interaction bandwidth, power and thermal constraints, privacy, and developer accessibility.
}

\subsection{Formalism and Definitions for Smart Glasses}
\label{sec:formulation}

The term \textit{smart glasses} now encompasses devices that differ substantially in sensing capabilities, feedback modalities, computing pathways, and system openness. A rigorous survey therefore requires a definition grounded in data-processing stack and underlying hardware rather than in product labels alone.

\sgpoint{Data flow formulation}{
Rather than relying on marketing terminology, we define smart glasses from a data-flow perspective. Consider a wearer continuously interacting with the physical environment. At time $t$, a smart-glasses system receives a first-person observation stream $\mathbf{o}^g_t$:
\begin{equation}
\mathbf{o}^g_t =
\left\{ I^{ego}_{t-k:t}, A_{t-k:t}, U_{t-k:t}, P_{t-k:t}, G_{t-k:t}, D_t \right\},
\label{eq:sg-observation}
\end{equation}
where $g$ denotes the glasses-mounted viewpoint; $I^{ego}_{t-k:t}$ denotes first-person visual observations in the past $k$ time intervals; $A$ denotes speech and ambient audio; $U$ denotes motion and localization cues obtained from sources such as Inertial Measurement Units (IMUs), Global Positioning System (GPS), and Visual-Inertial Odometry (VIO); $P$ denotes spatial and interaction-related cues, including gaze, hands, objects, and body pose; $G$ denotes explicit user interactions and other interaction signals; and $D_t$ denotes device state, permissions, and privacy policies. Recent work motivates treating $\mathbf{o}^g_t$ as a structured, temporally aligned stream rather than as a loose collection of sensor measurements. \Eg, Ego4D demonstrates the importance of long-form $I^{ego}_{t-k:t}$ for temporally extended activities and memory-oriented queries~\cite{Ego4D}, Project Aria highlights the value of synchronized and calibrated multimodal sensing for research-grade wearable platforms~\cite{ProjectAria}, and HoloAssist extended egocentric benchmarks from passive recognition toward context-aware, sensor-fused assistance~\cite{Holoassist}.
}

\sgpoint{From data flow to a closed-loop wearable system}{
Under this formulation, smart glasses are not merely cameras, displays, or assistant applications. Instead, they can be viewed as closed-loop wearable systems that transform first-person observations, internal state, user intent, and deployment constraints into user feedback, updated contextual state, and optional actions:
\begin{equation}
(y_t, s_{t+1}, \alpha_t) =
F_\theta\!\left(\mathbf{o}^g_{t-\tau:t}, s_t, q_t; \mathcal{B}\right),
\label{eq:sg-system}
\end{equation}
where $y_t$ denotes feedback delivered through audio, a HUD, AR overlays, or a companion device; $s_t$ denotes the system's evolving working context, spatial state, and personal memory; $\alpha_t$ denotes optional actions such as tool invocation, reminders, logging, remote collaboration, or interactions tied to the physical environment; $q_t$ denotes explicit or inferred user intent; and $\mathcal{B}$ denotes the deployment budget imposed by factors such as device weight, power consumption, thermal limits, network availability, and display requirements. The foundational capabilities discussed in \cref{sec:capabilities} decompose the perceptual, contextual, spatial, memory, agentic, and embodied-interface components underlying $F_\theta$, while \cref{sec:applications} examines how these capabilities support different user intents, tasks, environments, and risk conditions.
}

\sgpoint{Constrained objective}{
The research objective is not simply to maximize an offline model metric, but to maximize real-world task utility subject to wearable deployment constraints. A deployment-oriented objective can be expressed as:
\begin{equation}
\max_\theta \; \mathbb{E}\left[U(y_t,\alpha_t,s_{t+1})\right]
\quad
\mathrm{s.t.}\;
\ell_t \leq \ell_{\max},\;
e_t \leq e_{\max},\;
h_t \leq h_{\max},\;
r^{priv}_t \leq r_{\max},\;
c^{soc}_t \leq c_{\max},
\label{eq:sg-objective}
\end{equation}
where $\ell_t$, $e_t$, $h_t$, $r^{priv}_t$, and $c^{soc}_t$ denote latency, energy consumption, thermal load, privacy risk, and social disruption cost, respectively. This constrained formulation motivates the joint analysis of hardware capabilities, application requirements, and standardized evaluation. Without accounting for these deployment constraints, empirical performance measured in isolated settings may not reliably translate to practical smart-glasses systems operating in real-world environments.
}

\sgpoint{Claim-conditioned capability}{
A smart-glasses capability is not an intrinsic property of a device or model, but a conditional system claim. We represent such a claim as
\begin{equation}
\mathcal{C}
=
\left\langle
\mathcal{T},
\mathcal{H},
\mathcal{I},
\mathcal{Y},
\tau,
\mathcal{S},
\mathcal{A},
\mathcal{P},
\Omega,
\mathcal{V},
\mathcal{R}
\right\rangle,
\label{eq:capability-claim}
\end{equation}
where $\mathcal{T}$ denotes task scope; $\mathcal{H}$ the hardware and runtime profile; $\mathcal{I}$ and $\mathcal{Y}$ the available input and feedback channels; $\tau$ the temporal horizon and evidence-validity window; $\mathcal{S}$ the persistence and mutability of internal state; $\mathcal{A}$ the authority granted to external action; $\mathcal{P}$ the wearer, bystander, operator, and organizational participant structure; $\Omega$ the operating conditions; $\mathcal{V}$ the device, firmware, model, API, region, and service version; and $\mathcal{R}$ the task-specific risk tier. A body of evidence $\mathcal{E}$ supports the claim only under the stated tuple, denoted by $\mathcal{E}\models\mathcal{C}$. Changing any element produces a different capability claim, and empirical results should not be transferred across such changes without additional evidence.
}

\sgpoint{Analytical boundary}{
This survey focuses on smart glasses as first-person wearable computing systems. Neighboring device classes, including MR headsets, POV cameras, and robot-mounted or externally positioned cameras, are considered only as comparative references rather than as part of the same analytical class. Although these systems may share individual sensing, display, interaction, or intelligence capabilities with smart glasses, they differ in important assumptions regarding viewpoint, wearability, interaction, and deployment constraints. Establishing this boundary avoids the unqualified transfer of conclusions from adjacent systems to first-person intelligent eyewear designed under stringent wearable constraints. A detailed discussion of related concepts and their distinctions is provided in \cref{sec:related-fields}.
}

\subsection{Device/Platform Capability Consolidation}
\label{sec:hardware-axis}

\begin{figure}[t]
\centering
\includegraphics[width=\linewidth]{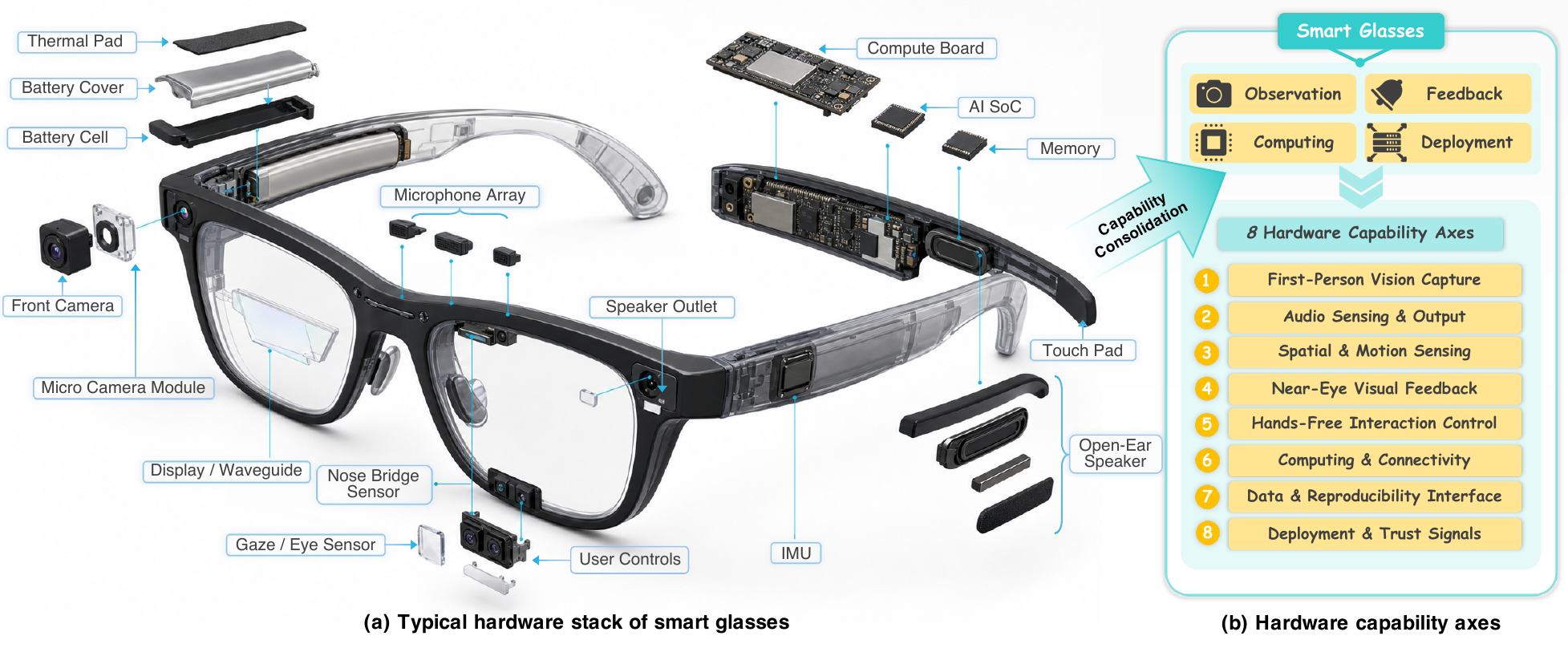}
\vspace{-1.8em}
\caption{\textbf{Hardware stack and capability-axis consolidation.} \textbf{Left panel (a)} presents the typical hardware components and placement constraints of smart glasses through a 45-degree exploded view, while \textbf{right panel (b)} consolidates the hardware profile into eight capability axes through four criteria: observation stream, feedback channel, computing path, and practical deployment. The exploded view is a conceptual schematic and does not represent a specific commercial device.
}
\label{fig:hardware-profile}
\vspace{-0.55em}
\end{figure}
\FloatBarrier

We organize hardware capabilities according to supported system functions in \cref{eq:sg-observation,eq:sg-system,eq:sg-objective}, guided by four diagnostic questions:
\textbf{\textit{i)} Observation stream.} Whether the component contributes to the first-person observation stream $\mathbf{o}^g_t$?
\textbf{\textit{ii)} Feedback channel.} Whether it determines the modality, bandwidth, user burden, or correctability of feedback $y_t$?
\textbf{\textit{iii)} Computing path.} Whether it shapes the local, edge, or cloud computation pathway of $F_\theta$, including interaction control and tool execution?
\textbf{\textit{iv)} Practical deployment.} Whether it affects latency, energy consumption, thermal load, privacy risk, social acceptability, or experimental reproducibility in real-world settings?

Following the criteria above, we consolidate the relevant hardware components and platform interfaces into eight capability axes, as illustrated in \cref{fig:hardware-profile}. These components and interfaces include cameras, microphones, IMUs, displays, interaction controls, systems-on-chip (SoC), connectivity modules, software development kits (SDKs), data-access interfaces, \etc Rather than treating hardware as a flat list of specifications, this consolidation provides a common analytical coordinate system for relating device capabilities to application requirements and standardized evaluation. \cref{tab:hardware-axis} further summarizes the atomic capabilities, representative algorithms, and application implications associated with each axis.

\begin{table}[t]
\centering
\caption{\textbf{Hardware capability axes and their corresponding research implications.}
}
\vspace{-0.55em}
\label{tab:hardware-axis}
\scriptsize
\begin{adjustbox}{width=\linewidth}
\begin{tabular}{
    >{\raggedright\arraybackslash}m{32mm}
    >{\raggedright\arraybackslash}m{42mm}
    >{\raggedright\arraybackslash}m{50mm}
    >{\raggedright\arraybackslash}m{50mm}
}
\toprule
\textbf{Capability Axis} & \textbf{Atomic Capabilities} & \textbf{Algorithmic Implications} & \textbf{Application Implications} \\
\midrule

\ding{172} \textbf{First-Person Vision Capture}
& RGB/wide-angle camera, multi-camera configuration, rolling/global shutter, low-light support, recording indicator, \etc
& Shapes egocentric VQA, OCR, active-object understanding, hand-object modeling, motion-robust perception, privacy-aware visual processing, \etc
& Supports everyday question answering, accessible reading, industrial documentation, mobile safety, first-person demonstration capture, \etc \\

\midrule
\ding{173} \textbf{Audio Sensing and Output}
& microphone array, beamforming, open-ear speaker, bone conduction, noise/wind suppression, audio-leakage control, \etc
& Shapes ASR, speaker and acoustic-event understanding, turn-taking, barge-in, speech-based assistance, alert delivery, \etc
& Supports translation, meetings, hearing assistance, sports feedback, low-burden conversational assistance, \etc \\

\midrule
\ding{174} \textbf{Spatial and Motion State Sensing}
& IMU, GPS/UWB, SLAM/VIO sensing, depth sensing, gaze and hand tracking, relocalization, calibration, \etc
& Supports pose continuity, semantic mapping, spatial memory, world-locked cues, object persistence, ego-exo alignment, \etc
& Enables indoor navigation, spatial reminders, AR annotation, skill training, embodied data collection, \etc \\

\midrule
\ding{175} \textbf{Near-Eye Visual Feedback}
& no display, monochrome HUD, monocular/binocular display, optical see-through waveguide, brightness, FOV, \etc
& Shapes feedback bandwidth, confirmation, captioning, spatial cueing, uncertainty presentation, distraction risk, \etc
& Supports low-vision assistance, procedural guidance, mobile safety, spatial-intelligence applications, context-aware visual assistance, \etc \\

\midrule
\ding{176} \textbf{Hands-Free Interaction Control}
& voice, button, touch, head gesture, gaze, ring, EMG, hand tracking, \etc
& Supports wake-up, pointing, confirmation, correction, undo, permission gating, control over system proactivity, \etc
& Shapes usability in public spaces, PPE-constrained environments, sports scenarios, accessibility settings, safety-critical interactions, \etc \\

\midrule
\ding{177} \textbf{Computing and Connectivity Architecture}
& glasses-only, phone-tethered, cloud-assisted, edge-cloud hybrid, compute puck, local enterprise server, \etc
& Shapes latency, energy consumption, model capacity, offline fallback, data boundaries, memory access, tool execution, \etc
& Affects all-day operation, offline industrial use, regulated deployments, cross-device agents, service reliability, \etc \\

\midrule
\ding{178} \textbf{Data and Reproducibility Interfaces}
& SDK, raw-sensor access, timestamps, calibration, pose/map API, memory API, log schema, model/firmware versioning, \etc
& Enables dataset release, benchmark construction, failure attribution, experimental replay, third-party auditing, longitudinal reproducibility, \etc
& Supports academic experimentation, enterprise integration, privacy auditing, ecosystem development, \etc \\

\midrule
\ding{179} \textbf{Deployment Envelope and Trust Signals}
& weight, battery life, thermal behavior, IP rating, prescription support, fit, audio leakage, physical shutter/mute, visible recording state, \etc
& Constrains duty cycle, long-term sensing quality, thermal throttling, wearing comfort, bystander awareness, \etc
& Shapes user acceptance, public-space deployment, scenario boundaries, regulatory compatibility, product maturity, \etc \\

\bottomrule
\end{tabular}
\end{adjustbox}
\end{table}

\begin{itemize}[leftmargin=1.4em,itemsep=0.18em,topsep=0.25em]

\item \textbf{First-Person vision capture}
directly determines whether and how reliably a smart-glasses system can form $I^{ego}$ within the observation stream $\mathbf{o}^g_t$. Camera field of view, single- or multi-camera geometry, shutter mechanism, low-light performance, and visible recording indicators jointly shape the fidelity, temporal stability, and social transparency of first-person visual observations. These properties, in turn, constrain the reliability of egocentric VQA, Optical Character Recognition (OCR), active-object understanding, hand-object modeling, motion-robust perception, and privacy-aware visual processing.

\item \textbf{Audio sensing and output}
couples environmental perception with low-burden interaction and feedback. Microphone arrays, beamforming, wind and noise suppression, open-ear speakers, bone conduction, and audio-leakage control shape Automatic Speech Recognition (ASR), speaker and acoustic-event understanding, turn-taking, barge-in, and alert delivery under mobile acoustic conditions. Audio should therefore be viewed not merely as a secondary modality to vision, but as a bidirectional interface through which environmental context, user intent, human emotion, system responses, and conversational state are continuously exchanged.

\item \textbf{Spatial and motion state sensing}
transforms transient first-person observations into persistent relationships among the wearer, surrounding objects, places, and ongoing actions. IMUs, GPS or Ultra-Wideband (UWB), sensing configurations supporting Simultaneous Localization and Mapping (SLAM) and Visual-Inertial Odometry (VIO), depth sensing, gaze tracking, hand tracking, relocalization, and calibration provide the foundation for pose continuity, semantic mapping, spatial memory, object persistence, and world-locked cues. Spatial capability should therefore be understood as a state-estimation and grounding substrate rather than as a synonym for near-eye visual display.

\item \textbf{Near-eye visual feedback}
defines the visual bandwidth through which a system can communicate results, uncertainty, guidance, and opportunities for correction to the wearer. The design space ranges from display-free devices and monochrome HUDs to monocular or binocular optical see-through displays, with brightness, Field of View (FOV), Pixels per Degree (PPD), outdoor readability, and occlusion jointly determining the usable visual-feedback envelope. Confirmation, captioning, directional guidance, uncertainty visualization, and spatial annotation are beneficial only when the information they provide justifies the associated visual, cognitive, and attentional costs, particularly during locomotion.

\item \textbf{Hands-free interaction control}
provides the operational layer through which users express intent, issue corrections, and authorize system actions. Voice, buttons, touch surfaces, head gestures, gaze, ring, Electromyography (EMG), and hand tracking can instantiate wake-up, pointing, selection, confirmation, correction, undo, and permission-gating mechanisms. In public spaces, sports settings, industrial environments involving Personal Protective Equipment (PPE), and accessibility scenarios, this interaction layer can define the practical boundary of system usability even before model accuracy becomes the dominant limiting factor.

\item \textbf{Computing and connectivity architecture}
determines where perception, inference, memory access, privacy filtering, and tool execution are performed. Glasses-only, phone-tethered, cloud-assisted, edge-cloud hybrid, compute-puck, and enterprise local-server architectures impose different latency distributions, energy budgets, model-capacity limits, offline fallback behavior, network dependencies, and data-governance boundaries. Consequently, empirical conclusions obtained using the same multimodal model may not transfer directly across on-device, phone-assisted, cloud-based, and enterprise-network deployments.

\item \textbf{Data and reproducibility interfaces}
determine the extent to which a smart-glasses platform can support reproducible scientific investigation beyond vendor demonstrations. SDKs, raw-sensor access, synchronized timestamps, calibration parameters, pose or map Application Programming Interfaces (APIs), memory APIs, log schemas, and model or firmware versioning enable benchmark construction, failure attribution, experimental replay, third-party auditing, dataset creation, and longitudinal comparison. For rigorous system research, the interface profile is therefore part of the experimental specification rather than ancillary product metadata.

\item \textbf{Deployment envelope and trust signals}
capture whether technical capabilities remain usable under sustained physical, environmental, and social constraints. Weight, battery life, thermal behavior, Ingress Protection (IP) rating, prescription-lens support, fit, audio leakage, physical shutter or mute controls, and visible recording indicators jointly shape duty cycle, wearing comfort, sensing stability, and bystander awareness. These factors connect device engineering to user acceptance, public-space deployment, and compliance requirements in medical, industrial, educational, and other risk-sensitive settings.

\end{itemize}

\providecommand{\sgfilledstar}{\textcolor{sgorange}{\ensuremath{\bigstar}}}
\providecommand{\sgemptystar}{\textcolor{lightgray}{\ensuremath{\bigstar}}}
\providecommand{\sgstars}[1]{%
  \ifcase#1\sgemptystar\kern0.08em\sgemptystar%
  \or\sgfilledstar\kern0.08em\sgemptystar%
  \or\sgfilledstar\kern0.08em\sgfilledstar%
  \else\sgemptystar\kern0.08em\sgemptystar%
  \fi}

\begin{table}[t]
\centering
\caption{\textbf{Hardware capability profiles and estimated levels of representative smart-glasses products.}
\textit{\textbf{Date}} denotes the public release or launch timing of commercial products and the announcement, pre-order, or research-access timing of developer kits and research platforms.
\textit{\textbf{Type}} summarizes the dominant hardware route.
\textit{Representative claim \textbf{Level}} follows the L0-L5 interpretation defined in \cref{sec:capability-levels}.
The eight capability columns correspond to the axes defined in \cref{sec:hardware-axis}, and ratings follow a two-star public-evidence scale (
\sgstars{0}: no publicly documented evidence of the capability, 
\sgstars{1}: basic support, and 
\sgstars{2}: strong or research-usable support).
\textit{Hardware Positioning} summarizes the primary research affordances and major limiting factors of each device.
The assigned level is task- and evidence-conditioned and should not be interpreted as a device maturity score.
}
\label{tab:product-hardware}

\scriptsize
\begin{adjustbox}{width=\linewidth}
\begin{tabular}{
>{\raggedright\arraybackslash}m{16mm}
>{\raggedright\arraybackslash}m{10mm}
>{\raggedright\arraybackslash}m{12mm}
>{\raggedright\arraybackslash}m{8mm}
cccccccc
>{\raggedright\arraybackslash}m{70mm}
}
\toprule
\textbf{Device} &
\textbf{Date} &
\textbf{Type} &
\textbf{Level} &
\textbf{Vision} &
\textbf{Audio} &
\textbf{Spatial} &
\textbf{Feedback} &
\textbf{Control} &
\textbf{Computing} &
\textbf{Interfaces} &
\textbf{Deployment} &
\textbf{Hardware Positioning} \\
\midrule

\rowcolor{aprilblue!20}
\multicolumn{13}{l}{\textit{\textbf{Camera/Audio-First Consumer Glasses}}} \\

Xiaomi AI Glasses~\cite{sgprod2026_xiaomi_xiaomi_ai_glasses}
& 2025-06
& camera/audio-first
& L2
& \sgstars{2} & \sgstars{2} & \sgstars{0} & \sgstars{1}
& \sgstars{1} & \sgstars{1} & \sgstars{0} & \sgstars{1}
& Offers tight integration with Chinese service ecosystems and assistant functions, but public evidence of raw-data access, gaze sensing, and persistent spatial state remains limited. \\
\cline{13-13}

Ray-Ban Meta Gen 2~\cite{sgprod2026_meta_ray_ban_ray_ban_meta_gen_2}
& 2025-09
& camera/audio-first
& L2
& \sgstars{2} & \sgstars{2} & \sgstars{0} & \sgstars{1}
& \sgstars{1} & \sgstars{1} & \sgstars{0} & \sgstars{2}
& Provides a strong daily-wear platform for L2 assistance and lightweight memory support, while persistent spatial state and open research interfaces remain limited. \\
\cline{13-13}

Oakley Meta Vanguard~\cite{sgprod2026_meta_oakley_oakley_meta_vanguard}
& 2025-10
& camera/audio-first
& L2
& \sgstars{2} & \sgstars{2} & \sgstars{0} & \sgstars{1}
& \sgstars{1} & \sgstars{1} & \sgstars{0} & \sgstars{2}
& Provides a strong deployment profile for sports and outdoor activities, highlighting the value of egocentric capture and open-ear feedback under high-activity conditions. \\
\cline{13-13}

Rokid AI Glasses Style~\cite{sgprod2026_rokid_rokid_ai_glasses_style}
& 2026-01
& camera/audio-first
& L2
& \sgstars{2} & \sgstars{1} & \sgstars{0} & \sgstars{1}
& \sgstars{1} & \sgstars{1} & \sgstars{0} & \sgstars{2}
& Its lightweight form factor exemplifies a wearability-oriented route for display-free AI glasses, although feedback bandwidth remains constrained. \\
\cline{13-13}

Solos AirGo V2~\cite{sgprod2026_solos_solos_airgo_v2}
& 2026-01
& camera/audio-first
& L2
& \sgstars{2} & \sgstars{1} & \sgstars{0} & \sgstars{1}
& \sgstars{1} & \sgstars{1} & \sgstars{1} & \sgstars{2}
& Modular camera support and multi-model access provide a relatively accessible platform for lightweight experimentation, but spatial capabilities remain limited. \\

\specialrule{0.25pt}{0.18em}{0.08em}
\rowcolor{aprilblue!20}
\multicolumn{13}{l}{\textit{\textbf{Camera-and-Display and AI-Enabled AR Consumer Glasses}}} \\

Meta Ray-Ban Display~\cite{sgprod2026_meta_ray_ban_meta_ray_ban_display}
& 2025-09
& camera-and-display
& potential L3
& \sgstars{2} & \sgstars{2} & \sgstars{1} & \sgstars{2}
& \sgstars{2} & \sgstars{1} & \sgstars{0} & \sgstars{1}
& Camera, visual display, and Neural Band interaction strengthen confirmation and correction loops, creating potential for L3 memory-centric assistance, but public evidence does not yet support mature L4 action loops. \\
\cline{13-13}

Quark AI Glasses S1~\cite{sgprod2026_alibaba_quark_alibaba_quark_ai_glasses_s1}
& 2025-11
& camera-and-display
& L2
& \sgstars{2} & \sgstars{2} & \sgstars{0} & \sgstars{2}
& \sgstars{1} & \sgstars{1} & \sgstars{0} & \sgstars{1}
& Integration with Chinese digital services, translation, and payment functions broadens tool access, but persistent spatial state, user-correctable memory, and reproducibility interfaces remain insufficiently documented for L3. \\
\cline{13-13}

RayNeo X3 Pro~\cite{sgprod2026_tcl_rayneo_rayneo_x3_pro}
& 2025-12
& AI-enabled AR glasses
& potential L3
& \sgstars{2} & \sgstars{1} & \sgstars{1} & \sgstars{2}
& \sgstars{1} & \sgstars{1} & \sgstars{1} & \sgstars{1}
& Binocular display and 6-DoF tracking enable spatial interfaces and spatiotemporal-memory experimentation, while all-day wearability, user-correctable memory, and ecosystem maturity require further validation. \\

\specialrule{0.25pt}{0.18em}{0.08em}
\rowcolor{aprilblue!20}
\multicolumn{13}{l}{\textit{\textbf{Lightweight Camera-Free HUD Glasses}}} \\

Even Realities G1~\cite{sgprod2026_even_realities_even_realities_g1}
& 2024-06
& camera-free HUD
& L1
& \sgstars{0} & \sgstars{1} & \sgstars{0} & \sgstars{2}
& \sgstars{1} & \sgstars{1} & \sgstars{0} & \sgstars{2}
& Privacy-oriented HUD design is well suited to captions, prompts, and teleprompter-style cues, but the absence of an onboard camera precludes first-person visual understanding. \\
\cline{13-13}

Vuzix Z100~\cite{sgprod2026_vuzix_vuzix_z100}
& 2024-11
& camera-free HUD
& L1
& \sgstars{0} & \sgstars{1} & \sgstars{0} & \sgstars{2}
& \sgstars{1} & \sgstars{1} & \sgstars{1} & \sgstars{2}
& Provides a clear enterprise cueing and power-efficient display profile, but is not designed as a general-purpose multimodal assistant. \\
\cline{13-13}

Halliday G2~\cite{Halliday_G2}
& 2026-07
& camera-free HUD
& L1
& \sgstars{0} & \sgstars{1} & \sgstars{0} & \sgstars{2}
& \sgstars{1} & \sgstars{1} & \sgstars{0} & \sgstars{1}
& Provides ambient assistance through a HUD-oriented consumer design that prioritizes everyday wearability and unobtrusive interaction over immersive spatial computing or research-grade sensing. \\

\specialrule{0.25pt}{0.18em}{0.08em}
\rowcolor{aprilblue!20}
\multicolumn{13}{l}{\textit{\textbf{AR Developer and Research-Oriented Platforms}}} \\

Pupil Labs Neon~\cite{sgprod2026_pupil_labs_pupil_labs_neon}
& 2023-02
& research platform
& partial L5
& \sgstars{2} & \sgstars{1} & \sgstars{1} & \sgstars{0}
& \sgstars{0} & \sgstars{1} & \sgstars{2} & \sgstars{1}
& High-quality gaze sensing and open data interfaces support HCI research, attention modeling, and learning from demonstration, while user-facing feedback and closed-loop assistance are largely absent. \\
\cline{13-13}

Project Aria~\cite{ProjectAria,sgprod2026_meta_reality_labs_research_aria_gen_2}
& Gen 1: 2020; Gen 2: 2025
& research platform
& L5 (data collection)
& \sgstars{2} & \sgstars{2} & \sgstars{2} & \sgstars{0}
& \sgstars{0} & \sgstars{1} & \sgstars{2} & \sgstars{1}
& Synchronized multimodal sensing, gaze and pose estimation, calibration, and raw-data access make the platform suitable for dataset construction and embodied-perception research rather than consumer-facing assistance. \\
\cline{13-13}

XREAL AURA~\cite{sgprod2026_xreal_google_xreal_aura_project_aura}
& 2025-05
& AR developer platform
& potential L4
& \sgstars{1} & \sgstars{1} & \sgstars{2} & \sgstars{2}
& \sgstars{1} & \sgstars{2} & \sgstars{2} & \sgstars{0}
& Its split-compute architecture and Android XR ecosystem provide a strong basis for spatial-application development, while device availability, battery life, and deployment maturity remain to be validated. \\
\cline{13-13}

SPECS~\cite{sgprod2026_snap_snap_specs_2026}
& 2026-06
& AR developer platform
& potential L4
& \sgstars{2} & \sgstars{1} & \sgstars{2} & \sgstars{2}
& \sgstars{2} & \sgstars{1} & \sgstars{2} & \sgstars{0}
& Offers a high ceiling for spatial display, hand tracking, and interactive AR, making it suitable for exploratory L4 spatial- intelligence research rather than mature all-day consumer deployment. \\

\bottomrule
\end{tabular}
\end{adjustbox}
\vspace{-1em}
\end{table}

\subsection{Typical Product Hardware Matrix}
\label{sec:product-matrix}

\cref{tab:product-hardware} maps representative products and research platforms onto route-aware hardware profiles discussed in \cref{sec:background-evolution}, thereby operationalizing the eight capability axes introduced in \cref{sec:hardware-axis}. The \textit{Level} column uses the cross-capability L0-L5 framework defined in \cref{sec:capability-levels}. Rather than ranking devices, the matrix is intended to clarify which classes of research claims are supported by publicly documented hardware capabilities and which remain insufficiently substantiated. Across the four routes, a central pattern emerges: capability progression is neither monotonic in sensor count nor proportional to display complexity. Instead, each route allocates the constrained smart-glasses design budget differently across the eight device/platform capability axes, yielding distinct strengths, limitations, and research affordances.

\textbf{\textit{i)}} Camera/audio-first consumer glasses such as Xiaomi AI Glasses~\cite{sgprod2026_xiaomi_xiaomi_ai_glasses}, Ray-Ban Meta Gen 2~\cite{sgprod2026_meta_ray_ban_ray_ban_meta_gen_2}, Oakley Meta Vanguard~\cite{sgprod2026_meta_oakley_oakley_meta_vanguard}, Rokid AI Glasses Style~\cite{sgprod2026_rokid_rokid_ai_glasses_style}, and Solos AirGo V2~\cite{sgprod2026_solos_solos_airgo_v2} provide the clear near-term evidence for L2 contextual assistance. However, limited persistent spatial state and restricted research interfaces constrain stronger claims regarding longitudinal memory or closed-loop action. 
\textbf{\textit{ii)}} Camera-and-display and AI-enabled AR consumer glasses, including Meta Ray-Ban Display~\cite{sgprod2026_meta_ray_ban_meta_ray_ban_display}, Quark AI Glasses S1~\cite{sgprod2026_alibaba_quark_alibaba_quark_ai_glasses_s1}, and RayNeo X3 Pro~\cite{sgprod2026_tcl_rayneo_rayneo_x3_pro}, increase feedback bandwidth and strengthen confirmation and correction loops, creating plausible entry points toward L3 when memory provenance, correction, deletion, and long-term stability can be verified. 
\textbf{\textit{iii)}} Lightweight camera-free HUD type, such as Even Realities G1~\cite{sgprod2026_even_realities_even_realities_g1}, Vuzix Z100~\cite{sgprod2026_vuzix_vuzix_z100}, and Halliday G2~\cite{Halliday_G2}, demonstrate the value of low-power, comparatively unobtrusive visual cueing while deliberately trading away first-person visual perception. 
\textbf{\textit{iv)}} AR developer and research-oriented platforms serve two distinct evidential roles. True-AR and developer platforms, represented by XREAL XREAL AURA~\cite{sgprod2026_xreal_google_xreal_aura_project_aura} and SPECS~\cite{sgprod2026_snap_snap_specs_2026}, raise the ceiling for spatial interaction, multimodal control, and developer-facing tool ecosystems. Their interpretation as deployable L4 systems, however, still depends on evidence of availability, thermal stability, battery endurance, and sustained field performance. Research sensing and gaze platforms, including Aria Gen 1 glasses~\cite{ProjectAria}, Pupil Labs Neon~\cite{sgprod2026_pupil_labs_pupil_labs_neon}, and Aria Gen 2 glasses~\cite{sgprod2026_meta_reality_labs_research_aria_gen_2}, occupy a complementary role: synchronized raw sensing, calibration, gaze and pose estimation, and open data interfaces make them particularly valuable for dataset construction, egocentric vision and embodied-AI research, and reproducible measurement.

\subsection{Related Concepts and Disambiguation}
\label{sec:related-fields}

In this section, we distinguish smart glasses from neighboring device and system classes that share subsets of their sensing, feedback, interaction, or agentic capabilities but operate under different physical and deployment assumptions. Rather than excluding these adjacent classes from consideration, we use them as boundary references to clarify which capabilities and methodological insights can inform smart-glasses research and which conclusions do not transfer directly to eyewear-constrained, first-person intelligent systems.

\begin{itemize}[leftmargin=1.4em,itemsep=0.18em,topsep=0.25em]

\item \textbf{Immersive headsets.}
Meta Quest 3~\cite{sgprod2026_meta_meta_quest_3}, Apple Vision Pro~\cite{apple_vision_pro}, Samsung Galaxy XR~\cite{samsung_galaxy_xr}, HoloLens 2~\cite{sgprod2026_microsoft_microsoft_hololens_2}, and Magic Leap 2~\cite{sgprod2026_magic_leap_magic_leap_2} fall outside the primary analytical class of smart glasses, despite providing high-end reference points for passthrough perception, hand tracking, eye tracking, spatial user interfaces, and dense multimodal sensing. Their occlusive or headset-like form factors, greater weight, shorter practical wearing duration, more restrictive public-space use, and larger power and thermal budgets distinguish them from lightweight eyewear intended for sustained use. Immersive headsets are therefore most useful as upper-bound references for spatial interaction and gaze/hand sensing, rather than as direct evidence of deployability or usability for smart glasses.

\item \textbf{Action cameras and body cameras.}
Insta360 X4~\cite{sgprod2026_insta360_insta360_x4} and GoPro MAX 2~\cite{sgprod2026_gopro_gopro_max_2} provide useful recording baselines for long-form Point-of-View (POV) capture, wide-angle video, and multi-view documentation. AoE~\cite{AoE} and Open-AoE~\cite{Open-AoE} leverage neck-mounted smartphones for in-the-wild egocentric human video collection for embodied AI. However, the absence of an eyewear form factor and functions such as wearer-aligned gaze sensing, natural-language interaction, user-facing feedback, and real-time agentic loops, places this class outside our analytical definition of smart glasses. Action cameras and body cameras are therefore useful for studying viewpoint bias, capture quality, and recording continuity, whereas claims about smart-glasses assistance additionally require evidence of interaction, feedback, privacy management, and situated reasoning.

\item \textbf{Egocentric capture equipment dedicated to embodied AI.}
Recent embodied-AI research~\cite{DexterityFromSmartLenses,Ego-Pi,Qwen-RobotManip,HOST} has boosted a distinct class of egocentric capture equipment designed specifically to acquire human demonstrations and multimodal first-person data for robot policy learning~\cite{RT-1,OpenVLA,Pi0}. These systems are typically head-mounted or body-worn rigs that prioritize synchronized sensing, stable calibration, and high-throughput data logging across RGB cameras, depth sensors, inertial units, microphones, and, where available, hand or eye-tracking interfaces. \Eg, DAS Ego is a dedicated head-worn capture system for recording first-person human behavior and demonstrations for embodied-intelligence datasets~\cite{DAS-Ego}, Pika Pro positions wearable sensing within an industrial workflow for acquiring operator demonstrations and supporting robot imitation learning~\cite{PikaPro}, Ropedia provides a full-stack platform for collecting, structuring, and managing in-the-wild human data~\cite{Ropedia}, EgoLive leverages a custom-designed head-mounted device, JoyEgoCam, for human behavior acquisition in real-world environments~\cite{EgoLive}, and Ego-OSCAR introduces an open-source, wearable stereo-capture system which supports geometric perception, 3D scene understanding, human–object interaction analysis, and learning from human demonstrations~\cite{Ego-OSCAR}. These systems are adjacent to, but remain outside the scope of, smart glasses as defined in this survey, though providing important methodological references for embodied data interfaces.

\item \textbf{Mobile and web agents.}
Mobile-Agent~\cite{MobileAgent}, MobileForge~\cite{MobileForge}, MemGUI-Agent~\cite{MemGUI-Agent}, and WebArena~\cite{Webarena} provide important references for agentic tool use in mobile and web environments, including screen understanding, application control, planning, and task execution through digital interfaces. These agents are grounded primarily in graphical user interfaces rather than in continuously changing first-person observations of the physical world, and therefore fall outside the primary scope of smart-glasses systems. Nevertheless, they offer transferable methodology for planning, tool invocation, recovery, and action-success evaluation. Smart-glasses agents additionally require mechanisms for wearable feedback, permission control, bystander awareness, situated perception, and physical-world risk management.

\item \textbf{Non-eyewear smart wearables.}
Non-eyewear smart wearables, including smartwatches (\eg, Apple Watch~\cite{apple_watch}), smart bands (\eg, Xiaomi Smart Band 9~\cite{xiaomi_smart_band_9}), smart rings (\eg, Oura Ring 4~\cite{sgprod2026_oura_ring_4} and Samsung Galaxy Ring~\cite{sgprod2026_samsung_galaxy_ring}), and earring-style hearables (\eg, NOVA H1 Audio Earrings~\cite{sgprod2026_nova_h1_audio_earrings}), likewise fall outside the analytical class because they lack a glasses-mounted first-person viewpoint, a near-eye feedback channel, \etc Their relationship to smart glasses is therefore primarily complementary rather than substitutive: physiological sensing, low-power state tracking, haptic or audio feedback, and cross-device context can enrich multi-device assistants. These devices are best interpreted as companion sensing and interaction substrates rather than as direct evidence for egocentric visual understanding, persistent spatial and temporal memory, or glasses-mounted agentic action.

\end{itemize}

%% file: sec/03_capabilities.tex
\section{Foundational Capabilities for Smart Glasses}
\label{sec:capabilities}

Smart glasses are formalized in \cref{sec:background} along three dimensions: data streams, system mappings, and constrained objectives, with the underlying hardware substrate further summarized through eight verifiable capability axes. Building on this foundation, this section examines \textit{\textbf{how hardware conditions support composable and evaluable mechanisms for first-person intelligence}}. These foundational capabilities therefore constitute an intermediate analytical layer that connects hardware profiles to datasets and benchmarks, models, application scenarios, and ultimately real-world deployment.

\begin{figure}[t]
\centering
\includegraphics[width=\linewidth]{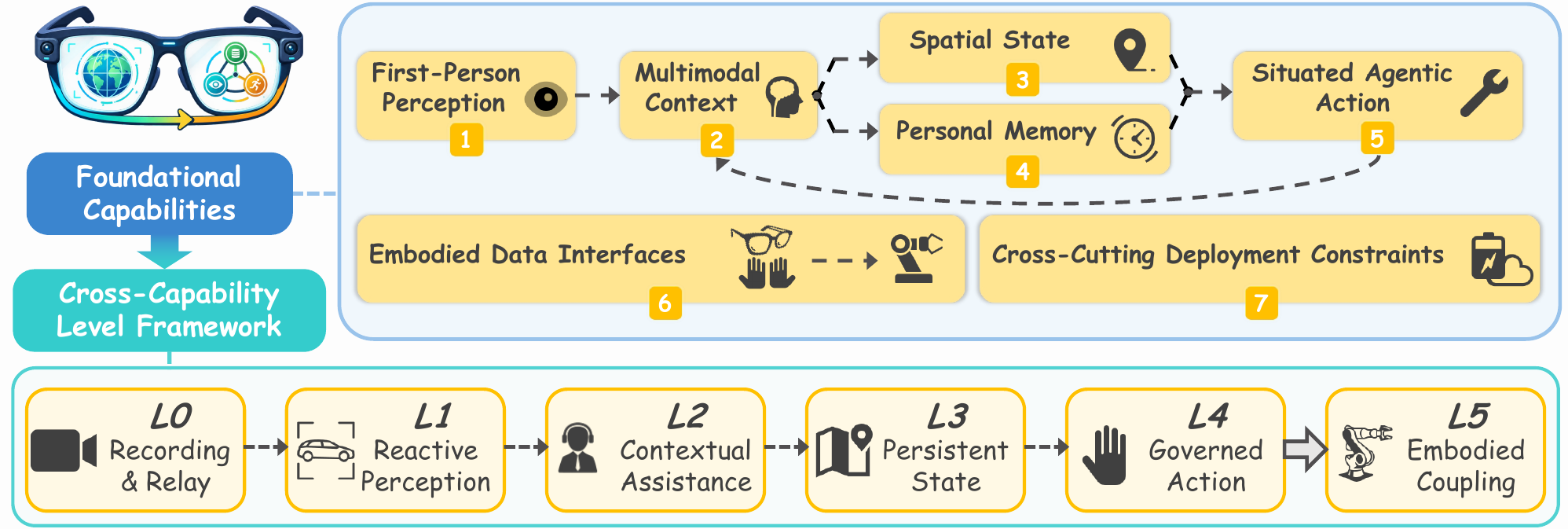}
\caption{\textbf{Foundational capabilities and L0-L5 cross-capability level framework.}
\textbf{\textit{1)}} First-person perception,
\textbf{\textit{2)}} multimodal context,
\textbf{\textit{3)}} spatial state,
\textbf{\textit{4)}} personal memory, and
\textbf{\textit{5)}} situated action form the core dependency structure of first-person intelligence;
\textbf{\textit{6)}} embodied data interfaces extend this structure toward robot learning and cross-embodiment transfer; and
\textbf{\textit{7)}} deployment constraints cut across all capabilities.
\textbf{Bottom}: L0-L4 form a wearer-facing progression, whereas L5 is an orthogonal cross-embodiment extension in \cref{sec:capability-levels}.
}
\label{fig:capability-ladder}
\end{figure}

As illustrated in \cref{fig:capability-ladder}, these capabilities exhibit a directional dependency structure rather than a simple linear pipeline. First-person perception establishes the evidential basis for downstream reasoning~\cite{Ego4D,Ego-Exo4D,H2O,EgoLive}; multimodal context modeling integrates heterogeneous observations while preserving temporal and cross-modal relationships~\cite{ContextAgent,ProAgent}; spatial state maintains persistent geometric and object-level context across locations~\cite{Ariadigitaltwin,Pandora}; personal memory extends system state across events and time~\cite{Egolife,LightMem-Ego,EgoMonth,EgoSchema}; and situated action closes the perception-state-action loop through context-aware and permission-governed execution~\cite{AI4Service,Ego2Web,VisionClaw}. Embodied data interfaces form an outward-facing branch that transforms first-person observations and internal representations into resources for capturing human experience, aligning ego-exo perspectives, and transferring knowledge across embodiments~\cite{Egomimic,Qwen-RobotManip,ACE-Ego-0}. Deployment constraints, by contrast, operate across sensing, inference, state maintenance, feedback, memory, and action rather than emerging only at the end of the pipeline~\cite{HCDesignAndFabrication,EPIC,OpenGlass}. Building on these seven foundational capabilities, we finally introduce the L0-L5 framework in \cref{sec:capability-levels} to connect capability-level analysis with scenario-based validation and deployment-oriented evaluation.

\FloatBarrier

\subsection{First-Person Perception}
\label{sec:first-person-perception}

First-person perception determines whether downstream systems receive evidence that is sufficiently aligned with wearer behavior, task context, and ongoing interaction. Ego4D and EPIC-KITCHENS establish the diversity and interaction density of egocentric activity~\cite{Ego4D,EK-100}, while Project Aria and HoloAssist demonstrate the importance of synchronized sensing and interactive task structure~\cite{ProjectAria,Holoassist}. The objective is therefore not merely to recognize visible entities, but to continuously recover reliable and temporally valid evidence about what the wearer sees, attends to, touches, and changes. This requires treating glasses-mounted video as a distinct sensing regime, jointly interpreting interaction-relevant cues, accounting for the acquisition process, and evaluating perception under streaming and deployment conditions.

\noindent\textbf{Egocentric observation regime.}
The central challenge is not simply to transfer existing vision models to glasses-mounted cameras, but to address the distribution shift induced by camera placement, egocentric framing, and continuous wearer motion. Ego4D documents long-horizon daily activities across diverse environments~\cite{Ego4D}, while EPIC-KITCHENS-100 formalizes fine-grained action-object interactions under strong hand occlusion and head motion~\cite{EK-100}. HoloAssist further emphasizes interactive assistance during real procedures~\cite{Holoassist}, and Ego-1K expands the scale and multiview diversity of first-person video~\cite{Ego-1K}. EgoObjects provides large-scale first-person imagery with fine-grained object annotations~\cite{EgoObjects}, and EgoTracks introduces a benchmark for persistent object tracking in egocentric video, where rapid viewpoint changes, occlusion, and object reappearance are common~\cite{EgoTracks}. Always-on collection efforts such as AoE additionally expose the long-tail, storage, annotation, and privacy burdens of continuous capture~\cite{AoE}. These works collectively support treating first-person video as a distinct observation regime rather than merely a change in viewpoint.

\noindent\textbf{Interaction-centric compositional perception.}
Hands, objects, contact relations, object-state changes, gaze, and deictic gestures should not be modeled as independent detection targets because their joint configuration provides evidence about task progress, wearer attention, interaction intent, and actionable affordances. HoloAssist couples egocentric observations with interactive task assistance~\cite{Holoassist}, while EGTEA Gaze+ shows that gaze and action understanding are mutually informative in first-person video~\cite{sgacad2018_egtea_gaze}. Referential reasoning studies demonstrate that pointing must be interpreted jointly with the visual scene rather than as an isolated gesture~\cite{MLLMs-Pointing}, and Project Aria provides synchronized eye-tracking, head pose, and multimodal streams for studying such alignment~\cite{ProjectAria}. Emerging work on inferring grasp pressure from egocentric video further illustrates that visible hand-object configurations can support richer interaction-state estimation beyond category recognition~\cite{EgoTactile}. The relevant capability boundary is therefore whether attention, physical contact, object-state transitions, and intent can be represented coherently within a shared temporal stream.

\noindent\textbf{Temporally reliable evidence acquisition.}
Compositional interpretation depends on the reliability of the sensing process itself. Project Aria foregrounds calibration, synchronization, and timestamped multimodal capture as prerequisites for research-grade first-person analysis~\cite{ProjectAria}, while EgoKit examines unified acquisition across heterogeneous low-cost devices~\cite{EgoKit}. Always-on collection systems such as AoE and Open-AoE make continuity, duty cycle, data management, and toolchain reliability explicit parts of the acquisition problem~\cite{AoE,Open-AoE}. Device and wearer conditions also matter: OCR performance can change with walking speed, camera placement, and camera type~\cite{EvaluatingOCRPerformance}, and lightweight visual-inertial odometry illustrates the resource constraints under which temporal alignment must be maintained~\cite{LEVIO}. Evaluation should therefore report frame rate, exposure, camera placement, timestamp quality, sensor dropout, and lens occlusion, and should test robustness under motion, low light, interference, and sustained use rather than attributing all performance differences to model quality.

\noindent\textbf{Evaluation beyond isolated recognition.}
Conventional object-detection or action-recognition accuracy captures only part of first-person perception. WearVQA and SuperGlasses test visual reasoning under wearable and smart-glasses interaction conditions~\cite{WearVQA,Superglasses}, while GLIMPSE highlights the combined demands of real-time text recognition and contextual understanding~\cite{GLIMPSE}. EgoSAT moves evaluation from isolated frames to streaming interaction understanding~\cite{EgoSAT}, and SAW-Bench broadens the target toward situated awareness in real environments~\cite{SAW-Bench}. Assistive datasets such as VizWiz additionally reveal the effects of imperfect framing, blur, occlusion, and user-generated capture on downstream question answering~\cite{VizWiz}. A complete evaluation should therefore include viewpoint alignment, temporal localization, active-object recall, interaction relevance, streaming latency, privacy-aware sampling, user correction, and cross-scene generalization, with the central objective being the continuous production of reliable, actionable, and temporally valid evidence.

\subsection{Multimodal Context Modeling}
\label{sec:multimodal-context}

Reliable first-person observations alone do not constitute contextual understanding. Project Aria demonstrates the availability of synchronized visual, inertial, audio, gaze, and pose streams~\cite{ProjectAria}, HoloAssist organizes such observations around interactive procedures~\cite{Holoassist}, Epic-Sounds provides a large-scale egocentric audio dataset that annotates aligned audible actions and sound events~\cite{Epic-Sounds}, and EgoSAT evaluates reasoning over continuous interaction streams~\cite{EgoSAT}. EgoVLPv2 introduces a backbone-level video–language fusion strategy for egocentric pre-training~\cite{EgoVLP}, and ContextAgent introduces context-aware proactive LLM agents with open-world sensory perceptions~\cite{ContextAgent}. Smart glasses must therefore transform heterogeneous and continuously changing inputs into an explicitly updatable state that represents what is happening, what the user currently intends, how the situation is changing, and which evidence supports each conclusion. Multimodal context modeling spans cross-modal integration, temporal state updating, bidirectional interaction, and provenance-aware uncertainty management rather than independent processing or simple concatenation of modality-specific features.

\noindent\textbf{Heterogeneous evidence integration.}
Multimodal context may include video, speech, environmental sound, IMU signals, gaze, location, device state, historical events, and prior system feedback, each with different sampling rates and uncertainty. Project Aria provides a synchronized multimodal research substrate~\cite{ProjectAria}, whereas HoloAssist demonstrates how visual, speech, and task signals interact during real-world assistance~\cite{Holoassist}. WearVQA and SuperGlasses show that wearable question answering must connect visual evidence with user queries and device-constrained interaction~\cite{WearVQA,Superglasses}, and ~\cite{AgenticLongVideoUnderstanding} targets complex queries that require integrating evidence distributed across distant portions of a long video. Meeting and social-interaction corpora such as AMI and AVA Active Speaker further motivate speaker-aware fusion of audio and visual evidence~\cite{AMI,AvaActiveSpeaker}. These settings support an explicit contextual state that aligns heterogeneous signals while retaining modality-specific confidence, rather than a collection of independently encoded inputs.

\noindent\textbf{Temporal state updating and event reasoning.}
Context must evolve as actions unfold, objects change state, users revise goals, and new evidence invalidates earlier interpretations. EgoSAT directly targets streaming interaction understanding and state change~\cite{EgoSAT}, while HoloAssist structures long procedural interactions around task steps and assistance events~\cite{Holoassist}. Daily-life assistants and long-horizon memory benchmarks show that relevant evidence may be separated by long temporal gaps~\cite{Egolife,EgoMemReason}, and EGOSTREAM makes online episodic updating a diagnostic target~\cite{EgoStream}. Procedural datasets such as Assembly101 additionally show that activity understanding requires tracking ordered steps, objects, and pose over extended sequences~\cite{Assembly101}. A useful contextual state should therefore preserve event order, temporal validity, task phase, unresolved dependencies, and the evidence responsible for each update instead of reconstructing the situation from scratch at every query. Long-form context also requires selective access to evidence that may be distributed across an extended observation stream. AdaVideoRAG constructs complementary text, visual, and graph indexes from clip captions, ASR, OCR, and visual features, and adaptively selects retrieval schemes according to query complexity~\cite{xue2025adavideorag}. This design provides a useful mechanism for query-conditioned context construction in smart-glasses systems, although it should be interpreted as evidence for long-video retrieval rather than for persistent, user-correctable personal memory.

\noindent\textbf{Bidirectional interaction and output-conditioned reasoning.}
The same channels that provide context often deliver system feedback, making multimodal modeling inherently bidirectional. Work on active noise cancellation for open-ear smart glasses shows that acoustic processing affects both environmental awareness and output intelligibility~\cite{ActiveNoiseCancellation}, while studies of everyday smart-glasses conversations document interruptions, misunderstandings, and repair as core usability phenomena~\cite{Conversational}. AMI and AVA Active Speaker motivate participant and speaker tracking in multi-person interaction~\cite{AMI,AvaActiveSpeaker}, and QMSum illustrates the need to condense long conversational context around a user query~\cite{QMSum}. AR captioning studies for deaf students and mixed-vision social activities further show that display modality, placement, and social context shape whether information is accessible and disruptive~\cite{EvaluatingARForDeafStudents,ReshapingInclusiveInterpersonalDynamics}. Reasoning should therefore be conditioned not only on input evidence but also on output bandwidth, interruption cost, privacy, and the confirmation affordances of audio-only, HUD, or spatial-display interfaces.

\noindent\textbf{Provenance, conflict, and uncertainty management.}
A contextual state should retain current observations, derived inferences, retrieved memories, and external-tool outputs without collapsing their evidential status. Cross-view memory tasks show that synchronized ego and exo streams may provide complementary or conflicting evidence for the same event~\cite{EgoExoMem}, while H2HMem requires attribution across speakers, modalities, and social interactions~\cite{H2HMem}. OpenEQA and SAW-Bench expose the need to ground answers in embodied and situated evidence rather than language priors alone~\cite{Openeqa,SAW-Bench}. Continuous proactive benchmarks such as IPIBench make uncertainty consequential because uncertain state estimates can trigger or suppress interventions~\cite{IPIBench}, and personalized visual context learning shows that prior user-specific information can alter interpretation~\cite{PersonalVisualContextLearning}. These settings motivate explicit source attribution, temporal validity, modality-conflict detection, and uncertainty communication as both evaluation criteria and gates for admitting information into memory or authorizing action.

\subsection{Persistent Spatial State}
\label{sec:spatial-state}

Multimodal context characterizes the evidence available at a given moment, whereas persistent spatial state organizes transient observations into durable and continuously revisable relations among the wearer, objects, places, paths, actionable regions, and task-relevant states. Project Aria and Aria Digital Twin connect wearable sensing to calibrated three-dimensional reconstruction~\cite{ProjectAria,Ariadigitaltwin}, while Pandora illustrates the move from geometry toward object-centric scene structure~\cite{Pandora}. This capability therefore extends beyond one-shot pose estimation or the presence of an AR display: it requires a geometric backbone, object- and action-centric semantics, cross-session revision, and feedback mechanisms that expose spatial state safely and intelligibly during movement.

\noindent\textbf{Geometric localization and mapping backbone.}
Persistent spatial state begins with reliable estimation of wearer motion and scene geometry. VINS-Mono establishes a visual-inertial state-estimation baseline for monocular sensing~\cite{Vins-mono}, ORB-SLAM3 supports visual, visual-inertial, and multimap SLAM~\cite{Orb-slam3}, and DROID-SLAM extends learned optimization across monocular, stereo, and RGB-D settings~\cite{Droid-slam}. LEVIO shows how visual-inertial odometry must be redesigned for resource-constrained devices~\cite{LEVIO}, while Project Aria demonstrates the calibration and sensor synchronization available on a glasses-based research platform~\cite{ProjectAria}. For smart glasses, evaluation should therefore cover pose error, drift, relocalization, map consistency, uncertainty, resource consumption, and robustness to rapid head motion, partial views, and illumination changes rather than reconstruction accuracy alone.

\noindent\textbf{Object-centric and actionable spatial semantics.}
Geometry becomes useful for assistance only when it is connected to objects, relations, reachability, and task relevance. Aria Digital Twin provides egocentric three-dimensional machine-perception data in calibrated environments~\cite{Ariadigitaltwin}, and Pandora represents articulated scenes through object-centric three-dimensional scene graphs~\cite{Pandora}. SpatialWorld evaluates interactive spatial reasoning by multimodal agents~\cite{SpatialWorld}, while EgoProx organizes near-body objects, distance, and reachability into a first-person proximity hierarchy~\cite{EgoProx}. Matterport3D and ScanNet provide broader precedents for object- and region-level indoor three-dimensional semantics~\cite{Matterport3d,Scannet}. Together, these works motivate spatial state that answers not only where the wearer is, but also what is where, whether it is reachable or traversable, and how those relations constrain the next action.

\noindent\textbf{Persistence, staleness, and cross-session revision.}
Actionable spatial semantics remain reliable only when the state can be revised across time and sessions. ORB-SLAM3's multimap setting provides a foundation for revisiting and relocalizing across spatial episodes~\cite{Orb-slam3}, while Aria Digital Twin and Pandora provide structured scene representations that can be compared as objects and articulated states change~\cite{Ariadigitaltwin,Pandora}. Latent Spatial Memory explicitly treats persistent spatial information as a component of video world models~\cite{LatentSpatialMemory}, and EgoForge frames egocentric world simulation around goal-directed state evolution~\cite{EgoForge}. These directions motivate map-age estimation, object-persistence modeling, contradiction detection, and user correction. Each spatial assertion should preserve when it was established or updated, which sensor, model, or user supplied the evidence, and how confidence decays as the environment changes.

\noindent\textbf{Spatial feedback and mobility safety.}
Spatial state becomes operational only through feedback that the wearer can interpret safely while moving. NavCog demonstrates navigational assistance for blind users~\cite{NavCog}, and urban risk-aware navigation uses event maps to communicate hazards and route-relevant conditions~\cite{UrbanRiskAwareNavigation}. Vision-and-language navigation and REVERIE provide established tasks for instruction following and referential grounding in indoor spaces~\cite{VLN,Reverie}, while OpenEQA evaluates embodied question answering about real environments~\cite{Openeqa}. Assistive systems and datasets such as VizWiz and LidSonic further emphasize imperfect sensing and the need for concise, accessible feedback~\cite{VizWiz,LidSonic}. Audio cues, HUD arrows, text, maps, and natural-language explanations should therefore be evaluated jointly with the spatial representation through latency, referential clarity, uncertainty, correction affordances, visual occlusion, distraction, and mobility risk.

\subsection{Auditable Long-Term Personal Memory}
\label{sec:long-term-memory}

Persistent spatial state captures how environmental relations endure and evolve, whereas auditable long-term personal memory governs how personally relevant evidence is admitted, retained, updated, retrieved, corrected, and revoked over time. EgoLife establishes daily-life assistance as a long-horizon retrieval problem~\cite{Egolife}, EgoMonth provides month-scale egocentric video sequences for testing persistent spatiotemporal memory across extended daily-life experiences~\cite{EgoMonth}, EgoMemReason focuses on reasoning across temporally separated egocentric events~\cite{EgoMemReason}, Memoro~\cite{Memoro} employs LLMs to support efficient retrieval and contextual use of episodic information during daily activities, and LightMem-Ego develops a tiered memory system for continuous first-person streams~\cite{LightMem-Ego}. Progress toward a persistent personal assistant therefore depends not on indiscriminate storage, but on memory that is hierarchically organized, provenance-preserving, temporally valid, correctable, permission-aware, and effectively deletable.

\noindent\textbf{Hierarchical memory organization.}
The memory layer may distinguish working context, episodic memory, semantic memory, spatial memory, user-confirmed facts, and action logs, with each class assigned a different temporal horizon and authorization policy. EgoLife studies memory retrieval in daily-life settings~\cite{Egolife}, while EgoMemReason emphasizes reasoning across temporally separated events~\cite{EgoMemReason}. EGOSTREAM extends the problem to online episodic memory~\cite{EgoStream}, EgoExoMem introduces synchronized cross-view memory evidence~\cite{EgoExoMem}, and H2HMem requires memory to preserve speakers, events, and social context~\cite{H2HMem}. EgoTrigger proposes using informative audio events to trigger selective image capture for human memory enhancement~\cite{EgoTrigger}, and LightMem-Ego explicitly routes queries across current, short-term, and long-term tiers according to temporal scope and intent~\cite{LightMem-Ego}. Together, these works motivate functional memory tiers rather than a uniform repository of all observations.

\noindent\textbf{Provenance-aware admission and evidential status.}
Auditability begins when information is written, not only when it is retrieved. EgoExoMem shows that the same event may be supported differently by ego and exo viewpoints~\cite{EgoExoMem}, and H2HMem introduces multiple people, modalities, and social roles whose contributions must remain distinguishable~\cite{H2HMem}. Personalized visual context learning demonstrates that user-specific information can alter later interpretation~\cite{PersonalVisualContextLearning}, while LightMem-Ego illustrates the operational need to decide which observations enter which memory tier~\cite{LightMem-Ego}. Privacy analyses of life-logging and wearer-bystander tensions show that persistence itself carries permission and governance consequences~\cite{Position,MindTheGap}. A memory item should therefore retain source, timestamp, spatial context, confidence, confirmation status, access permissions, and expiration policy, and admission rules should distinguish direct observations, model inferences, external retrieval, and user-confirmed facts.

\noindent\textbf{Spatiotemporal retrieval and validity.}
Useful recall recover not only semantic content but also when, where, and under what evidence an assertion was established. EgoMemReason evaluates reasoning across distant moments~\cite{EgoMemReason}, and EGOSTREAM tests whether episodic evidence can be maintained and queried as a stream evolves~\cite{EgoStream}. EgoLife places retrieval in everyday personal-assistance scenarios~\cite{Egolife}, while EgoTracks is intended to evaluate long-term identity preservation and temporally robust visual association~\cite{EgoTracks}. LightMem-Ego further aligns audiovisual evidence along a shared timeline and routes queries by temporal scope~\cite{LightMem-Ego}. Latent Spatial Memory shows that spatial persistence can be embedded within long-horizon video models~\cite{LatentSpatialMemory}, and OpenEQA illustrates the need to answer questions from situated environmental evidence~\cite{Openeqa}. Long-term retrieval should therefore jointly recover semantic, spatial, temporal, and provenance information and explicitly determine whether the recalled state remains valid at query time.

\noindent\textbf{Correction, revocation, and effective forgetting.}
A user-controllable memory must support provenance inspection, correction, person- or place-specific deletion, permission changes, and verification that those changes propagate to later behavior. Life-logging privacy analyses make deletion and purpose limitation central to the privacy-utility trade-off~\cite{Position}, while wearer-bystander studies show that consent and revocation may depend on social context rather than the wearer alone~\cite{MindTheGap}. VisGuardian explores lightweight privacy control for front-camera data from AR glasses~\cite{VisGuardian}, and UNSEEN explicitly investigates unlearning as a defense in AR-LLM systems~\cite{UNSEEN}. Personalized context learning further raises the possibility that user-specific information is absorbed into model behavior rather than retained only as an explicit record~\cite{PersonalVisualContextLearning}. Effective forgetting should therefore cover stored records, indexes, summaries, caches, and downstream policies, and should be verified through future retrieval and action rather than inferred from deleting one database entry.

\noindent\textbf{Evaluation of fidelity, validity, and deployability.}
Memory evaluation should test whether information is relevant, supported, still valid, controllable by the user, and available within the device budget. Personal Visual Context Learning provides evidence that individualized context can improve visual understanding~\cite{PersonalVisualContextLearning}, whereas online episodic QA on the edge exposes latency and resource constraints~\cite{MultimodalLMMs-EpisodicMemory}. EGOSTREAM and EgoMemReason motivate temporal localization, long-horizon reasoning, and false-recall analysis~\cite{EgoStream,EgoMemReason}; LightMem-Ego motivates tier-aware retrieval and system-level latency measurements~\cite{LightMem-Ego}; and life-logging privacy work motivates leakage and deletion tests~\cite{Position}. Evaluation should consequently include retrieval accuracy, source-attribution accuracy, answer-validity window, false-recall rate, correction persistence, privacy leakage, deletion effectiveness, latency, and resource consumption. Together, these criteria define the minimum evidential requirements for an L3 claim in \cref{sec:capability-levels}.

\subsection{Situated Agentic Action}
\label{sec:situated-agent}

Context and memory can provide evidence for an action, but they do not by themselves confer authority to execute it. VisionClaw frames smart glasses as an always-on agent interface~\cite{VisionClaw}, Egocentric Co-Pilot connects first-person perception to assistive web-native actions~\cite{EgocentricCoPilot}, and Pro2Assist studies continuous step-aware procedural assistance~\cite{Pro2Assist}. Situated agentic action therefore refers to a closed-loop capability that grounds goals in the wearer's current situation, determines whether and when intervention is appropriate, executes only authorized operations, monitors outcomes, and supports correction or recovery. The analytical focus shifts from answer generation to the reliability and governance of the complete observation-state-action loop.

\noindent\textbf{Closed-loop action architecture.}
A situated agent spans observation, grounding, inference, planning, action, feedback, and correction rather than merely attaching a conversational model to eyewear. VisionClaw studies always-on agents through smart glasses~\cite{VisionClaw}, and agentic long-video understanding provides mechanisms for selectively inspecting and reasoning over extended visual evidence~\cite{AgenticVideoUnderstanding}. Ego2Web grounds web-agent tasks in egocentric videos~\cite{Ego2Web}, while Mobile-Agent and SeeAct demonstrate visually grounded action in mobile and web interfaces~\cite{MobileAgent,Seeact}. WebArena provides a realistic environment in which action sequences have persistent external consequences~\cite{Webarena}. These works motivate evaluation of whether the agent has sufficient evidence to act, grounds operations to the current state, executes them reliably, and incorporates outcomes into subsequent state estimates.

\noindent\textbf{In-situ grounding, task phase, and personalization.}
The distinctive value of smart glasses lies in continuous access to the wearer's audiovisual, spatial, and task context rather than in transferring a phone assistant into a different form factor. Egocentric Co-Pilot explores assistive agents grounded in first-person experience~\cite{EgocentricCoPilot}, and Pro2Assist centers assistance on the current procedural step~\cite{Pro2Assist}. ~\cite{PlanWatchRecover} explicitly organizes proactive assistance around planning, observation, and recovery, while Ego-Pro-Bench evaluates personalized proactive interaction in continuous streams~\cite{EgoPro-Bench}. HoloAssist provides interactive procedural data for recognizing task progress and intervention needs~\cite{Holoassist}, and ~\cite{PersonalVisualContextLearning} shows how individualized context can change interpretation. ~\cite{Ego-Grounding} studies personalized question answering over egocentric video by grounding user-specific queries in the relevant temporal segments, objects, and activities captured from the wearer’s perspective. Situated assistance should therefore jointly model task phase, nearby entities, user preferences, prior corrections, and current uncertainty.

\noindent\textbf{Proactive intervention and timing.}
As the system moves from reactive assistance toward proactive intervention, it must balance missed opportunities against unnecessary, premature, or mistimed interruptions. Streaming Interventions evaluates whether video models can correct mistakes as they occur~\cite{StreamingInterventions}, and IPIBench places interactive proactive intelligence under continuous-stream conditions~\cite{IPIBench}. Pro2Assist~\cite{Pro2Assist} and ~\cite{PlanWatchRecover} both make step awareness and recovery central to procedural support, while Ego-Pro-Bench emphasizes personalization in the intervention policy~\cite{EgoPro-Bench}. AI for Service investigates AI glasses as a proactive service interface that continuously interprets the wearer’s context and delivers timely, task-relevant assistance without requiring explicit queries~\cite{AI4Service}. EgoSAT further links intent and interaction timing in streaming egocentric understanding~\cite{EgoSAT}. A practical policy should determine when to abstain, prompt, clarify, confirm, or act, with thresholds calibrated to uncertainty, user burden, task phase, and the consequences of failure.

\noindent\textbf{Authority, permission, and consequence-aware execution.}
Action authority should scale with the reversibility and external consequences of execution. WebArena, SeeAct, and Mobile-Agent demonstrate that visually grounded agents can modify persistent digital environments~\cite{Webarena,Seeact,MobileAgent}, while SayCan and RT-1 connect language-conditioned reasoning to physically executable robot actions~\cite{SayCan,RT-1}. AI-glasses communication research further shows that intent and system coordination become part of the action interface~\cite{IntentionAwareSemanticAgentCommunications}. At the same time, visual jailbreaks and real-time AR-LLM social-engineering attacks expose new attack surfaces when perceived content can influence model behavior~\cite{VisualAdversarialExamplesJailbreak,PhySE}. Increasing authority should therefore require stronger permission gates, explicit confirmation, least-privilege access, revocation, provenance, and audit logging, clearly separating the model's ability to propose an action from the system's authorization to execute it.

\noindent\textbf{Monitoring, recovery, and stage-wise evaluation.}
Execution should be monitored against expected outcomes, and deviations should trigger clarification, rollback, or recovery. Plan-Watch-Recover directly elevates recovery to a first-class component of proactive assistance~\cite{PlanWatchRecover}, while Streaming Interventions and IPIBench test online correction and interactive responses under continuous evidence~\cite{StreamingInterventions,IPIBench}. WebArena makes action outcomes observable in a persistent environment~\cite{Webarena}, and robot-control work such as RT-1 and SayCan demonstrates the importance of connecting planned actions to realizable outcomes~\cite{RT-1,SayCan}. Evaluation should jointly consider task success, grounding accuracy, unsafe-action rate, intervention frequency, interruption cost, user override, rollback or recovery success, and post-action state consistency, with failures attributed to observation, inference, authorization, execution, feedback, or recovery rather than only to the final outcome.

\subsection{Embodied Data Interfaces}
\label{sec:embodied-interface}

Smart glasses can also function as research interfaces that transform first-person observations and internal state into structured data for downstream embodied learning. Ego-Exo4D demonstrates the value of synchronized first- and third-person views~\cite{Ego-Exo4D}, Project Aria provides calibrated multimodal wearable sensing~\cite{ProjectAria}, and EgoZero and Open X-Embodiment illustrate the downstream objective of connecting human experience to robot learning across platforms~\cite{Egozero,Open-x-embodiment}. The distinctive value of this capability lies in capturing human experience from a naturally worn viewpoint and preserving the multimodal, spatial, behavioral, and task-relevant structure needed for later alignment across views and embodiments. It should be analyzed through acquisition, state recovery, cross-embodiment alignment, downstream validation, and data governance rather than equated directly with robot-policy execution.

\noindent\textbf{Synchronized capture of human experience.}
Glasses can jointly record video, gaze, speech, hand-object interaction, head motion, spatial context, and task outcomes from a first-person viewpoint. Ego-Exo4D aligns first-person recordings with multiple external views~\cite{Ego-Exo4D}, while Project Aria and Aria Digital Twin provide calibrated multimodal sensing and three-dimensional reference environments~\cite{ProjectAria,Ariadigitaltwin}. EgoExoMoCap further demonstrates a distributed capture paradigm in which two or more people wearing smart glasses provide mutually complementary ego- and exocentric observations for full-body human motion estimation in the global 3D world~\cite{EgoExoMoCap}. AoE and Open-AoE address always-on collection and an open toolchain for egocentric manipulation data~\cite{AoE,Open-AoE}, and EgoKit targets unified acquisition across heterogeneous devices~\cite{EgoKit}. ActiveGlasses and EgoMI further show how active viewpoint control and whole-body behavior can enrich demonstrations for manipulation learning~\cite{Activeglasses,EgoMi}. These systems position smart glasses as interfaces for producing synchronized records of human experience rather than direct proxies for robot policies.

\noindent\textbf{Task-state recovery and structured representation.}
An irreducible transformation pipeline separates raw recordings from reusable embodied-learning data. HoloAssist provides interactive procedural structure in egocentric assistance settings~\cite{Holoassist}, while Assembly101 and IKEA~ASM annotate actions, objects, and pose in complex assembly activities~\cite{Assembly101,IKEA-ASM}. COIN and CrossTask establish large-scale and weakly supervised formulations for recovering instructional steps from video~\cite{Coin,CrossTaskWeaklySupervisedLearning}. More recent egocentric manipulation work targets latent physical variables and robot-compatible demonstrations, including grasp pressure in EgoTactile~\cite{EgoTactile}, high-fidelity dexterous demonstrations in EgoEngine~\cite{EgoEngine}, and large-scale dexterous manipulation in EgoDex~\cite{Egodex}. The interface is therefore valuable only when temporal segments, objects, contacts, subgoals, outcomes, and uncertainty can be recovered reproducibly and exposed in a representation consumable by downstream learning systems.

\noindent\textbf{Ego-exo and cross-embodiment alignment.}
Human demonstrations must be related across viewpoints and mapped to embodiments with different morphology, kinematics, dynamics, sensing, and action spaces. Ego-Exo4D provides synchronized cross-view supervision~\cite{Ego-Exo4D}, while EgoMimic and EgoZero study imitation and robot learning from egocentric human video~\cite{Egomimic,Egozero}. HumanEgo focuses on entity-level hand-object representations~\cite{Humanego}, and EgoVLA learns vision-language-action models from egocentric video~\cite{EgoVLA}. EgoEngine and UniDex further transform human observations toward dexterous robot demonstrations and control~\cite{EgoEngine,UniDex}, while ActiveMimic incorporates active perception into egocentric pretraining~\cite{ActiveMimic}. Cross-embodiment transfer therefore requires viewpoint normalization, correspondence estimation, embodiment-aware state abstraction, and mappings from human actions or subgoals to robot-compatible representations.

\noindent\textbf{Downstream validation and evidential boundaries.}
A human head-mounted viewpoint is not equivalent to a robot-mounted camera, and first-person audiovisual recordings typically omit force, tactile, joint-state, and robot-side proprioceptive signals needed for closed-loop control. Open X-Embodiment, DROID, and BridgeData V2 establish robot-side diversity and evaluation substrates that differ fundamentally from passive human video~\cite{Open-x-embodiment,Droid,BridgeData}. RT-1 and SayCan demonstrate that successful physical execution requires robot observations, affordances, and action interfaces~\cite{RT-1,SayCan}, while R3M provides a robot-oriented visual representation learned for manipulation~\cite{R3M}. EgoVLA and HumanEgo represent attempts to bridge human egocentric evidence toward robot policies~\cite{EgoVLA,Humanego}. A defensible upstream claim is therefore that smart glasses reduce capture cost and improve representation learning or demonstration coverage; claims of policy transfer, reliable manipulation, or safe execution require downstream tests of transfer performance, sample efficiency, failure modes, and safety on the target embodiment.

\noindent\textbf{Consent, ownership, and downstream responsibility.}
Embodied data collection may capture bystanders, sensitive environments, proprietary workflows, and worker expertise. The privacy-utility trade-off in life-logging streams~\cite{Position}, context-dependent wearer-bystander tensions~\cite{MindTheGap}, and lightweight front-camera privacy control~\cite{VisGuardian} all show that governance must begin at acquisition. Earlier studies of bystander perspectives, opt-in and opt-out gestures, and in-the-wild social acceptability establish that camera-glasses consent is a situated social process~\cite{sgacad2014_bystanderprivacy,CameraGlassesPrivacy,sgacad2019_socialacceptability}. AoE and Open-AoE further make always-on collection and dataset toolchains part of the embodied-learning pipeline~\cite{AoE,Open-AoE}. An embodied data interface should therefore specify consent coverage, de-identification, provenance, ownership of demonstrated skills, permissible downstream use, data-use restrictions, and responsibility for robot-side validation as distinct from transfer performance.

\subsection{Cross-Cutting Deployment Constraints}
\label{sec:deployable-constraints}

Deployment constraints do not form a capability that follows perception, memory, or action in sequence. They define the cross-cutting feasibility envelope within which every preceding capability must remain usable, reliable, safe, and sustainable in an eyewear form factor. EPIC and OpenGlass show that system architecture and device partitioning shape real-time visual assistance~\cite{EPIC,OpenGlass}, while privacy and conversational studies demonstrate that social acceptability and interaction breakdowns can invalidate technically correct behavior~\cite{Position,Conversational}. The effective capability of a system is therefore jointly determined by computational scheduling, physical operating conditions, feedback latency and usability, privacy and security controls, and reproducibility under software, service, and environmental drift.

\noindent\textbf{Computation, energy, and thermal scheduling.}
Streaming inference, adaptive sampling, event-triggered perception, cascaded models, on-device filtering, and edge-cloud scheduling jointly determine observation quality, latency, energy consumption, and sustainable duty cycle. EPIC studies efficient egocentric perception on embodied AR glasses~\cite{EPIC}, and LEVIO targets visual-inertial odometry on resource-constrained devices~\cite{LEVIO}. At the model level, compact attention-based backbones such as EMO and EMOv2 provide transferable methods for balancing parameter count, computation, and visual recognition or dense-prediction performance, while token-adaptive knowledge distillation offers a complementary route for compressing language reasoning modules~\cite{zhang2023rethinking,zhang2025emov2,xie2026llm}. These works provide component-level efficiency evidence; their value for smart glasses still requires direct profiling of end-to-end latency, energy consumption, memory use, and thermal stability on the target device. ~\cite{OpenGlass} explores a sensing-computing split for local MLLM-driven assistance, ~\cite{OpenGlassUltraLowPower} presents an ultra-low-power AI-eyewear platform that uses event-based vision and on-device processing for continuous egocentric perception, EgoTrigger reduces the energy and storage costs of continuous visual recording through selective audio-driven image capture~\cite{EgoTrigger}, and EgoKit examines low-cost heterogeneous capture platforms~\cite{EgoKit}. Online episodic QA on the edge exposes memory and inference budgets~\cite{MultimodalLMMs-EpisodicMemory}, GLIMPSE emphasizes real-time text recognition and contextual VQA~\cite{GLIMPSE}, and active noise cancellation for open-ear glasses adds continuous audio processing to the same power envelope~\cite{ActiveNoiseCancellation}. These factors should be treated as part of the effective capability boundary rather than as implementation details considered after model evaluation.

\noindent\textbf{Wearable operating envelope and human factors.}
Battery state, thermal behavior, weight, camera placement, display brightness, audio leakage, network dependence, and firmware configuration can place the same model under substantially different runtime conditions. OCR studies show that walking speed, camera placement, and camera type materially affect assistive recognition performance~\cite{EvaluatingOCRPerformance}. Open-ear noise cancellation and everyday conversation studies expose the trade-off between intelligibility, environmental awareness, leakage, and interruption~\cite{ActiveNoiseCancellation,Conversational}. In-the-wild work on wearable-camera social acceptability and wearer-bystander tensions demonstrates that an operationally available sensor may still be socially unusable~\cite{sgacad2019_socialacceptability,MindTheGap}. Manufacturing reviews additionally show that head-mounted AR effectiveness depends on comfort, ergonomics, and integration with real workflows~\cite{ARinManufacturing}. Evaluation should therefore report device, firmware, battery, thermal state, ambient conditions, sensor duty cycle, and the wearability conditions under which the claimed capability remains available.

\noindent\textbf{Latency and feedback usability.}
A semantically correct response may still fail if it arrives after its useful time window, contains excessive detail, is delivered at an inappropriate volume, or obstructs the wearer's field of view. GLIMPSE and OpenGlass make real-time response a system-design objective~\cite{GLIMPSE,OpenGlass}, while conversational studies show that delayed or poorly timed responses create breakdowns and repair costs~\cite{Conversational}. AR support for deaf students demonstrates the importance of caption placement and communication access~\cite{EvaluatingARForDeafStudents}, and urban risk-aware navigation shows that feedback timing is safety-critical during movement~\cite{UrbanRiskAwareNavigation}. SUPERGLASSES and WearVQA further motivate evaluation under wearable interaction constraints rather than offline answer quality alone~\cite{Superglasses,WearVQA}. Feedback should therefore be measured through time to first useful feedback, interruptibility, confirmation cost, referential clarity, audio leakage, display occlusion, and mobility risk.

\noindent\textbf{Privacy, security, and lifecycle governance.}
Privacy-preserving operation must span acquisition, filtering, transmission, storage, retrieval, action, and revocation. Life-logging privacy work argues that utility and privacy are inseparable at the stream level~\cite{Position}, while Mind the Gap and earlier bystander studies show that consent expectations vary with context and social relationship~\cite{MindTheGap,sgacad2014_bystanderprivacy,CameraGlassesPrivacy}. VisGuardian explores local group-based control over front-camera data~\cite{VisGuardian}. Security threats also propagate across the stack: visual adversarial examples can jailbreak aligned multimodal models~\cite{VisualAdversarialExamplesJailbreak}, real-time AR-LLM systems can be exploited for social engineering~\cite{PhySE}, and UNSEEN studies unlearning-based defenses~\cite{UNSEEN}. Verifiable controls should therefore include bystander redaction, sensitive-audio filtering, local-first processing, consent logs, access control, adversarial robustness, audit trails, and effective deletion rather than relying only on final-output filtering or policy text.

\noindent\textbf{Drift, service dependence, and reproducibility.}
Effective capability boundaries are sensitive to firmware updates, model or API changes, deployment region, network configuration, subscription status, service availability, and environmental composition. OpenGlass emphasizes reproducible open prototypes and explicit computing partitions~\cite{OpenGlass}, while EgoKit and Open-AoE make device heterogeneity and toolchain specification visible in data collection~\cite{EgoKit,Open-AoE}. Project Aria demonstrates the importance of calibration and versioned sensing configurations~\cite{ProjectAria}, and EPIC and LEVIO show that implementation choices change latency and resource use on constrained hardware~\cite{EPIC,LEVIO}. A versioned deployment record should therefore capture device and firmware, model or API version, region, network, subscription and service status, sampling duty cycle, battery and thermal conditions, user population, scene composition, and representative failures. Evaluation should distinguish average performance from tail failures caused by thermal throttling, weak connectivity, low battery, regional differences, or changing software dependencies. These records provide the reproducibility basis for the standardized evaluation protocol introduced in \cref{sec:design}.

\begin{table}[t]
\centering
\caption{\textbf{L0-L5 cross-capability level framework} that distinguishes recording and relay, reactive perception, contextual assistance, persistent state, governed action, and embodied coupling. Each level defines a verifiable capability boundary, the minimum conditions required to support that boundary, and the primary targets through which the corresponding claim should be evaluated.}
\label{tab:levels}
\small
\begin{adjustbox}{width=\linewidth}
\begin{tabular}{
>{\raggedright\arraybackslash}m{10mm}
>{\raggedright\arraybackslash}m{22mm}
>{\raggedright\arraybackslash}m{70mm}
>{\raggedright\arraybackslash}m{70mm}
>{\raggedright\arraybackslash}m{50mm}
}
\toprule
\textbf{Level} & \textbf{Functional Role} & \textbf{Verifiable Capability Boundary} & \textbf{Minimum Supporting Conditions} & \textbf{Core Evaluation Targets} \\
\midrule

\textbf{L0}
& Recording and Relay
& Captures, records, or livestreams observations; relays notifications; or delivers basic audio/visual cues without establishing a task-level semantic assistance loop
& At least one sensing or feedback channel, such as a camera, microphone, HUD, or speaker
& Capture quality, operational continuity, battery endurance, recording-state visibility \\

\midrule
\textbf{L1}
& Reactive Perception
& Recognizes text, objects, speech, acoustic events, hazards, or simple actions and produces responses grounded primarily in current observations
& Relevant sensing and feedback channels together with OCR, ASR, detection, tracking, or equivalent reactive perception modules
& Recognition accuracy, false-positive/negative rate, streaming latency, energy consumption, robustness \\

\midrule
\textbf{L2}
& Contextual Assistance
& Answers questions, translates, captions, performs visual search, or provides procedural assistance by integrating current and short-term multimodal context
& Streaming multimodal input processing, contextual reasoning, grounding mechanisms, and an appropriate feedback channel
& Task correctness, faithfulness, grounding accuracy, source attribution, temporal consistency, time to first useful feedback \\

\midrule
\textbf{L3}
& Persistent State
& Maintains, retrieves, updates, corrects, and revokes traceable episodic, semantic, or spatial state across events and sessions
& Sustained state capture, persistent storage, provenance metadata, privacy controls, retrieval mechanisms, and user-facing correction and deletion interfaces
& Retrieval accuracy, false-recall rate, answer-validity window, source-attribution accuracy, correction persistence, deletion effectiveness \\

\midrule
\textbf{L4}
& Governed Action
& Performs goal tracking, planning, tool invocation, and proactive assistance through explicit authorization, monitored execution, revocation, and recovery
& Reliable feedback, permission and confirmation interfaces, tool or action executors, least-privilege access, safety monitoring, recovery mechanisms, and audit logging
& Task success, unsafe-action rate, intervention cost, user override, rollback/recovery success, action-provenance completeness \\

\midrule
\textbf{L5}
& Embodied Coupling
& Transforms first-person human experience into transferable embodied-learning data or shared physical-task state and, for system-level claims, demonstrates utility on a downstream embodied system; data-oriented instances are qualified as \textit{L5 data} or \textit{partial L5}
& Synchronized multimodal sensing, gaze and pose information, calibration, raw or structured data access, cross-view or cross-embodiment alignment, and downstream embodied validation for system-level claims
& Alignment quality, downstream transfer performance, sample efficiency, cross-embodiment failure, execution safety, reproducibility \\

\bottomrule
\end{tabular}
\end{adjustbox}
\end{table}

\subsection{Cross-Capability Level Framework}
\label{sec:capability-levels}

The preceding analysis identifies the mechanisms that constitute a smart-glasses system and the dependencies among them. We consolidate these mechanisms into the \textbf{L0-L5 framework} (L0–L4 form a wearer-facing progression, whereas L5 is an orthogonal cross-embodiment extension), in which each level denotes the principal capability regime that can be substantiated for a specified task, hardware profile, operating condition, and body of evidence. 
A level is therefore not an aggregate product score derived from sensor count, display specifications, or brand positioning, nor is it intended as a consumer recommendation. A meaningful level claim should specify the task scope, input and output channels, temporal horizon, state persistence, action authority, deployment conditions, and supporting evidence. Where only part of a regime is publicly substantiated, qualifiers such as \textit{L3 potential}, \textit{L5 data}, or \textit{partial L5} should be used to make the evidential boundary explicit.

\noindent\textbf{Capability progression.}
As summarized in \cref{tab:levels}, the framework distinguishes six capability regimes. L0 records or relays information; L1 performs reactive semantic perception over current observations; L2 provides assistance grounded in current or short-term multimodal context; L3 maintains persistent, traceable, and correctable state across events; L4 executes externally consequential actions through governed permission, monitoring, and recovery mechanisms; and L5 connects first-person human experience or shared physical-task state to downstream embodied learning and cross-embodiment transfer. From L0 to L4, the framework progressively expands the temporal horizon, persistence of internal state, and consequences of system intervention. L5 instead extends the framework across the embodiment boundary and should not be interpreted simply as a higher value on the same wearer-facing axis.

\begin{itemize}
    \item \textbf{L0: Capture and relay.} L0 provides the baseline that separates information capture and relay from semantic perception. It includes recording, livestreaming, notification relay, and basic audio or visual cueing without requiring a real-time semantic understanding loop. Evidence at this level therefore centers on sensing and capture quality, operational continuity, battery endurance, and the observability of recording and device state.
    
    \item \textbf{L1-L2: Reactive and contextual assistance.} The transition from L0 to L1 introduces semantic processing of current observations, while L2 incorporates short-term multimodal context into an assistance loop. Even Realities G1~\cite{sgprod2026_even_realities_even_realities_g1}, Halliday DigiWindow~\cite{sgprod2026_halliday_halliday_digiwindow_glasses}, and Vuzix Z100~\cite{sgprod2026_vuzix_vuzix_z100} illustrate an L1-oriented route centered on low-power HUD feedback, captions, prompts, and teleprompter-style assistance. These low-bandwidth interfaces can provide clear utility, but the absence of a first-person camera or accessible egocentric visual stream limits claims of first-person visual understanding. L2-oriented profiles span both camera/audio-first and camera+HUD designs, including Ray-Ban Meta Gen 2~\cite{sgprod2026_meta_ray_ban_ray_ban_meta_gen_2}, Oakley Meta Vanguard~\cite{sgprod2026_meta_oakley_oakley_meta_vanguard}, Xiaomi AI Glasses~\cite{sgprod2026_xiaomi_xiaomi_ai_glasses}, Rokid AI Glasses Style~\cite{sgprod2026_rokid_rokid_ai_glasses_style}, Solos AirGo V2~\cite{sgprod2026_solos_solos_airgo_v2}, and Alibaba Quark S1~\cite{sgprod2026_alibaba_quark_alibaba_quark_ai_glasses_s1}. Their sensing, audio or display feedback, and phone- or cloud-assisted services provide the basis for audiovisual question answering, translation, meeting assistance, exercise feedback, service invocation, and lightweight recording. Such profiles support an L2 claim only when timely multimodal assistance is demonstrated under the relevant device, network, and service conditions.
    
    \item \textbf{L3-L4: Persistent state and governed action.} The distinction between L3 and L4 is determined less by display complexity than by whether the closed loop extends from persistent state to governed external action. Meta Ray-Ban Display~\cite{sgprod2026_meta_ray_ban_meta_ray_ban_display} and RayNeo X3 Pro~\cite{sgprod2026_tcl_rayneo_rayneo_x3_pro} combine first-person sensing, near-eye feedback, and additional interaction mechanisms that can reduce the cost of confirmation, clarification, and correction, thereby providing prerequisites for L3-oriented systems. A mature L3 claim, however, still requires evidence of state provenance, correction, deletion, temporal validity, and longitudinal stability. Snap Specs 2026~\cite{sgprod2026_snap_snap_specs_2026} and XREAL AURA~\cite{sgprod2026_xreal_google_xreal_aura_project_aura} similarly provide several prerequisites for L4-oriented research, including spatial displays, hand or gesture interaction, spatial sensing, and developer-facing interfaces. Establishing a deployable L4 system additionally requires sustained service availability, battery and thermal stability, accessible permission and revocation mechanisms, least-privilege tool execution, outcome monitoring, recovery or rollback, and robust field performance.
    
    \item \textbf{L5: Embodied data and cross-embodiment transfer.} Research sensing platforms occupy a distinct evidential role because they prioritize measurement fidelity, synchronization, calibration, and data accessibility rather than wearer-facing autonomous assistance. Project Aria Gen 1~\cite{ProjectAria}, Project Aria Gen 2~\cite{sgprod2026_meta_reality_labs_research_aria_gen_2}, and Pupil Labs Neon~\cite{sgprod2026_pupil_labs_pupil_labs_neon} provide foundations for embodied dataset construction, HCI and gaze analysis, ego-exo alignment, and human-demonstration capture through synchronized sensing, gaze and pose estimation, timestamps, calibration, and raw-data access. These capabilities can substantiate \textit{L5 data} or \textit{partial L5} claims. An \textit{L5 system} claim, by contrast, requires downstream validation showing that the captured or aligned human experience improves learning, transfer, or execution on a target embodied system.
\end{itemize}

\noindent\textbf{Evidence thresholds and qualifiers.}
A level should be assigned only to the capability demonstrated under the stated conditions. An L2 claim requires sufficiently timely and context-grounded assistance, whereas L3 additionally requires persistent-state provenance, user correction and deletion, temporal validity, and stability across sessions. L4 further requires permission gates, least-privilege access, revocation or rollback, outcome monitoring, and failure recovery. L5 requires separate qualification: \textit{L5 system} denotes embodied coupling whose benefit has been validated on a target physical system, while \textit{L5 data} or \textit{partial L5} denotes support for embodied-data research through synchronized sensing, gaze and pose estimation, calibration, multimodal alignment, or raw-data access. The latter does not imply an L4-grade wearer-facing agentic loop. Consequently, a single device may occupy different levels for different tasks. 

\noindent\textbf{Category-aware product mapping.}
The comparative value of the framework emerges when these capability regimes are mapped to public product evidence. Each mapping should combine a level, an evidential qualifier, and the evidence supporting that claim. Because enterprise devices, consumer products, developer platforms, and research systems optimize for different objectives and operating conditions, comparisons should be made primarily within comparable device categories. Cross-category contrasts are useful for clarifying capability boundaries and architectural trade-offs, but not for deriving a unified product ranking. 
Accordingly, \cref{tab:product-hardware} presents category-aware capability profiles rather than a global product ranking. L1 HUD-oriented devices are primarily differentiated by display legibility, low-power feedback, and privacy characteristics; L2 camera/audio-first and camera+HUD devices by wearability, sensing quality, service integration, latency, and interface accessibility; L3- and L4-oriented systems by persistent-state support, confirmation and correction mechanisms, permission interfaces, governed tool execution, recovery, and longitudinal stability; and L5 data-oriented research platforms by sensor synchronization, gaze and pose quality, calibration fidelity, raw-data access, SDK support, and data reproducibility. The comparison therefore clarifies research claims, system trade-offs, and evidential boundaries without reducing heterogeneous products to a single maturity score.

%% file: sec/04_applications.tex
\section{Application Scenes for Smart Glasses}
\label{sec:applications}

Building on the first-person data-flow formulation, device/platform capability axes, and product profiles established in \cref{sec:background}, together with the foundational capabilities and L0-L5 framework introduced in \cref{sec:capabilities}, this section shifts the analysis from isolated system functions to real-world application scenes. Specifically, we examine \textbf{\textit{how combinations of sensing, reasoning, memory, feedback, and action capabilities translate into deployable requirements under concrete user activities, failure consequences, and responsibility structures}}. As illustrated in \cref{fig:framework} \textit{(bottom left)}, the discussion is organized around nine application scenes and four recurring questions: what activity the wearer is performing, what value the glasses are expected to provide, which capability loop is required, and who bears responsibility when the system is wrong. The first six scenes cover recurring wearer-facing and institutional activities and are summarized at the scene level in \cref{tab:application-evidence}. The next three extend the loop through persistent world state, multi-party participation, and cross-embodiment transfer, where \cref{tab:late-application-benchmarks} therefore decomposes them into finer-grained task blocks. The final subsection briefly identifies additional scenes for which smart glasses provide plausible entry points but the current evidence remains too fragmented for equally detailed treatment.

\begin{table}[t]
\centering
\caption{\textbf{Capability requirements and expanded representative evidence across six recurring smart-glasses application scenes.}
The ``datasets/benchmarks" column distinguishes direct wearable resources from task-specific or neighboring-domain proxies, while the final column separates research systems and platforms from product-level entry points. 
}
\label{tab:application-evidence}
\scriptsize
\begin{adjustbox}{width=\linewidth}
\begin{tabular}{
>{\raggedright\arraybackslash}m{30mm}
>{\raggedright\arraybackslash}m{49mm}
>{\raggedright\arraybackslash}m{67mm}
>{\raggedright\arraybackslash}m{67mm}
}
\toprule
\textbf{Application Scene} &
\textbf{Key Capability Requirement} &
\textbf{Datasets/Benchmarks} &
\textbf{Systems, Platforms, or Product Entries} \\
\midrule
Daily Situated Assistance (\cref{sec:daily-assistant})
& streaming visual question answering, OCR and translation, personalized episodic memory, context retrieval, multimodal dialogue, and permission-governed digital tool use
& \textit{\tb{Wearable and situated understanding:}} SuperGlasses~\cite{Superglasses}; WearVQA~\cite{WearVQA}; SAW-Bench~\cite{SAW-Bench}; GLIMPSE~\cite{GLIMPSE}; EgoSAT~\cite{EgoSAT}; TextVQA~\cite{TowardsVQAModelsThatCanRead}. \textit{\tb{Personal context and memory:}} EgoLife~\cite{Egolife}; EgoMonth~\cite{EgoMonth}; TeleEgo~\cite{TeleEgo}; PVCL~\cite{PersonalVisualContextLearning}; EgoMemReason~\cite{EgoMemReason}; EGOSTREAM~\cite{EgoStream}; EgoExoMem~\cite{EgoExoMem}; LightMem-Ego~\cite{LightMem-Ego}; online episodic-memory QA~\cite{MultimodalLMMs-EpisodicMemory}. \textit{\tb{Agentic-action proxies:}} agentic long-video understanding~\cite{AgenticVideoUnderstanding}; Ego2Web~\cite{Ego2Web}; WebArena~\cite{Webarena}; SeeAct~\cite{Seeact}; Mobile-Agent~\cite{MobileAgent}
& \textit{\tb{Research systems and platforms:}} VisionClaw~\cite{VisionClaw}; Egocentric Co-Pilot~\cite{EgocentricCoPilot}; OpenGlass~\cite{OpenGlass}; EPIC~\cite{EPIC}; Project Aria~\cite{ProjectAria}; Aria Gen 2~\cite{sgprod2026_meta_reality_labs_research_aria_gen_2}. \textit{\tb{Consumer and display entries:}} Ray-Ban Meta Gen 2~\cite{sgprod2026_meta_ray_ban_ray_ban_meta_gen_2}; Meta Ray-Ban Display~\cite{sgprod2026_meta_ray_ban_meta_ray_ban_display}; Alibaba Quark AI Glasses S1~\cite{sgprod2026_alibaba_quark_alibaba_quark_ai_glasses_s1}; Rokid AI Glasses Style~\cite{sgprod2026_rokid_rokid_ai_glasses_style}; Xiaomi AI Glasses~\cite{sgprod2026_xiaomi_xiaomi_ai_glasses}; Solos AirGo V2~\cite{sgprod2026_solos_solos_airgo_v2}; Even Realities G1~\cite{sgprod2026_even_realities_even_realities_g1}; Halliday DigiWindow Glasses~\cite{sgprod2026_halliday_halliday_digiwindow_glasses}; Vuzix Z100~\cite{sgprod2026_vuzix_vuzix_z100} \\
\midrule
Accessibility Assistance (\cref{sec:accessibility})
& reliable scene description and OCR, live captions and acoustic alerts, personalized multimodal feedback, hazard-aware navigation, cognitive cueing, and low-burden correction
& \textit{\tb{Visual access and reading:}} VizWiz~\cite{VizWiz}; WearVQA~\cite{WearVQA}; TextVQA~\cite{TowardsVQAModelsThatCanRead}; OCR-Wearable~\cite{EvaluatingOCRPerformance}; GLIMPSE~\cite{GLIMPSE}; SuperGlasses~\cite{Superglasses}. \textit{\tb{Situated and mobility assistance:}} SAW-Bench~\cite{SAW-Bench}; UrbanRiskVQA~\cite{UrbanRiskAwareNavigation}; egocentric pedestrian-intention understanding~\cite{DecodingPedestrianCrossingIntention}. \textit{\tb{Communication and inclusive interaction:}} AR-DeafEducation~\cite{EvaluatingARForDeafStudents}; MixedVision~\cite{ReshapingInclusiveInterpersonalDynamics}; ConversationBreakdowns~\cite{Conversational}; AVA-ActiveSpeaker~\cite{AvaActiveSpeaker}
& \textit{\tb{Assistive products:}} Envision Glasses~\cite{sgprod2026_envision_envision_glasses}; OrCam MyEye~\cite{sgprod2026_orcam_orcam_myeye_2_myeye_pro}; NuEyes E2+~\cite{sgprod2026_nueyes_nueyes_e2}; eSight Go~\cite{sgprod2026_esight_esight_go}. \textit{\tb{Navigation and audio systems:}} NavCog~\cite{NavCog}; LidSonic~\cite{LidSonic}; OpenEarANC~\cite{ActiveNoiseCancellation}. \textit{\tb{Research and low-burden display routes:}} OpenGlass~\cite{OpenGlass}; Pupil Labs Neon~\cite{sgprod2026_pupil_labs_pupil_labs_neon}; Even Realities G1~\cite{sgprod2026_even_realities_even_realities_g1}; Halliday DigiWindow Glasses~\cite{sgprod2026_halliday_halliday_digiwindow_glasses}; Vuzix Z100~\cite{sgprod2026_vuzix_vuzix_z100}; Meta Ray-Ban Display~\cite{sgprod2026_meta_ray_ban_meta_ray_ban_display} \\
\midrule
Industrial workflow support (\cref{sec:industrial})
& SOP-grounded step tracking, inspection and identification, proactive error detection and recovery, remote-expert collaboration, role-based authorization, offline fallback, and auditable logging
& \textit{\tb{Procedural and assembly understanding:}} HoloAssist~\cite{Holoassist}; Assembly101~\cite{Assembly101}; IKEA-ASM~\cite{IKEA-ASM}; COIN~\cite{Coin}; CrossTask~\cite{CrossTaskWeaklySupervisedLearning}; Ego-Exo4D~\cite{Ego-Exo4D}. \textit{\tb{Streaming intervention and recovery:}} Plan-Watch-Recover/Pro2Bench~\cite{PlanWatchRecover}; EgoPro-Bench~\cite{EgoPro-Bench}; Streaming Interventions~\cite{StreamingInterventions}; IPIBench~\cite{IPIBench}; EgoSAT~\cite{EgoSAT}. \textit{\tb{General first-person substrates:}} Ego4D~\cite{Ego4D}; Ego-1K~\cite{Ego-1K}; EPIC-KITCHENS-100~\cite{EK-100}
& \textit{\tb{Research assistance and edge systems:}} Pro2Assist~\cite{Pro2Assist}; VisionClaw~\cite{VisionClaw}; OpenGlass~\cite{OpenGlass}; EPIC~\cite{EPIC}; Project Aria~\cite{ProjectAria}; Aria Gen 2~\cite{sgprod2026_meta_reality_labs_research_aria_gen_2}. \textit{\tb{Enterprise and industrial entries:}} RealWear Navigator 520~\cite{sgprod2026_realwear_realwear_navigator_520}; Vuzix M400~\cite{sgprod2026_vuzix_vuzix_m400}; Google Glass Enterprise Edition 2~\cite{sgprod2026_google_google_glass_enterprise_edition_2}; Microsoft HoloLens 2~\cite{sgprod2026_microsoft_microsoft_hololens_2}; Magic Leap 2~\cite{sgprod2026_magic_leap_magic_leap_2} \\
\midrule
Healthcare and caregiving (\cref{sec:medical-care})
& procedure capture and documentation, professionally supervised guidance, rehabilitation and home-care reminders, longitudinal personal context, PHI governance, and clinically meaningful outcome validation
& \textit{\tb{Clinical procedure and skill:}} Cholec80/EndoNet~\cite{Endonet}; JIGSAWS~\cite{JIGSAWS}. \textit{\tb{Medical VQA proxies:}} VQA-RAD~\cite{ClinicallyGeneratedVQA}; SLAKE~\cite{Slake}. \textit{\tb{Wearable procedural and multiview proxies:}} HoloAssist~\cite{Holoassist}; Ego-Exo4D~\cite{Ego-Exo4D}; Ego4D~\cite{Ego4D}. \textit{Longitudinal care and memory proxies:} EgoLife~\cite{Egolife}; PVCL~\cite{PersonalVisualContextLearning}; EgoMemReason~\cite{EgoMemReason}; EGOSTREAM~\cite{EgoStream}; H2HMem~\cite{H2HMem}; LightMem-Ego~\cite{LightMem-Ego}
& \textit{\tb{Clinical-training and sensing platforms:}} AR healthcare-education systems surveyed in~\cite{ARinMedicalEducation}; Microsoft HoloLens 2~\cite{sgprod2026_microsoft_microsoft_hololens_2}; Magic Leap 2~\cite{sgprod2026_magic_leap_magic_leap_2}; Project Aria~\cite{ProjectAria}; Aria Gen 2~\cite{sgprod2026_meta_reality_labs_research_aria_gen_2}; Pupil Labs Neon~\cite{sgprod2026_pupil_labs_pupil_labs_neon}; Tobii Pro Glasses 3~\cite{sgprod2026_tobii_tobii_pro_glasses_3}. \textit{\tb{Assistive and caregiving entries:}} Envision Glasses~\cite{sgprod2026_envision_envision_glasses}; OrCam MyEye~\cite{sgprod2026_orcam_orcam_myeye_2_myeye_pro}; NuEyes E2+~\cite{sgprod2026_nueyes_nueyes_e2}; eSight Go~\cite{sgprod2026_esight_esight_go}; OpenGlass~\cite{OpenGlass} \\
\midrule
Education and skills training (\cref{sec:education-training})
& demonstration capture, skill decomposition, learner-state estimation, appropriately timed feedback, error-aware practice, reflection, accessibility, retention, and delayed transfer
& \textit{\tb{Demonstration and skill capture:}} Ego-Exo4D~\cite{Ego-Exo4D}; HoloAssist~\cite{Holoassist}; Ego-1K~\cite{Ego-1K}; Ego4D~\cite{Ego4D}; EPIC-KITCHENS-100~\cite{EK-100}. \textit{\tb{Instructional and procedural understanding:}} COIN~\cite{Coin}; CrossTask~\cite{CrossTaskWeaklySupervisedLearning}; Assembly101~\cite{Assembly101}; IKEA-ASM~\cite{IKEA-ASM}. \textit{\tb{Intervention and inclusive-learning evaluation:}} Plan-Watch-Recover/Pro2Bench~\cite{PlanWatchRecover}; EgoPro-Bench~\cite{EgoPro-Bench}; Streaming Interventions~\cite{StreamingInterventions}; IPIBench~\cite{IPIBench}; AR-DeafEducation~\cite{EvaluatingARForDeafStudents}; MixedVision~\cite{ReshapingInclusiveInterpersonalDynamics}
& \textit{\tb{Interactive coaching systems:}} Pro2Assist~\cite{Pro2Assist}; VisionClaw~\cite{VisionClaw}; OpenGlass~\cite{OpenGlass}. \textit{\tb{Immersive and display platforms:}} Microsoft HoloLens 2~\cite{sgprod2026_microsoft_microsoft_hololens_2}; Magic Leap 2~\cite{sgprod2026_magic_leap_magic_leap_2}; Meta Ray-Ban Display~\cite{sgprod2026_meta_ray_ban_meta_ray_ban_display}; Snap Specs~\cite{sgprod2026_snap_snap_specs_2026}; Vuzix Z100~\cite{sgprod2026_vuzix_vuzix_z100}. \textit{\tb{Learning-analysis and vocational routes:}} Aria Gen 2~\cite{sgprod2026_meta_reality_labs_research_aria_gen_2}; Pupil Labs Neon~\cite{sgprod2026_pupil_labs_pupil_labs_neon}; Tobii Pro Glasses 3~\cite{sgprod2026_tobii_tobii_pro_glasses_3}; RealWear Navigator 520~\cite{sgprod2026_realwear_realwear_navigator_520}; Vuzix M400~\cite{sgprod2026_vuzix_vuzix_m400} \\
\midrule
Mobility and transportation safety (\cref{sec:mobile-safety})
& robust localization and route guidance, hazard and traffic-intention understanding, low-latency multimodal alerts, environmental-audio preservation, outdoor robustness, and safe user override
& \textit{\tb{Hazard and traffic understanding:}} UrbanRiskVQA~\cite{UrbanRiskAwareNavigation}; JAAD~\cite{PedestrianCrosswalkBehavior}; PIE~\cite{Pie}; BDD100K~\cite{Bdd100k}; egocentric pedestrian-intention VLM~\cite{DecodingPedestrianCrossingIntention}. \textit{\tb{Navigation and embodied grounding proxies:}} Habitat~\cite{Habitat}; R2R~\cite{VLN}; REVERIE~\cite{Reverie}; EmbodiedQA~\cite{EmbodiedQA}; OpenEQA~\cite{Openeqa}. \textit{\tb{Spatial-map and real-device proxies:}} Aria Digital Twin~\cite{Ariadigitaltwin}; Matterport3D~\cite{Matterport3d}; ScanNet~\cite{Scannet}; SpatialWorld~\cite{SpatialWorld}
& \textit{\tb{Navigation, sensing, and localization systems:}} NavCog~\cite{NavCog}; LidSonic~\cite{LidSonic}; OpenEarANC~\cite{ActiveNoiseCancellation}; VINS-Mono~\cite{Vins-mono}; ORB-SLAM3~\cite{Orb-slam3}; DROID-SLAM~\cite{Droid-slam}; LEVIO~\cite{LEVIO}; Project Aria~\cite{ProjectAria}; Aria Gen 2~\cite{sgprod2026_meta_reality_labs_research_aria_gen_2}. \textit{\tb{Outdoor and visual-display entries:}} Oakley Meta Vanguard~\cite{sgprod2026_meta_oakley_oakley_meta_vanguard}; RayNeo V3~\cite{sgprod2026_tcl_rayneo_rayneo_v3_ai_shooting_glasses}; Meta Ray-Ban Display~\cite{sgprod2026_meta_ray_ban_meta_ray_ban_display}; RayNeo X3 Pro~\cite{sgprod2026_tcl_rayneo_rayneo_x3_pro}; XREAL AURA~\cite{sgprod2026_xreal_google_xreal_aura_project_aura} \\
\bottomrule
\end{tabular}
\end{adjustbox}
\end{table}

\subsection{Daily Situated Assistance}
\label{sec:daily-assistant}

Daily situated assistance treats smart glasses as a low-friction interface connecting the wearer's current first-person view, prior personal context, and external digital services. The scene spans reading, translation, conversational support, information retrieval, object finding, reminders, meeting assistance, travel and retail queries, and creative capture. Camera/audio-first products already provide practical entry points for user-initiated L2 assistance, whereas reliable cross-time memory and externally consequential service execution require L3 persistent-state management and L4 permission-governed action. The application therefore develops along a natural progression from understanding what is happening now, to remembering what happened before, and finally to acting on the wearer's behalf.

\sgpoint{Perceptual access and conversational support}{
At the immediate time scale, the glasses must convert a moving, wearer-centered stream into concise and correctly grounded assistance. Typical functions include reading signs and documents, translating visible or spoken language, answering questions about nearby objects, identifying relevant information during shopping or travel, and supporting conversations or meetings without requiring the wearer to hold a phone. SuperGlasses~\cite{Superglasses}, SAW-Bench~\cite{SAW-Bench}, GLIMPSE~\cite{GLIMPSE}, and EgoSAT~\cite{EgoSAT} collectively cover wearable question answering, text understanding, situated awareness, and continuous-stream interaction. Ray-Ban Meta Gen 2~\cite{sgprod2026_meta_ray_ban_ray_ban_meta_gen_2}, Xiaomi AI Glasses~\cite{sgprod2026_xiaomi_xiaomi_ai_glasses}, Rokid AI Glasses Style~\cite{sgprod2026_rokid_rokid_ai_glasses_style}, and Solos AirGo V2~\cite{sgprod2026_solos_solos_airgo_v2} illustrate camera/audio-first or lightweight feedback routes. Their usefulness depends not only on model accuracy, but also on wake-up reliability, speech robustness, time-to-first-useful-feedback, and whether the output can be verified through the available audio or visual channel.
}

\sgpoint{Personal context, memory, and anticipatory assistance}{
A more persistent assistant must connect current observations with the wearer's history: where an object was last seen, what was discussed earlier, which task remains unfinished, or which reminder is relevant to the present context. EgoLife~\cite{Egolife} moves toward an egocentric life assistant; PVCL~\cite{PersonalVisualContextLearning} studies personalized visual context; ContextAgent~\cite{ContextAgent} proposes a framework in which LLM agents continuously interpret open-world multimodal sensory streams and maintain contextual state to anticipate user needs; EgoMemReason~\cite{EgoMemReason}, EGOSTREAM~\cite{EgoStream}, and EgoExoMem~\cite{EgoExoMem} target long-horizon, streaming, and cross-view memory reasoning; and LightMem-Ego~\cite{LightMem-Ego} provides a further entry point for everyday personal memory. These directions extend assistance from isolated responses to L3 stateful support, but they also make memory provenance, uncertainty, correction, selective forgetting, and deletion first-class requirements. A false answer about the current scene may be immediately corrected, whereas a false stored memory can be repeatedly reused and can silently contaminate later reminders or decisions. Evaluation must therefore include cross-event consistency, false recall, user correction cost, memory retention policy, and usefulness over repeated or multi-day wear.
}

\sgpoint{Agentic retrieval and permission-governed service execution}{
Daily assistance becomes action-capable when first-person context is connected to web search, mobile services, communication tools, or other applications. Ego2Web~\cite{Ego2Web} grounds web-agent tasks in egocentric video, Egocentric Co-Pilot~\cite{EgocentricCoPilot} explores web-native smart-glasses agents, VisionClaw~\cite{VisionClaw} considers always-on agents through smart glasses, and agentic long-video understanding~\cite{AgenticVideoUnderstanding} provides a route for decomposing extended observations into tool-mediated reasoning steps. Such systems could retrieve venue information, prepare messages, organize captured content, or support bookings and purchases. However, the transition from recommendation to execution changes the responsibility boundary: the system must expose the intended action, request confirmation at an appropriate granularity, enforce account and task permissions, preserve an audit trail, and support cancellation or rollback. Payments, bookings, data sharing, and other externally consequential actions should therefore be evaluated separately from low-risk information retrieval rather than being folded into a single assistant score.
}

\sgpoint{End-to-end utility, robustness, and evidence gaps}{
A defensible daily-assistance evaluation should jointly report answer faithfulness, visual and temporal grounding, tail latency, interruption burden, false activation, user correction, privacy leakage, memory provenance, tool-action recovery, and repeated-use utility. The relevant operating conditions include background speech, changing illumination, partial visibility, intermittent connectivity, limited battery, and product or service updates. OpenGlass~\cite{OpenGlass} and EPIC~\cite{EPIC} illustrate system-level routes for efficient or split wearable perception, while PrivacyUtility~\cite{Position} and MindTheGap~\cite{MindTheGap} clarify why sustained capture must be evaluated together with privacy and contextual use. No cited resource yet combines sustained wear, memory editing, tool rollback, and product-version drift within a common protocol. Travel, retail, and creative-capture uses further require location and venue-policy updates, consumer privacy and payment security, and user control over captured or generated content. The appropriate evidence threshold should follow the consequence of the subtask: a question can often be clarified or retried, while an incorrect external action may affect money, privacy, or third parties.
}

\subsection{Accessibility Assistance}
\label{sec:accessibility}

Accessibility assistance uses smart glasses to improve independent access to textual, environmental, communicative, and mobility-related information. The relevant users and needs are heterogeneous: low-vision support, blindness assistance, hearing access, cognitive cueing, and mobility guidance require different sensing configurations, feedback modalities, timing constraints, and tolerance for error. Specialized products such as Envision Glasses~\cite{sgprod2026_envision_envision_glasses}, OrCam MyEye~\cite{sgprod2026_orcam_orcam_myeye_2_myeye_pro}, NuEyes E2+~\cite{sgprod2026_nueyes_nueyes_e2}, and eSight Go~\cite{sgprod2026_esight_esight_go} therefore represent different assistive routes rather than interchangeable solutions. The central deployment criterion is not generic model capability, but whether the complete system improves independent task performance for a clearly defined target population without imposing excessive workload or new safety risks.

\sgpoint{Visual access and environmental understanding}{
For blind or low-vision users, camera-equipped glasses can support scene description, text reading, object and person recognition, product identification, and visual-detail enhancement. VizWiz~\cite{VizWiz} captures questions submitted by blind users, EgoBlind~\cite{EgoBlind} explores how to convert continuously perceived surroundings and activities into timely guidance, OCR-Wearable~\cite{EvaluatingOCRPerformance} examines how walking speed and camera configuration affect recognition, and GLIMPSE~\cite{GLIMPSE} targets real-time text recognition and contextual understanding for wearables. These resources show why laboratory image accuracy is insufficient: the relevant input is continuously affected by head motion, framing, occlusion, viewing distance, and the user's inability to visually verify an answer. Useful feedback must prioritize task-relevant content, communicate uncertainty, and support rapid re-query or correction. Product configurations also differ between camera-based semantic assistance and optical visual enhancement, so evaluations should stratify results by functional need rather than aggregating distinct user groups.
}

\sgpoint{Communication access and auditory awareness}{
Live captions, speech translation, speaker identification, and acoustic-event alerts can improve participation in classrooms, workplaces, and everyday conversations. No-camera or lightweight HUD products such as Even Realities G1~\cite{sgprod2026_even_realities_even_realities_g1}, Halliday DigiWindow Glasses~\cite{sgprod2026_halliday_halliday_digiwindow_glasses}, and Vuzix Z100~\cite{sgprod2026_vuzix_vuzix_z100} illustrate low-burden visual cueing, while camera/audio systems can additionally use visual context to identify speakers or disambiguate references. AR-DeafEducation evaluates AR-mediated communication access for deaf students in experiential higher-education settings~\cite{EvaluatingARForDeafStudents}. The deployment challenge is to maintain caption timeliness, readability, and attribution without blocking the user's view or leaking private audio. Feedback granularity, font size, contrast, display position, and interruption policy must be personalized, and performance should be evaluated in noisy, multi-speaker, and mobile settings rather than only on clean speech.
}

\sgpoint{Mobility, hazard awareness, and cognitive cueing}{
Navigation and cognitive support require a longer and more safety-sensitive loop than isolated recognition. NavCog~\cite{NavCog} provides a system-level reference for navigation assistance, LidSonic~\cite{LidSonic} illustrates active sensing for visually impaired users, and UrbanRiskVQA~\cite{UrbanRiskAwareNavigation} targets urban-risk understanding. Smart glasses may also provide context-linked reminders for destinations, daily routines, or medication-related self-management. These functions depend on reliable localization, timely hazard detection, an appropriate balance between audio and visual feedback, and a mechanism for recovering when the system is uncertain or the environment changes. A missed obstacle, crossing hazard, or reminder and an unnecessary alarm impose different costs; both must be reported. Higher-consequence assistance should therefore include route-level testing, near-miss recording, false-alarm burden, recovery behavior, and clear boundaries indicating when the user should rely on another aid or human support.
}

\sgpoint{Personalization, independent outcomes, and responsible deployment}{
Evidence from VizWiz~\cite{VizWiz}, OCR-Wearable~\cite{EvaluatingOCRPerformance}, UrbanRiskVQA~\cite{UrbanRiskAwareNavigation}, and AR-DeafEducation~\cite{EvaluatingARForDeafStudents} spans different users, tasks, and operating conditions, reinforcing that average accuracy cannot represent assistive value across populations and activities. Evaluation should measure independent task completion, time and effort, cognitive load, visual and auditory demand, error recovery, near-miss events, personalization cost, and longitudinal adoption, with results stratified by functional need and task risk. The system should allow users to control feedback modality, detail, pace, volume, contrast, and memory, while caregivers or institutions should receive only the permissions necessary for the intended task. Bystander consent and recording transparency remain relevant whenever cameras or microphones capture surrounding people. Evidence from one population or controlled activity should not be generalized to another without validation, and improvements in component recognition should be distinguished from sustained gains in independence, confidence, participation, or self-management.
}

\subsection{Industrial Workflow Support}
\label{sec:industrial}

Industrial workflow support places smart glasses inside structured processes such as assembly, maintenance, inspection, warehouse picking and packing, laboratory operations, and field service. Unlike open-domain consumer assistance, these scenes are constrained by Standard Operating Procedures (SOPs), station- and role-specific permissions, quality targets, safety rules, and audit responsibilities. RealWear Navigator 520~\cite{sgprod2026_realwear_realwear_navigator_520}, Vuzix M400~\cite{sgprod2026_vuzix_vuzix_m400}, and Google Glass Enterprise Edition 2~\cite{sgprod2026_google_google_glass_enterprise_edition_2} emphasize ruggedized form factors, hands-free interaction, Personal Protective Equipment (PPE) compatibility, and device management. The application loop therefore progresses from recognizing the current work state, to providing timely guidance and recovery, and finally to producing reliable organizational evidence about what occurred.

\sgpoint{SOP-grounded execution and procedural guidance}{
Assembly, maintenance, and logistics tasks require the glasses to identify the current step, relevant object or tool, completed prerequisites, and the next permissible action. Assembly101~\cite{Assembly101}, IKEA-ASM~\cite{IKEA-ASM}, COIN~\cite{Coin}, and CrossTask~\cite{CrossTaskWeaklySupervisedLearning} provide complementary resources for procedural activity, instructional video, and step structure, LabOS~\cite{LabOS} exemplifies situated human–AI collaboration in scientific workspaces, while HoloAssist~\cite{Holoassist} introduces interactive errors and help-seeking behavior. A deployable assistant must go beyond offline step classification: it should maintain task progress under interruptions, distinguish acceptable variation from a true deviation, retrieve the correct local SOP version, and present concise instructions that do not occlude the work area. The value of the system should be measured through completion time, rework, first-pass quality, and worker workload rather than recognition accuracy alone.
}

\sgpoint{Inspection, identification, and operational documentation}{
Inspection and field workflows add barcode and OCR-based identification, equipment-state grounding, quality checks, evidence capture, and incident review. The manufacturing review~\cite{ARinManufacturing}, wearable OCR evidence~\cite{EvaluatingOCRPerformance}, and rugged product routes represented by RealWear Navigator 520~\cite{sgprod2026_realwear_realwear_navigator_520} and Vuzix M400~\cite{sgprod2026_vuzix_vuzix_m400} provide complementary support for this application block. Laboratory use further involves chemical labels, instrument settings, sample identity, and strict documentation. These tasks require high-resolution close-range perception, temporal comparison with an expected state, and explicit uncertainty when labels or components are partially visible. Because the glasses may also create compliance records, timestamps, identity, calibration state, and provenance must be preserved rather than reconstructed after the fact. Local or edge processing is valuable when sensitive production data cannot leave the site or when connectivity is weak. The resulting logs should be searchable and auditable, but data minimization and role-based access are necessary to prevent continuous workplace capture from becoming unnecessary worker surveillance.
}

\sgpoint{Remote expertise, proactive intervention, and error recovery}{
First-person streaming enables remote experts to see the wearer's task context, annotate relevant regions, and guide recovery without occupying the worker's hands. HoloAssist~\cite{Holoassist} provides a foundation for interactive assistance; Pro2Assist~\cite{Pro2Assist} studies continuous, step-aware proactive guidance; Plan-Watch-Recover/Pro2Bench~\cite{PlanWatchRecover} focuses on planning, monitoring, and recovery; and Streaming Interventions~\cite{StreamingInterventions} and IPIBench~\cite{IPIBench} examine correction or intervention under continuous streams. The key question is not only whether an error can be recognized, but whether assistance arrives before the error propagates, selects the correct level of intervention, and hands control to an expert when uncertainty is high. Field systems must also tolerate weak networks, expert-response delays, task handoffs, and partial video while preserving a consistent shared task state.
}

\sgpoint{Field robustness, governance, and organizational outcomes}{
The manufacturing review~\cite{ARinManufacturing} and enterprise product routes including RealWear Navigator 520~\cite{sgprod2026_realwear_realwear_navigator_520}, Vuzix M400~\cite{sgprod2026_vuzix_vuzix_m400}, and Google Glass Enterprise Edition 2~\cite{sgprod2026_google_google_glass_enterprise_edition_2} motivate evaluation under actual work constraints. Industrial validation must cover the operating envelope in which workers actually use the glasses: noise, gloves, dust or oil, temperature, visual occlusion, protective equipment, intermittent connectivity, and outdated or conflicting SOPs. Relevant outcomes include throughput, rework rate, safety incidents, recovery under network disruption, hands-free efficiency, authorization failures, log integrity, and auditability. Fleet deployment additionally requires configuration control, software and model versioning, device health monitoring, and revocation of permissions when workers or roles change. A credible deployment claim should demonstrate cross-station and cross-team robustness and should trace system errors through the organizational responsibility chain. Component benchmarks remain useful, but they do not substitute for field evidence that the complete system improves work outcomes without increasing distraction, privacy risk, or procedural ambiguity.
}

\subsection{Healthcare and Caregiving}
\label{sec:medical-care}

Healthcare and caregiving span clinical workflow capture, professional education, rehabilitation coaching, medication and daily-living reminders, elder care, and potentially decision support. The same perception or cueing mechanism can have very different consequences depending on whether it records a procedure, reminds a user of a routine, or influences diagnosis or treatment. As the application moves from passive capture toward clinical recommendation, the responsibility chain expands to include clinicians, patients, caregivers, institutions, and regulators. Smart-glasses evidence must therefore be interpreted according to the intended use and consequence of error rather than according to model capability alone.

\sgpoint{Clinical workflow capture and documentation}{
Hands-free first-person capture can document procedures, preserve a clinician's view, support later review, and reduce manual note-taking. Cholec80/EndoNet~\cite{Endonet} and JIGSAWS~\cite{JIGSAWS} provide references for surgical-phase recognition and skill or action analysis, although neither alone validates a wearable clinical system. In a smart-glasses deployment, the system must synchronize observations with the correct patient, procedure, time, and professional role, and it should distinguish automatically generated content from clinician-confirmed documentation. Missing events, incorrect attribution, or incomplete capture may be more consequential than ordinary video-recognition errors because they enter a legal and clinical record. Evaluation should therefore include documentation completeness, professional correction time, override behavior, provenance, tail latency, and failure detection, together with clear controls governing when recording starts, who can access it, and how long it is retained.
}

\sgpoint{Professional education and procedure-aware assistance}{
Medical education and supervised training are currently more strongly supported than autonomous clinical decision making. The systematic review of AR in medical education~\cite{ARinMedicalEducation} summarizes educational uses; HoloAssist~\cite{Holoassist} provides a procedural proxy for step understanding and help; and Ego-Exo4D~\cite{Ego-Exo4D} offers multiview capture of skilled activity. These resources motivate demonstration replay, viewpoint sharing, step-aware prompts, and post-hoc skill analysis. However, a training system and a clinical guidance system should not be treated as equivalent: the latter requires professional oversight, validated content, controlled update procedures, and explicit escalation when observations fall outside the supported use. Evaluations should separate knowledge or skill acquisition from immediate procedure completion and should report whether the wearer can recognize, reject, or correct an inappropriate prompt.
}

\sgpoint{Rehabilitation, elder care, and home support}{
In home and long-term care, smart glasses can support routine reminders, activity coaching, object or medication finding, remote caregiver communication, and accessible perception. Envision Glasses~\cite{sgprod2026_envision_envision_glasses} provides a product route for assistive perception, ~\cite{WearableARforRestorativeBreaks} investigates wearable augmented-reality narrative experiences as restorative breaks for young people, and EgoLife~\cite{Egolife}, PVCL~\cite{PersonalVisualContextLearning}, and related personal-memory resources motivate context-aware support across repeated activities. These scenes require persistent personal state, but they also expose household members and visitors to continuous sensing. A useful system must distinguish a missed reminder from a completed task, manage false-alarm escalation, allow the wearer to correct memory, and define what caregivers may view or change. Longitudinal outcomes such as adherence, independence, caregiver burden, and sustained acceptance are more informative than one-session model accuracy.
}

\sgpoint{Clinical evidence, PHI governance, and responsibility boundaries}{
VQA-RAD~\cite{ClinicallyGeneratedVQA} and SLAKE~\cite{Slake} provide medical-image question-answering proxies, but they are not direct evidence for smart-glasses deployment in real care pathways. System-level claims require professionally annotated workflow logs, validation under representative clinical conditions, and measurement of clinical or care outcomes. Protected Health Information (PHI) governance, consent, jurisdiction, liability, infection-control and hygiene requirements, device cleaning, network security, and audit trails jointly constrain the permissible design. Professional confirmation and override should be reported as system outcomes rather than treated as implementation details. Even when an underlying function resembles L2 cueing in everyday assistance, the validation threshold is higher because errors may propagate into regulated decisions, clinical records, or long-term care responsibilities.
}

\subsection{Education and Skills Training}
\label{sec:education-training}

Education and skills training use smart glasses to support durable knowledge and skill acquisition rather than merely completing the current task. Relevant activities include laboratory instruction, cooking, repair, sports, music, language learning, creative practice, and vocational education. The application loop can capture an expert demonstration, decompose it into meaningful steps, estimate the learner's current state, select feedback, and support later reflection. Its central criterion is whether assistance strengthens independent mastery and delayed transfer; a system that produces immediate compliance while creating persistent prompt dependence may be effective as a workflow aid but ineffective as an educational technology.

\sgpoint{Demonstration capture and skill decomposition}{
First-person and multiview recordings provide a natural substrate for showing what an expert attended to, which objects were manipulated, and how a procedure unfolded. Ego-Exo4D~\cite{Ego-Exo4D} aligns first- and third-person views of skilled activity, HoloAssist~\cite{Holoassist} captures interactive procedural behavior, Ego-1K~\cite{Ego-1K} broadens multiview egocentric analysis, and COIN~\cite{Coin}, CrossTask~\cite{CrossTaskWeaklySupervisedLearning}, Assembly101~\cite{Assembly101}, and IKEA-ASM~\cite{IKEA-ASM} provide instructional and procedural references. For training, the representation should preserve subgoals, key state changes, common deviations, and expert rationale rather than only action labels. Camera/audio-first glasses support naturalistic capture, whereas gaze- and pose-sensing platforms can support finer analysis of attention and technique. Dataset coverage, however, should not be mistaken for evidence that a particular representation improves learning.
}

\sgpoint{In-situ coaching and error-aware practice}{
During practice, the glasses can provide concise prompts, detect a missed or incorrect step, demonstrate a correction, or defer to a teacher. Pro2Assist~\cite{Pro2Assist}, Plan-Watch-Recover~\cite{PlanWatchRecover}, Streaming Interventions~\cite{StreamingInterventions}, and IPIBench~\cite{IPIBench} collectively motivate assistance that is aware of task progress and intervention timing. The educational objective changes the optimal policy: immediate correction may maximize short-term task success, while delayed hints, questions, or fading support may better promote recall and problem solving. The system should therefore model not only task state but also assistance history and learner response. Evaluation should report intervention precision, recovery, cognitive load, prompt dependence, and the learner's ability to continue after assistance is removed.
}

\sgpoint{Reflection, personalization, and inclusive learning}{
Smart glasses can also support post-practice review by linking first-person observations, external views, spoken explanations, errors, and feedback to specific moments, as enabled by the multiview and interaction structures in Ego-Exo4D~\cite{Ego-Exo4D} and HoloAssist~\cite{Holoassist}. Different learners may require different prompt detail, pacing, modality, and frequency; creative or open-ended skills additionally require preserving learner choice rather than optimizing toward a single canonical sequence. AR-DeafEducation~\cite{EvaluatingARForDeafStudents} demonstrates the importance of communication access in experiential higher education, while HUD and camera-and-display products provide alternative routes for captions, demonstrations, and private cues. Reflection tools should make model interpretations inspectable and allow learners or teachers to annotate mistakes, correct task state, and select what is retained. Personalized support is valuable only when the cost of calibration and the risks of inappropriate adaptation are included in evaluation.
}

\sgpoint{Retention, transfer, agency, and data governance}{
The current procedural-assistance resources, including Pro2Assist~\cite{Pro2Assist}, Plan-Watch-Recover~\cite{PlanWatchRecover}, Ego-Exo4D~\cite{Ego-Exo4D}, and HoloAssist~\cite{Holoassist}, primarily capture demonstrations or in-task assistance rather than long-term educational outcomes. A valid educational study should distinguish immediate completion from retention, delayed transfer to a new instance, and independent performance without the glasses. Excessive prompting can reduce learner agency; insufficient or poorly timed prompting can create repeated failures without learning benefit. Core outcomes therefore include retention, transfer, prompt fading, self-correction, confidence, cognitive load, and user or teacher override. Access rights also differ across private learner review, teacher supervision, classroom display, peer collaboration, and employer-managed vocational training. When minors or employees are involved, capture, sharing, retention, and performance analytics require explicit governance. Existing datasets mainly characterize demonstrations and training interactions; they do not yet provide a coherent longitudinal chain from wearable assistance to durable skill acquisition.
}

\subsection{Mobility and Transportation Safety}
\label{sec:mobile-safety}

Mobility and transportation safety cover pedestrian wayfinding, cycling and running guidance, hazard detection, nighttime mobility, public-transport reminders, and emergency alerts. These scenes are defined by continuous motion and a limited decision window: the system must perceive, reason, and communicate early enough for the wearer to act, while avoiding feedback that masks environmental sound or captures too much visual attention. Component-level navigation and traffic benchmarks are useful, but system-level safety claims require route-based evidence, user responses, and near-miss outcomes under realistic outdoor conditions.

\sgpoint{Wayfinding and route-level guidance}{
Pedestrian and public-transport assistance requires localization, route progress, turn selection, destination awareness, and recovery from deviation. NavCog~\cite{NavCog} provides a system-level navigation reference, while UrbanRiskVQA~\cite{UrbanRiskAwareNavigation} connects route understanding with urban risk. For smart glasses, route instructions must be synchronized with the wearer's position and movement and presented through a modality that remains intelligible without dominating attention. Persistent spatial state is needed to distinguish a temporary occlusion from a wrong turn and to update guidance when entrances, paths, or transit conditions change. Evaluation should include wrong-turn rate, recovery distance, cue timing, localization failure, user override, and performance across familiar and unfamiliar routes rather than only destination success.
}

\sgpoint{Hazard detection and traffic-intention understanding}{
Safety assistance must identify hazards that matter to the wearer and estimate whether other road users are likely to enter the wearer's path. JAAD~\cite{PedestrianCrosswalkBehavior}, PIE~\cite{Pie}, and BDD100K~\cite{Bdd100k} provide third-person road-scene and pedestrian-intention references, while an egocentric pedestrian VLM~\cite{DecodingPedestrianCrossingIntention} brings the task closer to the wearer's viewpoint. The transfer is not direct: a head-mounted camera has different motion, field of view, and occlusion patterns from a vehicle camera. Warnings should expose uncertainty and prioritize actionable hazards. Missed hazards and false alarms have asymmetric costs, so both must be reported together with reaction time, inappropriate user responses, alert habituation, and near-miss events.
}

\sgpoint{Sports mobility and outdoor multimodal interaction}{
Running and cycling add pace or route cues, capture, communication, and performance feedback under wind, sweat, vibration, and rapidly changing illumination. Oakley Meta Vanguard~\cite{sgprod2026_meta_oakley_oakley_meta_vanguard} illustrates a sports-oriented camera/audio profile, and RayNeo V3~\cite{sgprod2026_tcl_rayneo_rayneo_v3_ai_shooting_glasses} provides a lightweight capture-oriented entry point. OpenEarANC~\cite{ActiveNoiseCancellation} shows that environmental-sound preservation and wind-noise processing are themselves part of the assistance loop. Camera placement, Ingress Protection (IP) rating, battery endurance, microphone robustness, display legibility, and audio masking jointly define the operating envelope. Outdoor evaluation should therefore include motion-induced image degradation, wind and traffic noise, weather, nighttime conditions, battery depletion, and whether feedback changes the user's awareness of surrounding hazards.
}

\sgpoint{Safety-critical evaluation and operating boundaries}{
UrbanRiskVQA~\cite{UrbanRiskAwareNavigation}, NavCog~\cite{NavCog}, LidSonic~\cite{LidSonic}, OpenEarANC~\cite{ActiveNoiseCancellation}, and the egocentric pedestrian-intention study~\cite{DecodingPedestrianCrossingIntention} cover complementary parts of the loop but do not yet form a unified safety protocol. End-to-end warning latency, reaction time, hazard-miss rate, false-alarm burden, wrong-turn rate, display distraction, audio masking, near-miss events, and recovery behavior should be reported separately and stratified by illumination, weather, route complexity, traffic density, and user mobility characteristics. A false alarm can distract the wearer or erode trust, while a miss can leave a hazard entirely unreported; an aggregate accuracy score hides this asymmetry. Real-route trials, route replay, and synchronized user-response records are necessary to connect perception output to actual behavior. The system should also make its operating boundary explicit, including unsupported speeds, weather, road types, or visibility conditions, and should provide a safe fallback when sensing, localization, connectivity, or feedback becomes unreliable.
}

\begin{table}[thp!]
\centering
\caption{\textbf{Fine-grained capability requirements and expanded representative resources for spatial intelligence, social interaction and collaboration, and embodied intelligence.}
Unlike \cref{tab:application-evidence}, each of the final three major scenes is decomposed into task blocks. Direct smart-glasses resources are complemented by clearly identifiable spatial, dialogue, web-agent, or robot-side proxies where no unified wearable benchmark exists; system and product entries indicate implementation routes.}
\label{tab:late-application-benchmarks}
\scriptsize
\begin{adjustbox}{width=\linewidth}
\begin{tabular}{
>{\raggedright\arraybackslash}m{34mm}
>{\raggedright\arraybackslash}m{54mm}
>{\raggedright\arraybackslash}m{66mm}
>{\raggedright\arraybackslash}m{66mm}
}
\toprule
\textbf{Application Scene} &
\textbf{Key Capability Requirement} &
\textbf{Datasets/Benchmarks} &
\textbf{Systems, Platforms, or Product Entries} \\
\midrule
\rowcolor{aprilblue!20}
\multicolumn{4}{l}{\textit{\textbf{Spatial Intelligence}}} \\
\midrule
Spatial reconstruction and semantic mapping
& synchronized visual-inertial sensing, calibration, VIO/SLAM, egocentric 3D reconstruction, semantic mapping, and object/place grounding
& \textit{\tb{Egocentric and multiview resources:}} Aria Digital Twin~\cite{Ariadigitaltwin}; Ego-Exo4D~\cite{Ego-Exo4D}; Ego-1K~\cite{Ego-1K}. \textit{\tb{Indoor 3D reconstruction proxies:}} Matterport3D~\cite{Matterport3d}; ScanNet~\cite{Scannet}. \textit{\tb{Long-form first-person substrates:}} Ego4D~\cite{Ego4D}; EPIC-KITCHENS-100~\cite{EK-100}
& \textit{\tb{Wearable sensing and collection platforms:}} Project Aria~\cite{ProjectAria}; Aria Gen 1~\cite{ProjectAria}; Aria Gen 2~\cite{sgprod2026_meta_reality_labs_research_aria_gen_2}; EgoKit~\cite{EgoKit}. \textit{\tb{Localization and mapping systems:}} VINS-Mono~\cite{Vins-mono}; ORB-SLAM3~\cite{Orb-slam3}; DROID-SLAM~\cite{Droid-slam}; LEVIO~\cite{LEVIO}; EPIC~\cite{EPIC} \\
\midrule
Navigation and spatial question answering
& real-device localization, instruction following, remote-object grounding, world-locked cueing, route recovery, and attentional control
& \textit{\tb{Navigation and embodied QA:}} Habitat~\cite{Habitat}; R2R~\cite{VLN}; REVERIE~\cite{Reverie}; EmbodiedQA~\cite{EmbodiedQA}; OpenEQA~\cite{Openeqa}. \textit{\tb{Scene and map substrates:}} Matterport3D~\cite{Matterport3d}; ScanNet~\cite{Scannet}; Aria Digital Twin~\cite{Ariadigitaltwin}. \textit{\tb{Interactive and safety-oriented extensions:}} SpatialWorld~\cite{SpatialWorld}; UrbanRiskVQA~\cite{UrbanRiskAwareNavigation}
& \textit{\tb{Navigation and AR platforms:}} NavCog~\cite{NavCog}; Project Aria~\cite{ProjectAria}; LEVIO~\cite{LEVIO}; Microsoft HoloLens 2~\cite{sgprod2026_microsoft_microsoft_hololens_2}; Magic Leap 2~\cite{sgprod2026_magic_leap_magic_leap_2}. \textit{\tb{Lightweight visual-display entries:}} RayNeo X3 Pro~\cite{sgprod2026_tcl_rayneo_rayneo_x3_pro}; XREAL AURA~\cite{sgprod2026_xreal_google_xreal_aura_project_aura}; Meta Ray-Ban Display~\cite{sgprod2026_meta_ray_ban_meta_ray_ban_display}; Snap Specs~\cite{sgprod2026_snap_snap_specs_2026}; Vuzix Z100~\cite{sgprod2026_vuzix_vuzix_z100} \\
\midrule
Deictic reference, gaze, and near-body relations
& pointing and gaze-conditioned reference, 3D proximity reasoning, multi-object disambiguation, and low-cost confirmation
& \textit{\tb{Direct reference and proximity evaluation:}} PointingMLLM/EgoPoint-Bench~\cite{MLLMs-Pointing}; EgoProx~\cite{EgoProx}; EGTEA Gaze+~\cite{sgacad2018_egtea_gaze}. \textit{\tb{Interaction and spatial proxies:}} Ego-Exo4D~\cite{Ego-Exo4D}; HoloAssist~\cite{Holoassist}; REVERIE~\cite{Reverie}; SpatialWorld~\cite{SpatialWorld}
& \textit{\tb{Gaze and multimodal sensing platforms:}} Pupil Labs Neon~\cite{sgprod2026_pupil_labs_pupil_labs_neon}; Tobii Pro Glasses 3~\cite{sgprod2026_tobii_tobii_pro_glasses_3}; Project Aria~\cite{ProjectAria}; Aria Gen 1~\cite{ProjectAria}; Aria Gen 2~\cite{sgprod2026_meta_reality_labs_research_aria_gen_2}. \textit{\tb{World-locked interaction entries:}} Microsoft HoloLens 2~\cite{sgprod2026_microsoft_microsoft_hololens_2}; Magic Leap 2~\cite{sgprod2026_magic_leap_magic_leap_2}; Meta Ray-Ban Display~\cite{sgprod2026_meta_ray_ban_meta_ray_ban_display} \\
\midrule
Cross-session world state and digital twins
& persistent scene graphs, change detection, map staleness estimation, state provenance, cross-session consistency, correction, and rollback
& \textit{\tb{World models and persistent spatial state:}} Pandora~\cite{Pandora}; SpatialWorld~\cite{SpatialWorld}; EgoForge~\cite{EgoForge}; latent spatial memory~\cite{LatentSpatialMemory}; Aria Digital Twin~\cite{Ariadigitaltwin}. \textit{\tb{Cross-time and cross-view memory proxies:}} EgoMemReason~\cite{EgoMemReason}; EGOSTREAM~\cite{EgoStream}; EgoExoMem~\cite{EgoExoMem}; PVCL~\cite{PersonalVisualContextLearning}; EgoLife~\cite{Egolife}
& \textit{\tb{Persistent sensing and on-device systems:}} Project Aria~\cite{ProjectAria}; Aria Gen 2~\cite{sgprod2026_meta_reality_labs_research_aria_gen_2}; EPIC~\cite{EPIC}; VisionClaw~\cite{VisionClaw}; OpenGlass~\cite{OpenGlass}; EgoKit~\cite{EgoKit}. \textit{\tb{Mapping backbone:}} ORB-SLAM3 multimap support~\cite{Orb-slam3} \\
\midrule
Spatial action and device or robot handoff
& actionable-region grounding, permission-governed IoT or robot commands, confirmation, execution monitoring, and safe recovery
& \textit{\tb{Spatial grounding and embodied proxies:}} OpenEQA~\cite{Openeqa}; REVERIE~\cite{Reverie}; SpatialWorld~\cite{SpatialWorld}. \textit{\tb{Digital action and tool-use proxies:}} Ego2Web~\cite{Ego2Web}; WebArena~\cite{Webarena}; SeeAct~\cite{Seeact}; Mobile-Agent~\cite{MobileAgent}. \textit{\tb{Robot-action proxy:}} Open X-Embodiment~\cite{Open-x-embodiment}; no unified smart-glasses handoff benchmark
& \textit{\tb{Wearable agent and communication systems:}} VisionClaw~\cite{VisionClaw}; Egocentric Co-Pilot~\cite{EgocentricCoPilot}; intention-aware semantic agent communication~\cite{IntentionAwareSemanticAgentCommunications}. \textit{\tb{AR and display entries:}} Microsoft HoloLens 2~\cite{sgprod2026_microsoft_microsoft_hololens_2}; Magic Leap 2~\cite{sgprod2026_magic_leap_magic_leap_2}; Snap Specs~\cite{sgprod2026_snap_snap_specs_2026}; Meta Ray-Ban Display~\cite{sgprod2026_meta_ray_ban_meta_ray_ban_display}. \textit{\tb{Robot-side systems:}} SayCan~\cite{SayCan}; RT-1~\cite{RT-1} \\
\midrule
\rowcolor{aprilblue!20}
\multicolumn{4}{l}{\textit{\textbf{Social Interaction and Collaboration}}} \\
\midrule
Meeting and dialogue mediation
& speaker-turn tracking, captioning, translation, query-focused summarization, misunderstanding detection, and appropriately timed repair
& \textit{\tb{Meeting and dialogue resources:}} AMI~\cite{AMI}; QMSum~\cite{QMSum}; ConversationBreakdowns~\cite{Conversational}. \textit{\tb{Speaker and interaction extensions:}} AVA-ActiveSpeaker~\cite{AvaActiveSpeaker}; H2HMem~\cite{H2HMem}; EgoLife~\cite{Egolife}; EgoSAT~\cite{EgoSAT}
& \textit{\tb{Caption and display entries:}} Even Realities G1~\cite{sgprod2026_even_realities_even_realities_g1}; Halliday DigiWindow Glasses~\cite{sgprod2026_halliday_halliday_digiwindow_glasses}; Vuzix Z100~\cite{sgprod2026_vuzix_vuzix_z100}; Meta Ray-Ban Display~\cite{sgprod2026_meta_ray_ban_meta_ray_ban_display}. \textit{\tb{Audio-first and agent systems:}} Solos AirGo V2~\cite{sgprod2026_solos_solos_airgo_v2}; Ray-Ban Meta Gen 2~\cite{sgprod2026_meta_ray_ban_ray_ban_meta_gen_2}; OpenGlass~\cite{OpenGlass}; VisionClaw~\cite{VisionClaw} \\
\midrule
Audio-visual social perception and inclusive participation
& active-speaker detection, social-reference grounding, communication access, mixed-ability coordination, and stakeholder-specific feedback
& \textit{\tb{Audio-visual and inclusive-interaction evaluation:}} AVA-ActiveSpeaker~\cite{AvaActiveSpeaker}; MixedVision~\cite{ReshapingInclusiveInterpersonalDynamics}; AR-DeafEducation~\cite{EvaluatingARForDeafStudents}; ConversationBreakdowns~\cite{Conversational}; H2HMem~\cite{H2HMem}. \textit{\tb{Social-use evidence:}} wearable-camera social acceptability~\cite{sgacad2019_socialacceptability}
& \textit{\tb{Assistive and display products:}} Envision Glasses~\cite{sgprod2026_envision_envision_glasses}; OrCam MyEye~\cite{sgprod2026_orcam_orcam_myeye_2_myeye_pro}; Even Realities G1~\cite{sgprod2026_even_realities_even_realities_g1}; Halliday DigiWindow Glasses~\cite{sgprod2026_halliday_halliday_digiwindow_glasses}; Meta Ray-Ban Display~\cite{sgprod2026_meta_ray_ban_meta_ray_ban_display}. \textit{\tb{Interaction-research platforms:}} Pupil Labs Neon~\cite{sgprod2026_pupil_labs_pupil_labs_neon}; Tobii Pro Glasses 3~\cite{sgprod2026_tobii_tobii_pro_glasses_3}; Microsoft HoloLens 2~\cite{sgprod2026_microsoft_microsoft_hololens_2} \\
\midrule
Remote collaboration and expert handoff
& shared first-person context, spatial annotation, role-aware guidance, expert-response management, weak-network recovery, and task-state handoff
& \textit{\tb{Interactive procedural resources:}} HoloAssist~\cite{Holoassist}; Assembly101~\cite{Assembly101}; Ego-Exo4D~\cite{Ego-Exo4D}. \textit{\tb{Proactive intervention and recovery:}} Plan-Watch-Recover/Pro2Bench~\cite{PlanWatchRecover}; EgoPro-Bench~\cite{EgoPro-Bench}; Streaming Interventions~\cite{StreamingInterventions}; IPIBench~\cite{IPIBench}
& \textit{\tb{Research assistance systems:}} Pro2Assist~\cite{Pro2Assist}; VisionClaw~\cite{VisionClaw}; OpenGlass~\cite{OpenGlass}. \textit{\tb{Enterprise and spatial-collaboration entries:}} RealWear Navigator 520~\cite{sgprod2026_realwear_realwear_navigator_520}; Vuzix M400~\cite{sgprod2026_vuzix_vuzix_m400}; Google Glass Enterprise Edition 2~\cite{sgprod2026_google_google_glass_enterprise_edition_2}; Microsoft HoloLens 2~\cite{sgprod2026_microsoft_microsoft_hololens_2}; Magic Leap 2~\cite{sgprod2026_magic_leap_magic_leap_2} \\
\midrule
Shared memory and role-aware coordination
& attribution of people, events, decisions, and commitments; permissioned group memory; correction; selective sharing; and cross-view consistency
& \textit{\tb{Human-human and cross-view memory:}} H2HMem~\cite{H2HMem}; EgoExoMem~\cite{EgoExoMem}; EgoMemReason~\cite{EgoMemReason}. \textit{\tb{Streaming and personalized memory:}} EGOSTREAM~\cite{EgoStream}; LightMem-Ego~\cite{LightMem-Ego}; PVCL~\cite{PersonalVisualContextLearning}; EgoLife~\cite{Egolife}; online episodic-memory QA~\cite{MultimodalLMMs-EpisodicMemory}; EgoSAT~\cite{EgoSAT}
& \textit{\tb{Always-on and personal-agent systems:}} VisionClaw~\cite{VisionClaw}; Egocentric Co-Pilot~\cite{EgocentricCoPilot}; OpenGlass~\cite{OpenGlass}. \textit{\tb{Collaborative hardware entries:}} RealWear Navigator 520~\cite{sgprod2026_realwear_realwear_navigator_520}; Ray-Ban Meta Gen 2~\cite{sgprod2026_meta_ray_ban_ray_ban_meta_gen_2}; Aria Gen 2~\cite{sgprod2026_meta_reality_labs_research_aria_gen_2}; Meta Ray-Ban Display~\cite{sgprod2026_meta_ray_ban_meta_ray_ban_display} \\
\midrule
Consent, privacy, and social acceptability
& recording-state legibility, contextual consent, data minimization, local redaction, memory deletion, incident review, and bystander control
& \textit{\tb{Privacy and social-acceptability studies:}} CameraGlassesPrivacy~\cite{CameraGlassesPrivacy}; BystanderPrivacy~\cite{sgacad2014_bystanderprivacy}; wearable-camera social acceptability~\cite{sgacad2019_socialacceptability}; MindTheGap~\cite{MindTheGap}; PrivacyUtility~\cite{Position}. \textit{\tb{Security and attack/defense evidence:}} visual jailbreaks~\cite{VisualAdversarialExamplesJailbreak}; PhySE~\cite{PhySE}; UNSEEN~\cite{UNSEEN}
& \textit{\tb{Privacy-control and local-processing systems:}} VisGuardian~\cite{VisGuardian}; OpenGlass~\cite{OpenGlass}. \textit{\tb{Camera-glasses and AR product routes for evaluating state legibility and control:}} Ray-Ban Stories~\cite{sgprod2026_meta_ray_ban_ray_ban_stories}; Ray-Ban Meta Gen 2~\cite{sgprod2026_meta_ray_ban_ray_ban_meta_gen_2}; Aria Gen 2~\cite{sgprod2026_meta_reality_labs_research_aria_gen_2}; Snap Specs~\cite{sgprod2026_snap_snap_specs_2026}; Meta Ray-Ban Display~\cite{sgprod2026_meta_ray_ban_meta_ray_ban_display} \\
\midrule
\rowcolor{aprilblue!20}
\multicolumn{4}{l}{\textit{\textbf{Embodied Intelligence}}} \\
\midrule
First-person demonstrations and multiview data infrastructure
& long-form capture, temporal organization, ego-exo synchronization, gaze and pose sensing, calibration, task outcome annotation, and data governance
& \textit{\tb{Egocentric and multiview datasets:}} Ego4D~\cite{Ego4D}; Ego-Exo4D~\cite{Ego-Exo4D}; HoloAssist~\cite{Holoassist}; Ego-1K~\cite{Ego-1K}; AoE~\cite{AoE}; Open-AoE~\cite{Open-AoE}; EPIC-KITCHENS-100~\cite{EK-100}; Assembly101~\cite{Assembly101}; IKEA-ASM~\cite{IKEA-ASM}
& \textit{\tb{Wearable sensing and collection platforms:}} Project Aria~\cite{ProjectAria}; Aria Gen 1~\cite{ProjectAria}; Aria Gen 2~\cite{sgprod2026_meta_reality_labs_research_aria_gen_2}; Pupil Labs Neon~\cite{sgprod2026_pupil_labs_pupil_labs_neon}; Tobii Pro Glasses 3~\cite{sgprod2026_tobii_tobii_pro_glasses_3}; EgoKit~\cite{EgoKit}. \textit{\tb{Supplementary multiview capture entries:}} GoPro Hero13 Black~\cite{sgprod2026_gopro_gopro_hero13_black}; GoPro Max 2~\cite{sgprod2026_gopro_gopro_max_2}; Insta360 Ace Pro 2~\cite{sgprod2026_insta360_insta360_ace_pro_2}; Insta360 X4~\cite{sgprod2026_insta360_insta360_x4} \\
\midrule
Active vision and embodied state estimation
& active-view selection, head and body pose, hand-object state, temporal segmentation, efficient on-device perception, and viewpoint uncertainty
& \textit{\tb{First-person state-estimation substrates:}} Ego4D~\cite{Ego4D}; Ego-Exo4D~\cite{Ego-Exo4D}; HoloAssist~\cite{Holoassist}; EGTEA Gaze+~\cite{sgacad2018_egtea_gaze}; EPIC-KITCHENS-100~\cite{EK-100}; Aria Digital Twin~\cite{Ariadigitaltwin}; Open-AoE~\cite{Open-AoE}
& \textit{\tb{Active-perception and embodied systems:}} ActiveGlasses~\cite{Activeglasses}; EgoMI~\cite{EgoMi}; ActiveMimic~\cite{ActiveMimic}; EPIC~\cite{EPIC}. \textit{\tb{Pose, VIO, and mapping backbones:}} LEVIO~\cite{LEVIO}; VINS-Mono~\cite{Vins-mono}; ORB-SLAM3~\cite{Orb-slam3}; DROID-SLAM~\cite{Droid-slam}; Project Aria~\cite{ProjectAria}; Aria Gen 2~\cite{sgprod2026_meta_reality_labs_research_aria_gen_2} \\
\midrule
Contact, affordance, and dexterous skill recovery
& active-object and state recovery, contact and affordance representation, grasp or pressure estimation, subgoal discovery, and dexterous retargeting
& \textit{\tb{Direct dexterity and contact resources:}} EgoDex~\cite{Egodex}; EgoTactile~\cite{EgoTactile}; EgoScale~\cite{EgoScale}. \textit{\tb{Interaction and procedural substrates:}} Open-AoE~\cite{Open-AoE}; Ego-Exo4D~\cite{Ego-Exo4D}; HoloAssist~\cite{Holoassist}; EPIC-KITCHENS-100~\cite{EK-100}; IKEA-ASM~\cite{IKEA-ASM}
& \textit{\tb{Retargeting and dexterous-learning systems:}} EgoEngine~\cite{EgoEngine}; UniDex~\cite{UniDex}; ActiveGlasses~\cite{Activeglasses}; EgoMI~\cite{EgoMi}; EgoVLA~\cite{EgoVLA}. \textit{\tb{Wearable sensing entries:}} Aria Gen 2~\cite{sgprod2026_meta_reality_labs_research_aria_gen_2}; Pupil Labs Neon~\cite{sgprod2026_pupil_labs_pupil_labs_neon} \\
\midrule
Human-video-to-robot policy transfer
& egocentric representation learning, action-space mapping, embodiment alignment, imitation or VLA learning, sample efficiency, and failure-aware adaptation
& \textit{\tb{Human-video data sources:}} Ego4D~\cite{Ego4D}; Ego-Exo4D~\cite{Ego-Exo4D}; AoE~\cite{AoE}; Open-AoE~\cite{Open-AoE}; HumanNet~\cite{Humannet}. \textit{\tb{Robot-side data and evaluation proxies:}} Open X-Embodiment~\cite{Open-x-embodiment}; DROID~\cite{Droid}; BridgeData V2~\cite{BridgeData}
& \textit{\tb{Imitation, zero-shot, and representation routes:}} EgoMimic~\cite{Egomimic}; EgoZero~\cite{Egozero}; HumanEgo~\cite{Humanego}; HumanNet~\cite{Humannet}; R3M~\cite{R3M}. \textit{\tb{VLA and dexterous-transfer systems:}} EgoVLA~\cite{EgoVLA}; EgoScale~\cite{EgoScale}; EgoEngine~\cite{EgoEngine}; UniDex~\cite{UniDex}; ActiveMimic~\cite{ActiveMimic} \\
\midrule
Robot-side validation, safety, and governance
& downstream task success, cross-embodiment failure analysis, unsafe-execution detection, target-robot safety validation, skill ownership, consent, and responsibility attribution
& \textit{\tb{Large-scale robot datasets and suites:}} Open X-Embodiment~\cite{Open-x-embodiment}; DROID~\cite{Droid}; BridgeData V2~\cite{BridgeData}. \textit{\tb{Human-to-robot transfer evaluation routes:}} EgoZero~\cite{Egozero}; EgoMimic~\cite{Egomimic}; HumanEgo~\cite{Humanego}; EgoVLA~\cite{EgoVLA}
& \textit{\tb{Robot policy and representation systems:}} RT-1~\cite{RT-1}; R3M~\cite{R3M}; SayCan~\cite{SayCan}; ImitDiff~\cite{dong2025imitdiff}; RT-X models associated with Open X-Embodiment~\cite{Open-x-embodiment}. \textit{\tb{Cross-embodiment validation pipelines:}} EgoZero~\cite{Egozero}; EgoMimic~\cite{Egomimic}; HumanEgo~\cite{Humanego}; EgoEngine~\cite{EgoEngine}; UniDex~\cite{UniDex}; target robot platforms used for downstream validation \\
\bottomrule
\end{tabular}
\end{adjustbox}
\end{table}

\subsection{Social Interaction and Collaboration}
\label{sec:social-collaboration}

Social interaction and collaboration center on shared viewpoints, dialogue, and task state across wearers, remote experts, peers, teachers, patients, customers, and bystanders. Smart glasses can mediate conversations through captions and translation, provide first-person telepresence, preserve decisions and commitments, and support role handoffs during collaborative work. Unlike individual assistance, success depends on what multiple stakeholders understand, what information each is permitted to access, and whether the device's sensing and recording state is legible to people who are not wearing it. The capability loop consequently progresses from dialogue mediation, to multi-person perception and inclusive participation, to remote task coordination, shared memory, and explicit privacy governance.

\sgpoint{Meeting and dialogue mediation}{
In meetings and everyday conversations, glasses can provide captions, translation, speaker-turn cues, query-focused summaries, and reminders of unresolved points. AMI~\cite{AMI} and QMSum~\cite{QMSum} provide meeting and summarization resources, while ConversationBreakdowns~\cite{Conversational} examines conversational successes, misunderstandings, interruption, and repair in everyday smart-glasses use. The wearable setting changes the interaction cost: feedback that is technically correct may still be harmful if it arrives after the relevant turn, obscures eye contact, interrupts the speaker, or leaks private audio. Speaker attribution and temporal grounding are especially important when summaries or commitments are stored. Evaluation should therefore include caption latency, attribution, repair timing, interruption burden, summary faithfulness, and the ability of participants to correct the record. HUD-oriented products can reduce the need to look away from collaborators, but their limited display area makes prioritization and concise presentation essential.
}

\sgpoint{Audio-visual social perception and inclusive participation}{
Collaboration also requires understanding who is speaking, where attention is directed, and how participants with different perceptual abilities share information. AVA-ActiveSpeaker~\cite{AvaActiveSpeaker} provides an audio-visual active-speaker reference, MixedVision~\cite{ReshapingInclusiveInterpersonalDynamics} studies mixed-vision social activities, and AR-DeafEducation~\cite{EvaluatingARForDeafStudents} evaluates communication access for deaf students in experiential education. These works motivate speaker highlighting, visual descriptions, shared annotations, and modality adaptation. However, the appropriate outcome is stakeholder-specific: a feature that helps the wearer may increase distraction for a collaborator or reveal information about a bystander. Evaluations should measure participation balance, task contribution, communication repair, workload, and collaborator outcomes across different groups rather than only the wearer's task success. Personalization should remain visible and controllable so that the system does not silently infer ability, role, or social intent.
}

\sgpoint{Remote collaboration, shared viewpoints, and expert handoff}{
First-person streaming can allow a remote expert to observe the work context, point to relevant regions, and guide a wearer through a procedure. HoloAssist~\cite{Holoassist}, Pro2Assist~\cite{Pro2Assist}, and Plan-Watch-Recover/Pro2Bench~\cite{PlanWatchRecover} provide research routes for interactive and proactive procedural assistance, while RealWear Navigator 520~\cite{sgprod2026_realwear_realwear_navigator_520}, Vuzix M400~\cite{sgprod2026_vuzix_vuzix_m400}, and Google Glass Enterprise Edition 2~\cite{sgprod2026_google_google_glass_enterprise_edition_2} represent enterprise entry points. The system must maintain a shared task state despite occlusion, network loss, expert delay, or a handoff between personnel. Role permissions should determine who can view the stream, annotate, issue instructions, or approve an action. Relevant outcomes include expert-response latency, handoff failures, recovery after disconnection, annotation grounding, log integrity, and downstream task performance. Emergency or safety-critical collaboration requires a higher threshold than routine meeting support because delay or role confusion can directly affect operational decisions.
}

\sgpoint{Shared memory and role-aware coordination}{
Multi-party collaboration becomes stateful when the glasses remember people, prior interactions, decisions, commitments, and unresolved tasks. H2HMem~\cite{H2HMem} targets multimodal memory in human-human interactions, EgoExoMem~\cite{EgoExoMem} studies cross-view memory reasoning, and EgoMemReason~\cite{EgoMemReason} and PVCL~\cite{PersonalVisualContextLearning} provide complementary long-horizon and personalized-context routes. A social memory should distinguish what the wearer personally observed from what another participant said, what the system inferred, and what was later corrected. It should also enforce role- and relationship-specific access: a private reminder, a team decision, and a patient-clinician interaction cannot share the same retention and disclosure policy. Evaluation should include attribution accuracy, cross-view consistency, permission errors, correction propagation, selective sharing, and deletion. Without these controls, persistent assistance can amplify misremembered statements or expose information far beyond the original interaction.
}

\sgpoint{Consent, bystander privacy, and longitudinal social acceptance}{
Camera glasses affect people who may never interact with the system directly. CameraGlassesPrivacy~\cite{CameraGlassesPrivacy}, BystanderPrivacy~\cite{sgacad2014_bystanderprivacy}, and social-acceptability studies~\cite{sgacad2019_socialacceptability} examine recording concerns and device acceptance; MindTheGap~\cite{MindTheGap} maps wearer-bystander privacy tensions; PrivacyUtility~\cite{Position} frames the privacy-utility trade-off in life-logging streams; and VisGuardian~\cite{VisGuardian} explores local group-based privacy control. Deployment therefore requires legible recording state, contextual consent, data minimization, local redaction, memory deletion, and incident review. Aggregate task success cannot replace separate measures for wearers, collaborators, and bystanders, who may bear different benefits and risks. Social acceptance should be studied longitudinally in homes, workplaces, classrooms, and public spaces, because novelty effects and one-time consent do not establish durable legitimacy within ongoing relationships.
}

\subsection{Spatial Intelligence}
\label{sec:spatial-services}

Spatial intelligence turns smart glasses from a momentary perception interface into a persistent service that maintains, queries, and updates representations of objects, places, paths, and actionable regions. It covers indoor wayfinding, object finding, spatial reminders, home or workplace digital twins, museum and retail guidance, shared-space annotation, and handoff to Internet of Things (IoT) devices or robots. This scene operationalizes the foundational spatial-state capability introduced in \cref{sec:spatial-state}: geometry, semantics, localization, memory, and action must remain mutually consistent across motion and across sessions. Because stale maps or incorrect references can propagate into later guidance or actions, spatial intelligence is best analyzed as a sequence from map construction, to spatial query and reference, to persistent world-state maintenance, and finally to governed action.

\sgpoint{Spatial reconstruction and semantic mapping}{
The first layer is a calibrated representation of the environment. Smart glasses must combine first-person cameras, inertial measurements, and device calibration to estimate pose, reconstruct geometry, and attach semantic labels to objects, surfaces, and places. Aria Digital Twin~\cite{Ariadigitaltwin} and Project Aria~\cite{ProjectAria} provide egocentric capture and 3D machine-perception resources, while Matterport3D~\cite{Matterport3d} and ScanNet~\cite{Scannet} offer indoor reconstruction references. VINS-Mono~\cite{Vins-mono}, ORB-SLAM3~\cite{Orb-slam3}, DROID-SLAM~\cite{Droid-slam}, and LEVIO~\cite{LEVIO} illustrate visual-inertial or visual localization routes relevant to resource-constrained devices. A wearable map, however, must be evaluated beyond offline reconstruction quality: calibration drift, relocalization after device removal, dynamic objects, partial coverage, sensing duty cycle, and map timestamps determine whether the representation remains usable. Semantic state should preserve provenance and uncertainty so that a user can distinguish an observed fact from an inferred or stale one.
}

\sgpoint{Navigation and spatial question answering}{
Once localized, the glasses can provide route guidance and answer questions such as where an object is, which entrance is accessible, or how to reach a destination. Habitat~\cite{Habitat}, R2R~\cite{VLN}, REVERIE~\cite{Reverie}, EmbodiedQA~\cite{EmbodiedQA}, and OpenEQA~\cite{Openeqa} cover navigation, remote-object reference, instruction following, and embodied question answering. Translating these tasks to glasses requires accounting for head motion, wearer intent, display or audio bandwidth, and the fact that the user executes the movement rather than an autonomous agent. Camera/audio-first profiles can provide short-horizon spoken guidance, camera-and-display or HUD profiles can reduce the cost of confirmation, and true-AR devices can place world-locked cues. Evaluation should jointly measure localization error, wrong turns, cue timing, attentional load, safe recovery, \etc A route-completion score alone cannot reveal whether the guidance was late, visually distracting, or based on an outdated map.
}

\sgpoint{Deictic reference, gaze, and near-body relations}{
Everyday spatial interaction often relies on expressions such as ``this,'' ``that one,'' ``over there,'' or ``the object beside my hand.'' PointingMLLM/EgoPoint-Bench~\cite{MLLMs-Pointing} evaluates egocentric pointing and referential reasoning, EgoProx~\cite{EgoProx} targets 3D proximity reasoning, and EGTEA Gaze+~\cite{sgacad2018_egtea_gaze} links gaze and first-person actions. Research devices such as Pupil Labs Neon~\cite{sgprod2026_pupil_labs_pupil_labs_neon}, Tobii Pro Glasses 3~\cite{sgprod2026_tobii_tobii_pro_glasses_3}, and Aria Gen 2~\cite{sgprod2026_meta_reality_labs_research_aria_gen_2} provide sensing routes for gaze, pose, and synchronized observations. These signals can reduce language burden, but they also introduce device-induced gaze error, calibration changes, and ambiguity among nearby objects. A robust system should expose competing referents, ask for confirmation when ambiguity is consequential, and avoid treating gaze as equivalent to intention. Referential hallucination, distance error, confirmation cost, and the effect of head or eye-tracking uncertainty should therefore be evaluated together.
}

\sgpoint{Cross-session world state and digital twins}{
Persistent spatial services must answer not only where something is, but whether the remembered state is still valid. Pandora~\cite{Pandora} represents articulated 3D scene graphs from egocentric vision, SpatialWorld~\cite{SpatialWorld} benchmarks interactive spatial reasoning, EgoForge~\cite{EgoForge} explores goal-directed egocentric world simulation, and latent spatial memory~\cite{LatentSpatialMemory} studies persistent representations for video world models. These directions support object histories, room- or workspace-level state, and reasoning across visits. The main deployment difficulty is change: objects move, spaces are rearranged, permissions differ by user, and an old map may be locally accurate but operationally wrong. The system therefore needs map-staleness detection, cross-session relocalization, state timestamps, confidence, user correction, and rollback. Shared homes and workplaces additionally require ownership and access rules for spatial memories, because a world model can reveal sensitive routines and object locations even when no raw video is retained.
}

\sgpoint{Spatial action interfaces and end-to-end validation}{
The final step connects spatial state to actions: opening a device interface, placing a world-locked annotation, handing a target location to a robot, or triggering an IoT service. True-AR and display-oriented products such as Microsoft HoloLens 2~\cite{sgprod2026_microsoft_microsoft_hololens_2}, Magic Leap 2~\cite{sgprod2026_magic_leap_magic_leap_2}, Snap Specs~\cite{sgprod2026_snap_snap_specs_2026}, Meta Ray-Ban Display~\cite{sgprod2026_meta_ray_ban_meta_ray_ban_display}, RayNeo X3 Pro~\cite{sgprod2026_tcl_rayneo_rayneo_x3_pro}, and XREAL AURA~\cite{sgprod2026_xreal_google_xreal_aura_project_aura} provide alternative interface routes. An L4 spatial action should specify the target, coordinate frame, confidence, permission, and expected effect, and it should verify whether execution succeeded. Tourism, museums, retail spaces, homes, and workplaces also require current venue maps, content ownership, and public-space recording policies. No cited benchmark yet integrates cross-day mapping, deictic grounding, world-locked feedback, user correction, and device or robot execution, so current results should be interpreted as component evidence rather than validation of a complete persistent spatial service.
}

\subsection{Embodied Intelligence}
\label{sec:embodied-ai-app}

Embodied intelligence positions smart glasses not only as wearer-facing assistants, but also as interfaces through which human experience can be captured, structured, and transferred to robotic systems. The glasses can record demonstrations, task narratives, failures and recoveries, gaze, body and hand motion, hand-object relations, and spatial pose, while also providing feedback during data collection or robot supervision. This scene therefore spans a longer chain than ordinary assistance: sensing and data quality, embodied state recovery, contact and affordance representation, human-to-robot mapping, policy learning, and downstream robot validation. The evidential boundary must remain explicit: high-quality human data support an \textit{L5 data} or \textit{partial L5} claim, whereas an \textit{L5 system} claim requires successful and safe behavior on the target robot.


\begin{wrapfigure}{r}{0.7\linewidth}
    \centering
    \vspace{-1.25em}
    \includegraphics[width=1\linewidth]{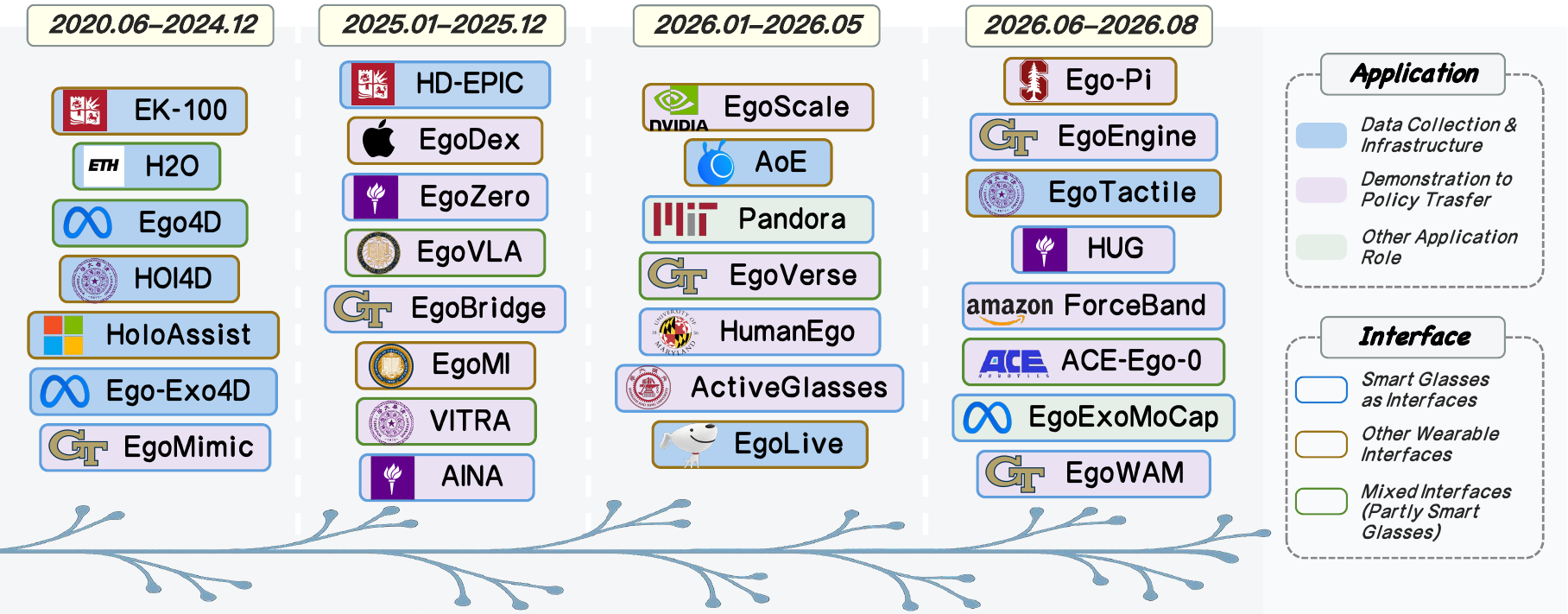}
    \caption{\textbf{Temporal evolution and taxonomy of representative work on smart glasses and related wearable interfaces for embodied intelligence.}
    As detailed in \cref{sec:embodied-ai-app}, the branching timeline organizes representative work from 2020.06 to 2026.08 into three application roles using distinct fill colors, while outline colors indicate the corresponding data-interface categories.}
    \label{fig:timeline_embodied}
    \vspace{-0.55em}
\end{wrapfigure}

\sgpoint{First-person demonstrations and multiview data infrastructure}{
Ego4D~\cite{Ego4D}, Ego-Exo4D~\cite{Ego-Exo4D}, HD-EPIC~\cite{HD-EPIC}, and Open-AoE~\cite{Open-AoE} provide complementary routes for long-form activity capture, procedural interaction, multiview alignment, and always-on or open egocentric data collection. Aria Gen 2~\cite{sgprod2026_meta_reality_labs_research_aria_gen_2}, Pupil Labs Neon~\cite{sgprod2026_pupil_labs_pupil_labs_neon}, Tobii Pro Glasses 3~\cite{sgprod2026_tobii_tobii_pro_glasses_3}, and EgoKit~\cite{EgoKit} expose or unify sensing interfaces for video, pose, gaze, or heterogeneous capture. A useful data pipeline must synchronize devices, calibrate coordinate frames, segment long activities, link actions to object-state changes and outcomes, and quantify usable data per hour rather than raw recording duration. It must also preserve consent coverage, de-identification, workspace privacy, and ownership of demonstrated skills. These requirements determine whether human experience can serve as reliable training data before any claim about robot transfer is made.
}

\sgpoint{Active vision and embodied state estimation}{
Human head motion is not only noise; it reflects active information seeking and can reveal which viewpoints are useful for manipulation. ActiveGlasses~\cite{Activeglasses} and EgoMI~\cite{EgoMi} study active vision and manipulation from egocentric demonstrations, while ActiveMimic~\cite{ActiveMimic} explores egocentric pretraining with active perception. EPIC~\cite{EPIC} and LEVIO~\cite{LEVIO} provide complementary system and lightweight visual-inertial directions for resource-constrained embodied glasses. The system must recover head and body pose, hands, active objects, object states, and temporal subgoals while accounting for rapid camera motion and self-occlusion. It must also distinguish informative viewpoint changes from incidental motion. Evaluation should include calibration and synchronization, state-estimation accuracy, active-view utility, missing observations, computational latency, and robustness across wearers and device placements. These representations form the bridge between raw video and a task state that can be mapped to a robot.
}

\sgpoint{Contact, affordance, and dexterous skill recovery}{
Robot learning requires more than recognizing visible actions. It must infer which object is active, where contact occurs, how the object state changes, what affordances are relevant, and which subgoal the human is pursuing. EgoDex~\cite{Egodex} targets dexterous manipulation from large-scale egocentric video, EgoScale~\cite{EgoScale}, EgoEngine~\cite{EgoEngine}, and UniDex~\cite{UniDex} extend the pipeline toward high-fidelity or universal dexterous robot demonstrations and control, EgoTactile~\cite{EgoTactile} further studies grasp-pressure inference, and ForceBand~\cite{ForceBand} targets the acquisition of forceful manipulation skills that are difficult to recover from egocentric vision alone. The principal limitation is that ordinary smart glasses do not directly observe force, tactile feedback, joint torque, or full proprioception. Visual estimates of contact or pressure should therefore carry uncertainty and should be validated against downstream physical interaction. Retargeting must also account for differences between human hands and robot grippers, reachability, kinematics, and safety constraints rather than assuming that visual similarity implies executable equivalence.
}

\sgpoint{Human-to-robot representation and policy transfer}{
EgoMimic~\cite{Egomimic}, EgoZero~\cite{Egozero}, EgoBridge~\cite{EgoBridge}, VITRA~\cite{VITRA}, HUG~\cite{HUG}, HumanEgo~\cite{Humanego}, HumanNet~\cite{Humannet}, EgoVLA~\cite{EgoVLA}, and EgoWAM~\cite{EgoWAM} explore complementary routes from human egocentric experience to imitation, zero-shot transfer, large-scale human-video representation learning, and Vision-Language-Action (VLA) policies. A credible transfer pipeline must align human and robot viewpoints, convert human motion or task state into the robot's action space, identify invariant subgoals and affordances, and adapt to different embodiments and environments. Evaluation should isolate representation quality, sample efficiency, transfer success, and failure type rather than reporting only pretraining loss or human-video understanding. Cross-embodiment errors are especially important: the model may correctly understand what the human did yet produce an unreachable, unstable, or unsafe robot action. Downstream robot experiments are therefore necessary to determine which information in smart-glasses data is genuinely actionable.
}

\sgpoint{Robot-side validation, safety, and governance}{
Open X-Embodiment~\cite{Open-x-embodiment}, DROID~\cite{Droid}, and BridgeData V2~\cite{BridgeData} provide robot-side data references, while RT-1~\cite{RT-1}, R3M~\cite{R3M}, and SayCan~\cite{SayCan} illustrate policy, representation, and affordance-grounded control routes against which transfer can be assessed. ImitDiff provides a complementary robot-side robustness reference: it transfers vision-language foundation-model priors into pixel-level task semantics, combines global and local visual evidence through a dual-resolution policy, and evaluates real-time visuomotor control under increased scene complexity, visual distractions, and novel objects~\cite{dong2025imitdiff}. An L5 system claim should report target-robot task success, sample efficiency, cross-embodiment failure rate, failure severity, unsafe execution, recovery, and safety validation under the intended operating conditions. Responsibility must also be traced across the human demonstrator, data curator, model developer, and robot operator. Demonstrations collected in homes, factories, kitchens, and laboratories may reveal private spaces or proprietary skills, so consent, de-identification, data-use boundaries, and skill ownership are part of the evaluation rather than external administrative details. 
An L5 system claim is justified only when the complete chain from human capture to safe robot outcome is substantiated.
}

\subsection{Other Potential Application Scenes}
\label{sec:other-scenes}

Beyond the nine major scenes above, smart glasses provide plausible entry points for a broader set of emerging applications whose evidence is currently distributed across products, prototypes, and neighboring research traditions. These include tourism and cultural-heritage interpretation, museum and exhibition guidance, location-aware retail and commerce, sports performance and first-person coaching, creative capture and live content production, entertainment and shared augmented-reality experiences, public-safety and emergency response, scientific fieldwork and laboratory observation, agriculture and outdoor field service, logistics and delivery coordination, smart-home and ambient IoT control, personal authentication and context-aware security, communication-efficient cooperation among wearable or embodied agents, \etc Sports-oriented products such as Oakley Meta Vanguard~\cite{sgprod2026_meta_oakley_oakley_meta_vanguard}, display and AR routes such as Meta Ray-Ban Display~\cite{sgprod2026_meta_ray_ban_meta_ray_ban_display}, Snap Specs~\cite{sgprod2026_snap_snap_specs_2026}, RayNeo X3 Pro~\cite{sgprod2026_tcl_rayneo_rayneo_x3_pro}, and XREAL AURA~\cite{sgprod2026_xreal_google_xreal_aura_project_aura}, enterprise devices such as RealWear Navigator 520~\cite{sgprod2026_realwear_realwear_navigator_520}, and research platforms such as Project Aria~\cite{ProjectAria} and Pupil Labs Neon~\cite{sgprod2026_pupil_labs_pupil_labs_neon} illustrate relevant hardware entry points. Intention-aware semantic agent communication for AI glasses~\cite{IntentionAwareSemanticAgentCommunications} suggests an additional direction in which glasses coordinate information exchange with other agents. At the same time, visual jailbreaks~\cite{VisualAdversarialExamplesJailbreak}, real-time AR-LLM social-engineering attacks~\cite{PhySE}, and corresponding unlearning defenses~\cite{UNSEEN} indicate that security and trust may themselves become application-defining requirements. These scenes warrant separate treatment once representative tasks, stakeholders, field outcomes, and end-to-end benchmarks become sufficiently coherent.

Across these application scenes, hardware profiles constrain what can be observed, computed, remembered, and communicated to the wearer. State persistence determines how spatial and memory errors accumulate over time; action consequences and stakeholder structure determine validation and governance thresholds; and the available evidence limits how far conclusions can be extrapolated beyond the evaluated setting. \textbf{No foundational capability admits a universal pass criterion: as task risk, state duration, action authority, or the number of affected stakeholders increases, evaluation must progress from local benchmark performance toward end-to-end field evidence, longitudinal outcomes, and appropriate independent auditing.}

%% file: sec/05_design_framework.tex
\section{Design Framework and Evaluation}
\label{sec:design}

The nine application domains characterized in \cref{sec:applications} differ in user activity, temporal urgency, action consequence, and participant structure. These differences imply that an intelligent-glasses system cannot be assessed solely through a list of functions or isolated model scores. This section therefore translates application-level requirements into a deployment-oriented design framework and a standardized evaluation protocol. The central objective is to determine how hardware, runtime, models, persistent state, interaction, external action, and governance jointly delimit a defensible capability claim, and what evidence is required to support that claim under realistic operating conditions.

\textbf{Overall design framework.}
We take the closed-loop embodied assistance process, rather than the individual model, as the primary design object. A deployable smart glass system must repeatedly decide what to sense, which observations are sufficiently reliable for reasoning, where computation should occur, what state may persist across time, when assistance should be delivered, whether an external action is authorized, and how the outcome should be verified and audited. 
The resulting closed-loop workflow follows this sequence: \emph{perception $\rightarrow$ inference $\rightarrow$ state $\rightarrow$ feedback/action $\rightarrow$ verification and recovery}. This loop must operate strictly within well-defined hardware profiles, operating-condition envelopes, and risk boundaries. 
User correction and override, intelligible device-state signaling, bystander awareness, organizational authorization, provenance, and failure recovery are therefore part of the capability itself rather than auxiliary product features. 
This perspective also makes runtime placement explicit. On-device processors, companion phones or pucks, cloud models, local enterprise servers, and external tools are distinct runtime nodes whose placement changes latency, data exposure, state consistency, failure responsibility, and the set of functions available during degraded connectivity (\cref{fig:framework}, \textit{bottom-right}). Accordingly, we organize the design space into \textbf{nine coupled dimensions} as follows. 
For each dimension, the discussion identifies the design object, scenario-dependent constraints, characteristic failure propagation paths, and admissible validation evidence. No single dimension is sufficient in isolation: a strong model cannot compensate for missing sensors, stale state, delayed feedback, unauthorized execution, or an unreproducible runtime.

\subsection{Hardware Profile and Form-Factor Design}
\label{sec:form-factor}

The hardware profile defines the physical envelope within which every higher-level capability must operate. It determines which parts of the wearer's environment can be observed, which modalities can be fused, how assistance can be returned, and whether sensing and interaction can be sustained for minutes, hours, or an entire workday. Form-factor design must therefore jointly consider sensing geometry, output bandwidth, compute and communication placement, power capacity, thermal dissipation, recording controls, facial fit, weight distribution, aesthetic integration, and long-term comfort rather than treating camera resolution, display field of view, or battery capacity as independent specifications.

\textbf{Sensing-feedback envelope.}
Sensor count alone does not establish useful first-person perception. Camera placement and field of view determine point-of-view alignment and hand-object visibility; microphone geometry affects speaker separation and robustness to environmental noise; IMU, gaze, depth, and pose sensors determine whether observations can be registered in a persistent spatial frame; and calibration and timestamp integrity determine whether these streams can be fused. Research platforms such as Project Aria and Aria Gen~2 emphasize synchronized multimodal sensing, calibration, gaze, and pose access, while Pupil Labs Neon and Tobii Pro Glasses~3 emphasize wearable gaze measurement and research-oriented data access~\cite{ProjectAria,sgprod2026_meta_reality_labs_research_aria_gen_2,sgprod2026_pupil_labs_pupil_labs_neon,sgprod2026_tobii_tobii_pro_glasses_3}. Output configuration is equally consequential: open-ear audio supports unobtrusive assistance but can be masked or leak to bystanders, whereas HUD or AR displays enable visual confirmation and world-referenced cues but impose optical, brightness, field-of-view, and thermal constraints.

\textbf{Representative form-factor routes.}
Existing devices allocate these budgets through several recurring routes. \textbf{\textit{1) Camera/audio-first glasses.}} These systems prioritize familiar appearance, everyday wearability, first-person capture, and voice interaction, but provide limited visual confirmation or spatially anchored feedback; Ray-Ban Meta Gen~2 and Rokid AI Glasses Style are representative references~\cite{sgprod2026_meta_ray_ban_ray_ban_meta_gen_2,sgprod2026_rokid_rokid_ai_glasses_style}. \textbf{\textit{2) Camera-free HUD glasses.}} These systems reduce recording ambiguity and sensing overhead while supporting captions, notifications, and lightweight prompts, as exemplified by Even Realities G1; without direct scene sensing, however, grounding must be supplied by another device or service~\cite{sgprod2026_even_realities_even_realities_g1}. \textbf{\textit{3) Camera-and-display glasses.}} These systems combine egocentric observation with immediate visual confirmation, but tighten the joint budget for weight, battery, brightness, and heat, as illustrated by Meta Ray-Ban Display~\cite{sgprod2026_meta_ray_ban_meta_ray_ban_display}. \textbf{\textit{4) True-AR systems.}} These systems add spatial anchoring, richer hand or pose interaction, and broader developer interfaces at the cost of more complex optics, compute, power, and ecosystem requirements; Snap Specs and established optical see-through platforms provide corresponding references~\cite{sgprod2026_snap_snap_specs_2026,sgprod2026_magic_leap_magic_leap_2}. \textbf{\textit{5) Research-sensing platforms.}} These devices prioritize synchronized raw sensing, calibration, gaze, and pose over consumer-facing feedback or all-day wearability, as represented by Aria Gen~2, Pupil Labs Neon, and Tobii Pro Glasses~3~\cite{sgprod2026_meta_reality_labs_research_aria_gen_2,sgprod2026_pupil_labs_pupil_labs_neon,sgprod2026_tobii_tobii_pro_glasses_3}.

\textbf{Wearability as a sustained operating condition.}
A nominally capable device may fail as an intelligent-glasses platform if users cannot wear it long enough to obtain continuous context. Total mass, front-back balance, temple pressure, skin-contact temperature, prescription-lens compatibility, visual obstruction, audio leakage, and appearance all affect adherence and data continuity. Social acceptability is also part of the effective hardware envelope: a visible camera, an ambiguous recording indicator, or a form factor perceived as intrusive can change the behavior of both the wearer and surrounding participants, thereby altering the very data and interactions being evaluated~\cite{sgacad2019_socialacceptability,CameraGlassesPrivacy,sgacad2014_bystanderprivacy}. Consequently, short laboratory trials should not be interpreted as evidence of stable all-day operation.

\textbf{Task-conditioned selection and evidence.}
There is no universally optimal form factor. A navigation aid may require reliable localization and an always-visible output channel; a memory assistant may prioritize unobtrusive capture and endurance; industrial guidance may tolerate a heavier device in exchange for a bright display, robust controls, and local enterprise connectivity; and an embodied-data platform may prioritize synchronization and calibration over visual feedback. Hardware claims should therefore be reported as a category-normalized profile that includes sensing and output configuration, point-of-view alignment, calibration, weight and fit, p50/p95 power draw, thermal behavior, recording-state signaling, and longitudinal adherence. Marketing specifications can establish nominal component availability, but sustained sensing, comfort, and end-to-end capability require laboratory measurement and field evidence.

\subsection{Runtime and Resource Management}
\label{sec:low-latency}

Within the physical envelope established by the hardware profile, runtime design determines how sensing, computation, state, and external services are distributed over time. Common topologies include glasses-phone-cloud, glasses-puck-cloud, glasses-local-server, and primarily on-device execution with selective offloading. These alternatives should not be reduced to a simple edge-versus-cloud choice: each topology creates different data-flow, trust, latency, availability, energy, and accountability boundaries.

\textbf{Placement as a data and responsibility boundary.}
A deployable runtime profile should specify which raw or processed streams leave the glasses, where redaction occurs, which node hosts working and persistent state, how caches are synchronized, where model and tool credentials reside, and which component owns timeout handling and recovery. Local preprocessing can reduce bandwidth and raw-data exposure, whereas cloud inference can provide larger models and broader tools; local enterprise servers may satisfy organizational control requirements but introduce their own synchronization and availability constraints. State versioning is especially important when perception, memory, and action are split across devices: an external tool must not act on a stale map, an obsolete user correction, or a cache that differs from the state displayed to the wearer.

\textbf{Latency as temporal validity rather than a single speed score.}
Responsiveness should be measured from observation capture to the first useful feedback and, where applicable, to externally visible action. Mean model inference time omits sensing delay, buffering, uplink and downlink variability, context construction, tool invocation, rendering, and retries. Moreover, the same delay has different consequences for casual question answering, live captioning, navigation, hazard alerts, step-by-step assistance, and physical execution. Evaluation should therefore report p50/p95/p99 capture-to-feedback and capture-to-action latency, timeout frequency, jitter, recovery time, and the fraction of outputs delivered inside the validity window of the supporting evidence. Streaming benchmarks that emphasize temporally valid recall and interventions that correct errors as they unfold reinforce the need to couple latency with evidence freshness rather than report speed in isolation~\cite{EgoStream,StreamingInterventions,IPIBench}.

\textbf{Sustainable duty cycle and graceful degradation.}
Always-on operation is bounded by battery drain, thermal derating, memory and context growth, network usage, and competition among sensing, inference, display, and communication. EPIC, LEVIO, EgoKit, and OpenGlass illustrate complementary approaches to efficient egocentric perception, embedded visual-inertial odometry, heterogeneous capture infrastructure, and sensing-computing split architectures~\cite{EPIC,LEVIO,EgoKit,OpenGlass}. Edge-oriented episodic-memory studies further show that online multimodal reasoning must be evaluated under device-level resource constraints rather than only through server-side accuracy~\cite{MultimodalLMMs-EpisodicMemory}. A robust system should define explicit degradation modes. Examples include reducing sampling rate, disabling computationally expensive reasoning modules, preserving safety-critical local functions, queuing non-urgent requests, or switching to phone-only feedback, with all such state transitions made visible to the end user. Energy per successful task, on-device processing ratio, offline task retention, thermal throttling frequency, and post-degradation task success jointly characterize sustainable operation.

\subsection{Perception and Inference Stack}
\label{sec:foundation-models}

The perception and inference stack converts continuous first-person streams into evidence that can support immediate feedback, persistent state, or external action. Its central design challenge is to reconcile the low cost required for continuous operation with the open-vocabulary and long-context reasoning required by unconstrained real-world tasks. This motivates a layered architecture rather than continuous invocation of a single large multimodal model.

\textbf{Always-on perceptual front end.}
Lightweight modules should continuously handle functions whose latency, energy, or safety requirements are incompatible with repeated large-model calls, including wake-word detection, ASR, OCR, object and hand cues, privacy redaction, motion estimation, SLAM/VIO, and basic safety filtering. GLIMPSE exemplifies real-time text recognition and contextual understanding for wearable VQA, while VINS-Mono, ORB-SLAM3, and LEVIO provide relevant references for visual-inertial state estimation under different computational assumptions~\cite{GLIMPSE,Vins-mono,Orb-slam3,LEVIO}. These modules also serve as triggers and compressors: rather than forwarding every frame, they can identify state changes, active objects, speech segments, low-confidence intervals, or task-relevant clips for deeper reasoning.

\textbf{Selective multimodal escalation.}
Open-vocabulary scene understanding, cross-modal disambiguation, video reasoning, and complex instruction following should be invoked over selected spatiotemporal evidence with an explicit context-construction policy. AdaVideoRAG operationalizes this principle for long-video understanding by using a lightweight intent classifier to route queries from direct inference through naive retrieval to graph-based retrieval, supported by complementary caption, ASR, OCR, visual, and graph indexes~\cite{xue2025adavideorag}. SuperGlasses, WearVQA, SAW-Bench, and EgoSAT collectively expose the breadth of intelligent-glasses reasoning, authentic wearable VQA, situated awareness, and continuous ego-view interaction that such models must support~\cite{Superglasses,WearVQA,SAW-Bench,EgoSAT}. Their use also highlights that benchmark performance depends on input resolution, frame sampling, audio availability, temporal context, prompt construction, and model/API version. A capability claim should therefore specify the exact perception-to-model interface rather than attribute the result to the model name alone.

\textbf{Grounding, reference resolution, and temporal validity.}
First-person interaction contains frequent deictic expressions such as ``this,'' ``that,'' ``over there,'' or ``the one I used earlier.'' These references may depend on gaze, pointing, hand-object contact, prior dialogue, and a changing spatial frame. EgoPoint-Bench reveals that apparently fluent multimodal models can still hallucinate referents in egocentric pointing, while EgoProx and SAW-Bench emphasize observer-centric spatial and situated reasoning~\cite{MLLMs-Pointing,EgoProx,SAW-Bench}. The stack should therefore represent what evidence supports each referent, detect unresolved or conflicting modalities, and distinguish a currently visible object from one inferred from memory. Time stamps, observation age, and state version should accompany the output whenever freshness affects its validity.

\textbf{Admission gates to feedback, memory, and action.}
Model output is not automatically admissible as persistent state or executable intent. Before crossing these boundaries, the system should apply source attribution, cross-modal consistency checks, uncertainty calibration, grounding verification, abstention, clarification, and task-specific fallback. A failure to read text, a failure to see the target, an ambiguous referent, a stale observation, and an unavailable external tool require different user messages and recovery strategies. Evaluation should therefore measure not only answer accuracy, but also grounding accuracy, faithfulness, attribution quality, modality-conflict handling, appropriate abstention, temporal consistency, and the downstream consequences of incorrect admission.

\subsection{Persistent State and Memory Lifecycle}
\label{sec:memory-design}

Intelligent glasses become longitudinal assistants only when information can persist across moments, sessions, and locations. Persistence, however, changes an ephemeral inference into a durable system claim that may later influence feedback or action. Memory must therefore be designed as a controlled lifecycle with typed state, provenance, validity, correction, access, and deletion policies rather than as an unrestricted vector store.

\textbf{Typed memory hierarchy.}
Working context, episodic memory, semantic memory, persistent spatial state, preference memory, user-confirmed facts, and action logs serve different purposes and should have different schemas and retention policies. EgoLife and LightMem-Ego explore everyday egocentric assistance and personal memory, while EgoMemReason and EGOSTREAM focus on long-horizon and streaming episodic reasoning~\cite{Egolife,LightMem-Ego,EgoMemReason,EgoStream}. A practical memory entry should associate content with temporal and spatial context, sensor or tool provenance, model and pipeline version, confidence, confirmation status, access scope, and expiration policy. This representation preserves the distinction among direct observation, model-generated inference, imported external information, and an explicitly confirmed user fact.

\textbf{Provenance-aware write admission.}
A state write is an authorization decision, not the default endpoint of inference. Unconfirmed outputs should remain candidate states until they satisfy a task-dependent write policy or receive user confirmation. This is particularly important for identities, preferences, commitments, health-related observations, and descriptions of other people. The memory interface should allow the wearer to inspect why an item was stored, correct its content, narrow its scope, or reject the write. Personal-context learning and cross-view memory reasoning further motivate retaining the evidence and viewpoint from which a memory was constructed, rather than storing only a compressed textual conclusion~\cite{PersonalVisualContextLearning,EgoExoMem,H2HMem}.

\textbf{Temporal and spatial validity.}
Persistent state can become dangerous when it remains retrievable after the underlying world has changed. Episodic entries should record observation time and a validity or refresh policy; spatial entries should additionally encode map age, relocalization status, coordinate frame, object-persistence estimate, and the evidence for state changes. Aria Digital Twin and SpatialWorld provide relevant foundations for egocentric 3D perception and interactive spatial reasoning, while EgoExoMem illustrates that memory conclusions may depend on synchronized viewpoints~\cite{Ariadigitaltwin,SpatialWorld,EgoExoMem}. Corrections must propagate to summaries, indexes, derived preferences, and action plans so that a superseded belief is not reintroduced through another memory layer.

\textbf{Revocation, forgetting, and longitudinal evaluation.}
Users should be able to delete or suspend memory by item, person, place, time interval, or category, and sensitive contexts should support automatic non-recording or non-persistence policies. Expiration and forgetting are not merely storage optimizations: they are mechanisms for preventing stale assistance and reducing privacy exposure. Evaluation should report retrieval accuracy, temporal localization, answer-validity window, false-recall rate, correction persistence, stale-state use, cross-session relocalization, object-persistence accuracy, deletion effectiveness, and privacy leakage. Benchmark evidence from EgoMemReason, EGOSTREAM, and related memory resources should be complemented by longitudinal replay and field studies, because static question answering does not establish correction propagation or effective forgetting in a running system~\cite{EgoMemReason,EgoStream,MultimodalLMMs-EpisodicMemory}.

\subsection{Feedback and Intervention}
\label{sec:affective-interaction}

Feedback is the point at which internal system state becomes a user-facing intervention. Its utility depends not only on whether the content is correct, but also on whether the modality, timing, duration, referent, and level of initiative match the user's task and current environment. Poorly timed or poorly grounded assistance can impose cognitive burden, obscure environmental cues, or induce an incorrect action even when the underlying prediction is nominally accurate.

\textbf{Modality-task matching.}
Audio, captions, monocular HUD cues, binocular AR overlays, haptics through a companion device, and phone relay offer different combinations of privacy, salience, persistence, and visual or auditory load. Open-ear audio may preserve visual attention but is vulnerable to masking and leakage; display-based feedback can provide persistent text and spatial confirmation but may distract, occlude, drift, or become unreadable under bright illumination. Active noise-control research for open-ear glasses and studies of conversational successes and breakdowns show that output quality must be evaluated in the acoustic and social conditions in which it is used~\cite{ActiveNoiseCancellation,Conversational}. For referential tasks, the output should identify not only what to do but also which object, direction, or region the instruction concerns, with pointing and proximity benchmarks exposing common grounding failures~\cite{MLLMs-Pointing,EgoProx}.

\textbf{Intervention timing and proactive assistance.}
A proactive assistant must decide whether an intervention is necessary, when it remains useful, and whether the expected benefit exceeds interruption cost. Pro$^2$Assist, Plan-Watch-Recover, Ego-Pro-Bench, Streaming Interventions, and IPIBench move evaluation beyond retrospective recognition toward continuous step awareness, personalized initiative, online error correction, and interactive proactive intelligence~\cite{Pro2Assist,PlanWatchRecover,EgoPro-Bench,StreamingInterventions,IPIBench}. These works motivate separate measurements of missed interventions, premature interventions, late interventions, unnecessary interruptions, and successful recoveries. The optimal policy is risk-dependent: a low-risk reminder may tolerate delay, whereas a hazard warning or procedural correction may become invalid within seconds.

\textbf{Correctability and escalation.}
Every intervention should expose an appropriate correction path. Confirmation, clarification, repeat, dismiss, user override, undo, and controls for disabling proactive assistance are not interchangeable: their necessity depends on whether the output is informative, directive, or action-triggering. When evidence is ambiguous, the system should ask a focused clarification rather than convert uncertainty into an assertive instruction. When feedback is repeatedly ignored or a task becomes unsafe, escalation may involve switching modality, requesting explicit confirmation, deferring the task, or contacting an authorized human operator. Evaluation should include confirmation cost, override rate, time to first useful feedback, interruption cost, referential-hallucination rate, alert fatigue, and post-correction task success.

\textbf{Bounded use of inferred user state.}
Transient cues of workload, confusion, urgency, gaze allocation, or conversational breakdown may be used to adapt modality and timing, but they do not by themselves substantiate stable claims about personality, psychological condition, or health. Such adaptation should be purpose-limited, visible to the wearer, and accompanied by retention and deletion controls. A user should be able to inspect which cue caused an intervention policy to change and disable that inference channel. Child-facing, clinical, educational, and workplace deployments additionally may require guardian or organizational authorization and a narrower definition of permissible inference. The design objective is to support interaction timing, not to anthropomorphize the system or displace user judgment.

\subsection{Action and External-System Orchestration}
\label{sec:multi-agent}

When intelligent glasses move from describing or recommending to executing, the system transfers not only data but also authority and responsibility. Orchestration may involve a companion phone, watch, earphones, IoT devices, web services, enterprise software, remote experts, or robots. The design problem is therefore broader than cooperation among models: it concerns stateful handoff, authorization, execution provenance, consequence-aware control, and recovery across heterogeneous systems.

\textbf{Stateful handoff across devices and services.}
Each handoff should record the supporting observation, relevant memory and spatial-state version, requested operation, executing entity, authorization basis, credentials or permission scope, returned result, and recovery status. Intention-aware communication for AI glasses highlights the importance of transmitting task-relevant intent rather than indiscriminately forwarding raw streams~\cite{IntentionAwareSemanticAgentCommunications}. In practice, the glasses may provide first-person evidence and immediate feedback, a phone may host application permissions, a cloud model may plan, an enterprise server may enforce organizational policy, and an external device may execute. Unattributed outputs should not silently enter memory, and a result from an external node should be reconciled with the state visible to the wearer before further action.

\textbf{Consequence-tiered authority.}
Action authority should be stratified by consequence rather than by model confidence alone. A useful hierarchy distinguishes \emph{read-only observation}, \emph{advisory output}, \emph{reversible digital operations}, \emph{operations with external side effects}, \emph{physical-world execution}, and \emph{high-risk or regulated actions}. Higher tiers require progressively stronger evidence, explicit confirmation, least-privilege access, separation of duties, undo or rollback, audit logging, and human escalation. The same model and task planner therefore support different capability claims in read-only and execution modes. A high task-success rate is insufficient when rare failures create irreversible financial, physical, legal, or clinical consequences.

\textbf{Grounded web, mobile, and enterprise action.}
Ego2Web and Egocentric Co-Pilot directly connect first-person video to web-native assistance, while Mobile-Agent, WebArena, and SeeAct provide broader methodological foundations for visual mobile and web planning~\cite{Ego2Web,EgocentricCoPilot,MobileAgent,Webarena,Seeact}. These systems are important evidence for planning and interface grounding, but they do not by themselves validate smart-glasses deployment. Transfer requires first-person privacy filtering, device-level confirmation, robust handoff across screens and services, account and regional availability, and recovery from partial execution. Evaluation should therefore distinguish plan correctness, grounding accuracy, permission violations, execution failure, unsafe action, rollback success, and action-provenance completeness.

\textbf{Embodied-data transfer and downstream robot validation.}
Smart glasses can also serve as an interface for collecting human demonstrations, aligning ego-exo observations, learning active perception, or synthesizing robot training data. Ego-Exo4D, ActiveGlasses, EgoMI, EgoMimic, EgoZero, EgoVLA, and Ego2Robot illustrate complementary routes from egocentric human activity to embodied representation, active vision, imitation learning, VLA training, and robot-data synthesis~\cite{Ego-Exo4D,Activeglasses,EgoMi,Egomimic,Egozero,EgoVLA,Ego2Robot}. This route introduces additional requirements for camera-body calibration, time synchronization, hand and object state estimation, action retargeting, consent, de-identification, and embodiment-gap analysis. Human-video scale or alignment quality cannot be treated as evidence of robot competence on its own; claims must ultimately be validated through downstream robot success, sample efficiency, cross-embodiment generalization, intervention rate, and safety failures.

\subsection{Reliability and Recovery}
\label{sec:trust}

Because intelligent glasses operate as a composition of sensing, inference, memory, feedback, tools, and external execution, an upstream error can be amplified by later modules. Reliability should therefore be designed around failure attribution and recovery trajectories rather than a single confidence score or aggregate accuracy measure.

\textbf{Failure-source decomposition.}
At minimum, the system should distinguish \emph{observation failure} (the event was not captured or the target was outside the field of view), \emph{perception or reasoning failure}, \emph{reference ambiguity}, \emph{stale or inconsistent state}, \emph{network or tool unavailability}, and \emph{execution failure}. These sources require different responses: reacquisition for a missed observation, clarification or abstention for ambiguous evidence, state refresh or relocalization for stale context, fallback or deferred execution for unavailable services, and rollback or execution recovery after a failed action. A generic ``low confidence'' warning hides the causal source and prevents both the user and the system from selecting an appropriate response.

\textbf{Barriers before error propagation.}
Grounding verification, calibrated uncertainty, appropriate abstention, modality-conflict detection, state-validity checks, and authorization gates should be applied before an inference enters persistent memory or triggers action. Referential failures exposed by EgoPoint-Bench and temporally invalid recall exposed by EGOSTREAM show why fluent answers and high average accuracy do not guarantee safe downstream use~\cite{MLLMs-Pointing,EgoStream}. These barriers should be tested under low light, motion blur, acoustic interference, missing modalities, model updates, network degradation, corrupted state, and adversarial content, rather than only on clean benchmark samples.

\textbf{Recovery trajectory and versioned incident evidence.}
A defensible reliability claim should record the initiating failure, whether and when it was detected, the recovery policy selected, user or operator intervention, rollback outcome, and final post-recovery task result. Plan-Watch-Recover and Streaming Interventions provide relevant references for monitoring and correcting procedural execution as events unfold~\cite{PlanWatchRecover,StreamingInterventions}. Controlled fault injection, failure replay, model-update ablation, and incident review should complement standard task evaluation. Useful outcomes include failure-detection rate, calibration error, inappropriate-abstention rate, stale-state use, unsafe-action rate, rollback success, recovery success, time to recovery, and explanation usefulness. Versioned logs are essential because an unexplained model, firmware, or API update can otherwise make a previously observed failure impossible to reproduce.

\subsection{Privacy, Security, and Governance}
\label{sec:privacy-security}

First-person video, speech, speaker identity, gaze, location, household layout, screens, health-related cues, and longitudinal memory create a data surface that is broader and more persistent than that of a conventional handheld assistant. Once the device can invoke tools or control external systems, privacy and security failures can also produce downstream actions. Governance must therefore cover the complete sensing-state-action chain and all affected participants, not only the wearer who owns the device.

\textbf{Data minimization and lifecycle control.}
The system should collect and transmit only the spatial, temporal, and modal evidence needed for the current task. On-device filtering, face or screen redaction, speech segmentation, event-triggered capture, permission isolation, encrypted transmission, secure logging, short-lived credentials, and scoped retention reduce exposure before data reach a model or persistent store. VisGuardian provides a reference for lightweight privacy control over front-camera data in home environments, while work on life-logging emphasizes that privacy-utility trade-offs are task-dependent and cannot be eliminated by a single global policy~\cite{VisGuardian,Position}. Recording indicators, physical shutters or mute controls, private modes, place-based policies, memory dashboards, deletion, and data export make the device state and lifecycle observable and correctable.

\textbf{Bystander and multi-party governance.}
The wearer's consent does not automatically authorize the capture, inference, storage, or sharing of information about bystanders, household members, coworkers, patients, students, or remote service personnel. Prior HCI studies show that bystanders care about recording awareness, understandable indicators, and practical ways to mediate or opt out of camera-glasses capture~\cite{sgacad2014_bystanderprivacy,CameraGlassesPrivacy}. Mind the Gap further frames wearer-bystander tensions as context-dependent rather than solvable through a single indicator or gesture~\cite{MindTheGap}. Evaluation should therefore measure recording-state recognition, consent violations, bystander understanding, social discomfort, policy comprehension, and whether context-specific restrictions are actually enforced.

\textbf{Perceptual attacks and least-privilege execution.}
Environmental content can function as an instruction channel. Visual or audio prompt injection, malicious interface content, social engineering, tool-use abuse, and model-update drift can convert passive perception into unsafe reasoning or action. Visual adversarial jailbreaks demonstrate that image content can manipulate aligned multimodal models, PhySE studies real-time social-engineering risks in AR-LLM systems, and UNSEEN proposes cross-stack defensive mechanisms~\cite{VisualAdversarialExamplesJailbreak,PhySE,UNSEEN}. Defenses should combine input provenance, instruction-data separation, contextual policy checks, least-privilege tools, confirmation for consequential actions, anomaly detection, rollback, and incident review. Security claims must remain scoped to the evaluated threat model; success against one localized attack does not establish end-to-end security across device, model, memory, network, and tool layers.

\textbf{Organizational accountability and auditable policy.}
Enterprise, educational, clinical, and public-sector deployments require explicit rules for data ownership, retention, provider access, model and cloud responsibilities, cross-border processing, authorized locations, employee or patient rights, incident response, and independent audit. Technical controls should map to accountable entities and leave evidence of who accessed which state, under what policy, for what purpose, and with what result. Relevant outcomes include raw-data exposure, redaction recall, speaker or location leakage, consent-violation rate, least-privilege violations, attack success, unsafe-action rate, deletion effectiveness, audit completeness, and policy compliance. These outcomes should remain disaggregated because a single privacy or trust score can conceal materially different failure modes.

\subsection{Developer Interfaces and Reproducibility}
\label{sec:developer-ecosystem}

Developer access determines whether a system can be independently studied, extended, replayed, and audited. \textbf{\textit{1) Instrumentation and interfaces.}} A reproducible platform should expose sensor timestamps, calibration metadata, permission APIs, privacy controls, audio and display output, pose and map access where applicable, memory and tool interfaces, and structured execution logs. Project Aria and related research platforms illustrate the scientific value of synchronized sensing and calibration, while EgoKit and OpenGlass illustrate open capture and system-level prototyping routes~\cite{ProjectAria,EgoKit,OpenGlass}. \textbf{\textit{2) Versioning and replay.}} Device, firmware, model, API, prompt or policy, region, account tier, service availability, and subscription status should be pinned or recorded, because any of them may change observable behavior. Cross-device replay, deterministic test inputs, model-update regression tests, and failure reconstruction are necessary to attribute a result to hardware, perception, network, state, permission, or execution. \textbf{\textit{3) Openness is distinct from deployability.}} Raw-sensor access, calibration fidelity, synchronization, licensing, and evaluation scripts support research reproducibility, but do not establish consumer-grade comfort, endurance, safety, or field readiness. Conversely, a commercially available product may be wearable yet impossible to evaluate independently if raw data, version history, or logs are unavailable. Aria Gen~2, Pupil Labs Neon, and Tobii Pro Glasses~3 expose different combinations of sensing and developer access, and should therefore be compared according to the research question rather than through a single openness score~\cite{sgprod2026_meta_reality_labs_research_aria_gen_2,sgprod2026_pupil_labs_pupil_labs_neon,sgprod2026_tobii_tobii_pro_glasses_3}.

\begin{table}[t]
\centering
\caption{\textbf{Standardized evaluation protocol for smart glasses.}
The nine deployment-oriented design dimensions are evaluated under a shared claim and reproducibility context. Each result is conditioned on task scope, hardware route, L0-L5 claim, action-risk tier, participant structure, operating condition, evidence type, and confidence. The final column distinguishes benchmark evidence, controlled experiments, field studies, and validation that remains necessary for a deployment claim.}
\label{tab:evaluation-protocol}
\scriptsize
\begin{adjustbox}{width=\linewidth}
\begin{tabular}{
>{\raggedright\arraybackslash}m{27mm}
>{\raggedright\arraybackslash}m{56mm}
>{\raggedright\arraybackslash}m{66mm}
>{\raggedright\arraybackslash}m{56mm}
}
\toprule
\textbf{Dimension} & \textbf{Evaluation object and controlled variables} & \textbf{Primary observed outcomes} & \textbf{Evidence source / experiment} \\
\midrule

Claim and reproducibility context
& task scope; hardware route; L0-L5 claim; action-risk tier; wearer, bystander, operator, and organizational roles; device/firmware/model/API version; region/date; service and subscription status; sensor/output configuration
& claim applicability; missing-evidence rate; evidence confidence; confidence interval; version drift; independent-verification status; conflict-of-interest disclosure
& protocol record; versioned evidence profile; public artifact and audit trail \\

\midrule
Hardware profile and form factor
& camera, microphone, IMU, gaze/depth/pose where available; speaker/display; calibration; battery; thermals; weight distribution; FOV; fit; recording controls
& SNR and WER under noise; display brightness and PPD/FOV; audio leakage; POV alignment; timestamp/calibration error; battery drain; thermal throttling; comfort degradation; long-term adherence
& official specifications; laboratory measurement; independent teardown; calibration replay; field diary; longitudinal wear study \\

\midrule
Runtime and resource management
& capture-to-feedback/action pathway; compute and state placement; fallback policy; network condition; sampling duty cycle; context budget; thermal and resource scheduling
& p50/p95/p99 latency; timeout and jitter; energy per successful task; offline degradation; network sensitivity; thermal derating; on-device processing ratio; validity-window success
& controlled stress test; network ablation; energy profiling; degraded-mode replay; field trace \\

\midrule
Perception and inference stack
& OCR; object/action and audio-event recognition; ASR; SLAM/VIO; gaze/hand tracking; VQA; translation; spatial and temporal reasoning; instruction following
& task accuracy; mAP; WER; tracking accuracy; ATE/RPE; gaze error; timestamp skew; motion/low-light robustness; grounding and faithfulness; source attribution; conflict resolution; appropriate abstention; temporal consistency
& SuperGlasses~\cite{Superglasses}; SAW-Bench~\cite{SAW-Bench}; WearVQA~\cite{WearVQA}; EgoSAT~\cite{EgoSAT}; EgoPoint-Bench~\cite{MLLMs-Pointing}; device-stream replay \\

\midrule
Persistent state and memory
& working, episodic, semantic, spatial, preference, and action memory; write/update/expiration/deletion policy; provenance; confirmation; map and object validity
& retrieval accuracy; temporal localization; answer-validity window; false recall; map staleness; object-persistence accuracy; cross-session relocalization; correction propagation; deletion effectiveness; privacy leakage
& EgoLife~\cite{Egolife}; EgoMemReason~\cite{EgoMemReason}; EGOSTREAM~\cite{EgoStream}; EgoExoMem~\cite{EgoExoMem}; Aria Digital Twin~\cite{Ariadigitaltwin}; SpatialWorld~\cite{SpatialWorld}; longitudinal replay and deletion audit \\

\midrule
Feedback and intervention
& wake-up behavior; captions; audio alerts; HUD/AR cues; pointing and referential feedback; confirmation and override; proactive and user-state-adaptive intervention
& time to first useful feedback; confirmation and interruption cost; missed/premature/late intervention; referential hallucination; wrong turn; cue drift; display distraction; audio masking; override rate; false user-state inference
& EgoPoint-Bench~\cite{MLLMs-Pointing}; EgoProx~\cite{EgoProx}; Conversational Breakdowns~\cite{Conversational}; Pro$^2$Assist~\cite{Pro2Assist}; Ego-Pro-Bench~\cite{EgoPro-Bench}; IPIBench~\cite{IPIBench}; route and user studies \\

\midrule
Action and external-system orchestration
& tool use; planning; permissions; multi-device handoff; web/mobile/enterprise execution; rollback; physical action; embodied-data alignment and transfer
& plan and task success; grounding error; permission violation; unsafe-action and execution-failure rate; intervention and rollback success; provenance completeness; alignment quality; usable data per hour; robot success; sample efficiency; cross-embodiment failure
& Ego2Web~\cite{Ego2Web}; Egocentric Co-Pilot~\cite{EgocentricCoPilot}; Ego-Exo4D~\cite{Ego-Exo4D}; ActiveGlasses~\cite{Activeglasses}; EgoMI~\cite{EgoMi}; EgoMimic~\cite{Egomimic}; EgoZero~\cite{Egozero}; downstream tool and robot experiments \\

\midrule
Reliability and recovery
& observation failure; reasoning and grounding error; stale state; network/tool unavailability; execution failure; model update; recovery and escalation policy
& failure-detection rate; calibration error; inappropriate abstention; stale-state use; unsafe-action rate; rollback and recovery success; time to recovery; post-recovery task outcome; explanation usefulness
& controlled fault injection; failure replay; model-update ablation; adversarial and degraded-condition tests; incident review \\

\midrule
Privacy, security, and governance
& raw-data exposure; redaction; recording state; consent; retention and deletion; prompt injection; least privilege; credentials; audit logs; organizational and place-based policy
& redaction recall; consent violation; speaker/location leakage; recording-state recognition; bystander understanding; attack success; privilege violation; unsafe action; deletion effectiveness; audit completeness; policy compliance; social discomfort
& CameraGlassesPrivacy~\cite{CameraGlassesPrivacy}; Mind the Gap~\cite{MindTheGap}; VisGuardian~\cite{VisGuardian}; visual jailbreaks~\cite{VisualAdversarialExamplesJailbreak}; PhySE~\cite{PhySE}; UNSEEN~\cite{UNSEEN}; privacy/security audits and field studies \\

\midrule
Developer interfaces and reproducibility
& SDK and raw-sensor access; timestamps; calibration metadata; pose/map APIs; permissions; memory/tool interfaces; logging schema; model/firmware versioning; licensing
& API coverage; timestamp integrity; calibration traceability; replay success; version drift; missing-data rate; failure reconstructability; third-party reproducibility; license clarity
& SDK audit; benchmark harness; cross-device replay; public scripts and configurations; independent replication \\

\bottomrule
\end{tabular}
\end{adjustbox}
\end{table}
\FloatBarrier

\subsection{Structured Evaluation Protocol}
\label{sec:standard-evaluation}

The purpose of standardized evaluation is not to produce a universal ranking of heterogeneous products, but to determine whether a particular capability claim is supported under its stated task, hardware, runtime, participant, and risk conditions. The protocol in \cref{tab:evaluation-protocol} therefore treats the nine design dimensions as coupled evaluation objects and augments them with a common claim and reproducibility context.

\textbf{Claim-conditioned unit of evaluation.}
The atomic unit is a capability claim conditioned jointly on task scope, hardware route, sensing and output configuration, L0-L5 level, action-risk tier, participant structure, operating environment, evidence type, and confidence. For example, ``L3 memory assistance for household object finding under a camera-and-audio profile'' is a different claim from ``L4 autonomous purchasing from the same observations,'' even when both use the same underlying model. N/A denotes unavailable evidence rather than zero capability, and a benchmark or prototype result should not be generalized to longitudinal field use without corresponding evidence. Comparisons should first be made among systems following similar hardware and interaction routes before cross-category differences are interpreted.

\textbf{Controlled variables and disaggregated outcomes.}
Each experimental record should fix the device, firmware, model or API version, region and date, service availability and subscription status, sensor and output configuration, and participant and scenario composition. Compute split, sampling duty cycle, context budget, network condition, battery state, temperature, illumination, acoustic noise, motion, task risk, and permission state can then be varied systematically. Outcomes should remain decomposed into task accuracy, grounding and faithfulness, p50/p95/p99 latency, energy per successful task, state freshness, correction and deletion behavior, intervention and rollback outcomes, bystander comprehension, security violations, and version drift. Benchmarks such as SuperGlasses, EgoSAT, EGOSTREAM, EgoPoint-Bench, EgoProx, and IPIBench provide localized evidence for different components of this profile, but none alone establishes end-to-end deployability~\cite{Superglasses,EgoSAT,EgoStream,MLLMs-Pointing,EgoProx,IPIBench}.

\begin{table}[t]
\centering
\caption{\textbf{Design checklist for deployable smart glasses.}
The nine rows correspond one-to-one with the design dimensions in Sections~\ref{sec:form-factor}-\ref{sec:developer-ecosystem}. Each item is interpreted with respect to task scope, hardware route, L0-L5 claim, action-risk tier, evidence source, confidence, and missing evidence. N/A denotes unavailable evidence rather than a default low score.}
\label{tab:design-checklist}
\scriptsize
\begin{adjustbox}{width=\linewidth}
\begin{tabular}{
>{\raggedright\arraybackslash}m{31mm}
>{\raggedright\arraybackslash}m{42mm}
>{\raggedright\arraybackslash}m{43mm}
>{\raggedright\arraybackslash}m{42mm}
>{\raggedright\arraybackslash}m{35mm}
}
\toprule
\textbf{Design dimension} & \textbf{Baseline evidence condition} & \textbf{Strengthened design} & \textbf{Unacceptable failure} & \textbf{Validation test} \\
\midrule

\ding{172} Hardware profile and form factor
& documented sensing/output configuration, calibration, weight and balance, battery life, thermals, brightness/FOV, audio leakage, fit, and recording controls
& category-normalized hardware profile, observable physical controls, point-of-view validation, and longitudinal comfort/adherence evidence
& marketing specifications or component count treated as evidence of model, closed-loop, or all-day capability
& laboratory measurement + calibration replay + longitudinal field diary \\

\midrule
\ding{173} Runtime and resource management
& documented end-to-end latency, energy use, compute/state placement, network state, fallback behavior, and runtime failure rate
& tail-latency control, state consistency, explicit degraded modes, safety-critical local fallback, and thermal/resource-aware scheduling
& high-risk feedback or action delivered after its supporting evidence has become invalid, without visible degradation or escalation
& latency/energy stress test + network and node-failure ablation \\

\midrule
\ding{174} Perception and inference stack
& fixed input configuration and model/API version, with observable grounding, uncertainty, and failure signaling
& layered local/cloud inference, selective context escalation, modality-conflict detection, source attribution, and appropriate abstention
& unprovenanced, temporally stale, or ambiguously grounded output entering persistent state or triggering action
& benchmark evaluation + device-stream replay + missing-modality test \\

\midrule
\ding{175} Persistent state and memory
& user access to stored state, provenance, confirmation status, correction, retention, and deletion mechanisms
& typed memory, write admission, map-age and validity tracking, automatic expiration, scoped policies, and correction propagation
& model inference stored as a confirmed fact, stale state reused as current evidence, or deletion/correction failing to propagate
& memory QA + longitudinal/spatial replay + correction and deletion audit \\

\midrule
\ding{176} Feedback and intervention
& confirmation, clarification, dismiss, override, and controls for disabling proactive intervention
& risk-aware modality/timing selection, referential clarification, interruption-cost modeling, and bounded user-state adaptation
& uncorrectable instruction, persistent intervention outside its validity window, or stable psychological/health labeling from isolated cues
& route/procedural user study + interruption, referential, and inferred-state audit \\

\midrule
\ding{177} Action and external-system orchestration
& explicit authorization for external actions; documented handoff, permission, returned state, and recovery; synchronization/calibration and consent scope for embodied data
& consequence-tiered authority, least privilege, separation of duties, undo/rollback, end-to-end provenance, de-identification, and auditable action logs
& irreversible autonomous action without appropriate authorization, or robot-capability claims inferred from human data without downstream robot validation
& situated tool task + permission/unsafe-action test + embodied-data audit + robot experiment \\

\midrule
\ding{178} Reliability and recovery
& failures distinguished among missed observation, reasoning/grounding error, ambiguous reference, stale state, tool unavailability, and execution failure
& pre-action verification, calibrated abstention, failure-specific recovery, user escalation, versioned replay, and incident review
& generic confidence signaling that obscures failure source, or failed recovery without rollback, escalation, or an auditable record
& controlled fault injection + degraded-condition replay + end-to-end recovery evaluation \\

\midrule
\ding{179} Privacy, security, and governance
& observable recording controls, data minimization, consent and retention management, deletion pathways, scoped credentials, and audit logging
& on-device redaction, context/place policies, instruction-data separation, prompt-injection defense, least privilege, rollback, and organizational incident procedures
& covert or incomprehensible recording, unauthorized tool execution, unrevocable sensitive state, or policy that cannot be technically audited
& privacy and deletion audit + prompt-injection/privilege test + bystander and organizational study \\

\midrule
\ding{180} Developer interfaces and reproducibility
& traceable API, firmware, model, region, and service versions, together with timestamps, calibration metadata, permissions, and execution logs
& pose/map and memory/tool APIs, version pinning, evaluation harnesses, cross-device replay, public configurations, and reproducibility scripts
& an independent third party cannot identify the evaluated system version, reconstruct a failure, or reproduce the reported result under the stated conditions
& SDK and license audit + regression replay + independent replication \\

\bottomrule
\end{tabular}
\end{adjustbox}
\end{table}
\FloatBarrier

\textbf{Iterative evidence ladder.}
Evaluation is a claim-specific and iterative evidence chain rather than a linear certification sequence that every product must pass in the same manner. Official documentation establishes nominal availability; laboratory measurement establishes physical and runtime boundaries; fixed benchmarks assess localized capabilities; device-stream replay and controlled fault injection examine cross-module composition; end-to-end task studies measure closed-loop behavior; longitudinal field studies expose adherence and context drift; and privacy, security, and governance audits evaluate data and authority boundaries. Any failure occurring at any stage shall feed back into its corresponding design dimension, including sensing, sampling, model, state policy, feedback, authorization, runtime, and interface. The revised system must then undergo re-evaluation against a newly versioned operational profile.

\textbf{Reporting, comparison, and checklist rules.}
Every result should report missing data, confidence intervals where applicable, evidence provenance, independent-verification status, and conflicts of interest. Evidence confidence should reflect both source type and experimental control: a vendor statement, a third-party laboratory measurement, a public benchmark, and a longitudinal deployment provide different forms of support. The protocol table specifies what to control and observe, whereas the design checklist in \cref{tab:design-checklist} maps the same dimensions to baseline evidence conditions, strengthened safeguards, unacceptable failures, and validation tests. Neither table defines a task-agnostic passing threshold, and the presence of a recommended test should not be interpreted as evidence that a product has already passed it.

%% file: sec/06_conclusion.tex
\section{Conclusion and Future Prospects}
\label{sec:conclusion}

\subsection{Challenges and Future Roadmap}
\label{sec:risks-future-roadmap}

Despite rapid advances in sensing hardware, egocentric multimodal models, proactive agents, and near-eye interaction, smart glasses remain far from becoming reliable general-purpose platforms for embodied intelligence. The central difficulty is no longer an isolated deficiency in recognition, generation, or interaction accuracy, but the need to sustain an end-to-end loop that continuously observes the world, maintains temporally valid state, decides whether and how to intervene, communicates with the wearer, invokes external tools or embodied systems, and recovers from failure under strict resource, privacy, and social constraints. These requirements are tightly coupled: improving one component may increase the cost or risk of another, while local errors can propagate across perception, memory, interaction, and action. We organize the remaining barriers into eight challenges, following the path through which failures enter and propagate across the smart-glasses system.

\sgpoint{Hardware budgets and sustained closed-loop operation}{
Weight, battery capacity, thermal dissipation, optical efficiency, camera placement, microphone geometry, sensor synchronization, and wireless connectivity jointly determine whether a smart-glasses system can remain useful beyond a short demonstration. These constraints propagate throughout the closed loop: thermal throttling may reduce sensing or inference frequency, unstable connectivity may make retrieved context stale, poor camera placement may degrade hand-object visibility, and wind or environmental noise may undermine speech interaction. System-oriented efforts on efficient egocentric perception, embedded visual-inertial estimation, sensing-computing separation, and open-ear audio illustrate the importance of treating hardware and inference as a co-designed stack rather than independent modules \cite{EPIC,LEVIO,OpenGlass,ActiveNoiseCancellation}. Future research should therefore report not only average model accuracy or latency, but also tail latency, energy per useful intervention, thermal recovery, frame and sensor dropout, network-degradation behavior, and performance over hours of continuous wear. Long-horizon stress tests should further quantify how resource adaptation changes downstream grounding, memory consistency, interaction quality, and task success, rather than assuming that offline capability remains unchanged after deployment on a wearable device.
}

\sgpoint{Longitudinal data and annotation bottlenecks}{
Smart glasses are expected to reason over activities, places, objects, people, and routines that unfold across days or months, yet most data pipelines are still optimized for bounded recording sessions and relatively static annotation targets. Large-scale egocentric datasets and emerging always-on or month-level benchmarks have substantially broadened the temporal and environmental scope of first-person data \cite{Ego4D,Ego-Exo4D,AoE,Open-AoE,EgoMonth}, but sustained collection remains constrained by bystander privacy, sensitive locations, sparse critical events, annotation cost, device-specific fields of view, calibration drift, and changes in firmware or hardware. These factors directly affect cross-session relocalization, personal memory, preference learning, event retrieval, and the validity of long-term world state. Future datasets should consequently record not only audiovisual observations, but also calibration and device metadata, sensor quality, consent state, user corrections, uncertainty, state expiration, and deletion requests. Annotation should move from isolated frame labels toward temporally persistent entities, events, relationships, and state transitions, supported by automated filtering, active learning, multimodal synchronization, and human verification. Without such lifecycle-aware data, improvements in long-context modeling may still produce brittle or unverifiable long-term assistance.
}

\sgpoint{Grounding, memory, and model uncertainty}{
First-person intelligence requires more than recognizing visible objects or answering questions about the current frame. A system must determine what a user is referring to, whether an observation is sufficiently reliable, how it relates to previous events and spatial state, and whether the resulting inference should be stored or acted upon. Motion blur, occlusion, speech-recognition errors, ambiguous pointing, viewpoint changes, stale maps, and changing object or tool states can compound across the transitions from observation to inference, inference to memory, and memory to action. Benchmarks on egocentric referential reasoning, interactive spatial reasoning, proximity understanding, and long-horizon episodic memory expose different parts of this problem \cite{MLLMs-Pointing,SpatialWorld,EgoProx,EgoMemReason,EgoStream}. Future systems should jointly model source attribution, temporal validity, confidence calibration, memory-admission criteria, contradiction detection, abstention, and correction propagation. Evaluation should determine whether uncertainty is recognized and communicated before unsupported state is persisted, whether later evidence can revise earlier conclusions, and whether a local grounding failure remains contained rather than becoming a false memory, incorrect reminder, misleading instruction, or unsafe external action.
}

\sgpoint{Personalization, accessibility, and population diversity}{
The utility of smart glasses is inherently user-dependent. Differences in visual and auditory ability, speech patterns, language, head motion, gait, hand preference, cultural norms, technical experience, cognitive workload, and tolerance for interruption can substantially alter both sensing quality and the appropriateness of system behavior. Research on human-centered wearable assistance, visual assistance for blind or low-vision users, communication access for deaf users, and inclusive mixed-vision interaction demonstrates that a single interaction and assistance policy is unlikely to serve all populations equally \cite{HCDesignAndFabrication,EgoBlind,EvaluatingARForDeafStudents,ReshapingInclusiveInterpersonalDynamics}. At the same time, personalization itself introduces sparse-data, privacy, adaptation, and evaluation problems: a system may overfit to recent behavior, infer sensitive traits, fail after a change in routine, or improve aggregate performance while worsening outcomes for particular user groups. Future work should distinguish adaptation to stable preferences from adaptation to temporary context, support explicit user correction and preference control, and report group-stratified performance rather than only population averages. Personalization should be evaluated not merely by predictive accuracy, but by whether it improves accessibility, reduces interaction burden, preserves user agency, and remains robust across changing physical, social, and environmental conditions.
}

\sgpoint{Interaction timing, proactivity, and correctability}{
The value of proactive assistance depends not only on what a system predicts, but also on whether it intervenes at an appropriate moment, through an appropriate modality, and with a suitable degree of confidence and reversibility. Everyday smart-glasses studies and emerging benchmarks for continuous procedural assistance, mistake correction, and interactive proactive intelligence increasingly shift attention from passive response generation to intervention policy \cite{Conversational,Pro2Assist,PlanWatchRecover,StreamingInterventions,IPIBench}. However, a technically correct message may still be harmful when it arrives too late, interrupts a safety-critical action, obscures the visual field, repeats information the wearer already knows, or demands confirmation during high workload. Conversely, excessive conservatism can eliminate the practical value of proactivity. Future work should jointly optimize intervention timing, modality, duration, specificity, confirmation threshold, override, and rollback while accounting for task risk and individual preference. Evaluation should include missed-intervention cost, false-alarm burden, attention disruption, recovery time, repeated-error rate, user override behavior, and failure severity. Correctability must be treated as a first-class system property, ensuring that users can inspect, reject, revise, postpone, or reverse both system conclusions and consequential actions.
}

\sgpoint{Ecosystem interoperability and action reliability}{
Smart glasses rarely operate as self-contained devices. Their practical capabilities emerge from a distributed ecosystem involving on-device processors, phones, edge or cloud models, retrieval services, personal data stores, web agents, enterprise applications, and, increasingly, robots or other embodied platforms. Smart-glasses and mobile-agent systems have begun to explore this transition from situated perception to web- and tool-mediated action \cite{VisionClaw,Ego2Web,EgocentricCoPilot,MobileAgent,Seeact}. Yet distribution introduces failure modes that are not captured by conventional perception benchmarks, including state divergence across devices, stale capability descriptions, incompatible coordinate or identity representations, partial tool completion, duplicated side effects, and actions executed after the originating context is no longer valid. Future systems need explicit capability discovery, typed action and state schemas, identity and permission propagation, idempotent execution, transactional confirmation, timeout handling, and compensating rollback. The reliability of an action should be evaluated from the initial user intent through grounding, tool selection, execution, verification, and recovery, rather than inferred from the accuracy of an intermediate plan. Interoperability standards must also preserve provenance so that users can determine which device, model, service, or external agent produced each consequential outcome.
}

\sgpoint{Privacy, security, and social governance}{
Continuous body-proximate sensing places wearers, bystanders, families, coworkers, service providers, employers, and institutions within a shared sensing, storage, inference, and accountability ecosystem. The resulting risks extend beyond conventional data leakage. Life-logging can expose identity, location, relationships, routines, screens, conversations, and sensitive spaces, while visual prompt injection, adversarial content, unauthorized memory writes, and deceptive augmented-reality cues may manipulate downstream inference or action \cite{Position,MindTheGap,VisGuardian,PhySE,UNSEEN}. Earlier studies of bystander attitudes and privacy-mediating gestures further show that visible recording indicators alone do not resolve multi-party expectations and social acceptability \cite{sgacad2014_bystanderprivacy,CameraGlassesPrivacy}. Future deployment therefore requires layered protections combining data minimization, local processing, encryption, access isolation, memory provenance, content-origin detection, permission-aware tool invocation, and auditable action logs. These mechanisms must be complemented by understandable recording states, context-sensitive consent and revocation, organizational authorization, location-specific policy, retention limits, cross-border data handling, and incident accountability. Privacy and security should be evaluated under realistic adversarial and social conditions, including whether deletion propagates across derived memories and models, whether bystanders can exercise meaningful control, and whether failures can be attributed and remediated.
}

\sgpoint{Evaluation, version drift, and reproducibility}{
Existing benchmarks have begun to cover wearable question answering, streaming interaction understanding, situated awareness, and smart-glasses agent behavior \cite{WearVQA,Superglasses,EgoSAT,SAW-Bench}. Nevertheless, benchmark scores are frequently overgeneralized into claims about complete products or deployment readiness, even though real systems vary in sensor configuration, device placement, firmware, model version, API availability, regional support, subscription features, and network conditions. Studies of wearable OCR, for example, illustrate that physical factors such as motion and camera placement can alter downstream performance in ways that conventional static evaluation may not capture \cite{EvaluatingOCRPerformance}. Future evaluation should therefore express capability claims atomically and associate them with a versioned evidence profile specifying the device, sensors, model, software stack, operating condition, task scope, temporal horizon, action authority, and risk level. Standard protocols should combine controlled replay with real-device field testing, include latency and energy distributions rather than averages alone, and report recovery, uncertainty, privacy, and failure severity alongside task success. The objective should not be a timeless leaderboard, but a traceable record of which claims remain reproducible under which system versions and deployment conditions.
}

These challenges are analytically distinct but operationally inseparable. Increasing sensing frequency may strengthen temporal grounding while shortening battery life, increasing thermal load, and expanding privacy exposure. Larger models may improve complex reasoning while increasing response latency and dependence on remote computation. More persistent memory may improve continuity while amplifying the consequences of an incorrect or unauthorized state update. Similarly, broader tool access can increase task utility while enlarging the attack surface and raising the burden of authorization, confirmation, and rollback. Smart-glasses research should therefore move from optimizing isolated components toward co-designing the full perception-state-interaction-action loop. Against this background, \textbf{\textit{we outline six complementary research directions as follows}}:

\begin{itemize}
    \item \textbf{Reproducible device and system profiles.} A first priority is to transform one-off functional demonstrations into reproducible, versioned, and inspectable system artifacts. A unified profile should document sensor placement and sampling, calibration, device form factor, firmware and model versions, compute allocation, offloading strategy, network assumptions, display and audio configuration, privacy filters, action interfaces, and recovery mechanisms. Research platforms and system frameworks for multimodal egocentric sensing, heterogeneous-device collection, efficient wearable perception, embedded odometry, and sensing-computing separation provide useful foundations for such profiles \cite{ProjectAria,EgoKit,EPIC,LEVIO,OpenGlass}. Each profile should be accompanied by long-duration resource traces, representative failure cases, correction and rollback logs, and controlled degradation experiments across battery, thermal, illumination, motion, and connectivity conditions. Open replay interfaces would allow new models to be evaluated against the same sensor streams and action histories, while reference hardware tiers could prevent conclusions from silently depending on unavailable compute or sensors. The goal is not to prescribe a single product form, but to establish comparable experimental objects and clearly delimit the conditions under which a system-level capability claim remains valid.
    
    \item \textbf{Privacy-aware longitudinal data engines.} Future progress will require data infrastructure that treats collection, curation, annotation, consent, correction, and deletion as a continuous lifecycle rather than a one-time dataset release. Existing large-scale, multi-view, always-on, open-toolchain, and month-level egocentric efforts indicate a progression toward longer and more diverse first-person observations \cite{Ego4D,Ego-Exo4D,AoE,Open-AoE,EgoMonth}. The next generation of data engines should support heterogeneous glasses and companion devices, synchronized audiovisual and inertial streams, calibration and quality metadata, persistent entity and event representations, user corrections, and machine-readable consent states. Automated agents can assist with quality filtering, privacy redaction, temporal segmentation, cross-view alignment, uncertainty estimation, and candidate annotation, while humans verify sensitive or high-impact labels. Such infrastructure should also support permission-aware subsets, federated or on-device learning, auditable data lineage, and verifiable removal of raw observations and derived annotations. Rather than maximizing recorded hours alone, dataset design should measure coverage of activities, environments, users, rare failures, long-term state changes, and downstream capability gain. This would make longitudinal data a governed research substrate for continual assistance rather than an opaque archive of personal experience. 
    
    \item \textbf{Auditable memory and continual world models.} Persistent assistance requires memory systems that distinguish direct observations from model-generated interpretations, user-confirmed facts, inferred preferences, spatial state, and action history. Emerging benchmarks for long-horizon, streaming, cross-view, and everyday egocentric memory provide important components for studying this problem \cite{EgoMemReason,EgoStream,EgoExoMem,LightMem-Ego}, while spatial-memory and egocentric world-model research points toward representations that support prediction and interaction rather than retrieval alone \cite{LatentSpatialMemory,EgoForge}. Future systems should maintain explicit provenance graphs linking each stored state to its supporting observations, confidence, timestamp, access policy, and subsequent revisions. Memory admission, consolidation, contradiction resolution, expiration, forgetting, and deletion should be jointly optimized, with high-impact states requiring stronger evidence or user confirmation. Cross-device portability should preserve both state and provenance, while population-level learning should separate transferable regularities from personally identifying information. Evaluation should cover false-memory creation, stale-state detection, correction propagation, deletion effectiveness, cross-session consistency, and the consequences of memory errors on subsequent interaction or action. The intended outcome is not unlimited recall, but a correctable and accountable world model whose continuity remains useful without becoming opaque or irreversible.
    
    \item \textbf{Adaptive interaction and inclusive proactivity.} Future smart glasses should learn not only what assistance to provide, but also when to ask, wait, warn, summarize, display, or act. Research on proactive procedural assistance, recovery, personalized interventions, gaze-supported reference resolution, and alternative wearable input illustrates the range of signals that can inform such policies \cite{Pro2Assist,PlanWatchRecover,EgoPro-Bench,GazePointAR,FingerGlass}. A unified interaction controller should reason over task phase, predicted risk, wearer attention, environmental noise, motion, display availability, confidence, and learned preference before selecting a feedback channel or intervention level. It should support graded behavior ranging from silent state maintenance, through subtle audio or visual cues, to explicit confirmation and emergency interruption. Personalization must remain inspectable: users should be able to specify interruption preferences, accessibility needs, sensitive situations, and actions that always require confirmation. Longitudinal field studies should measure not only task completion, but also cognitive load, trust calibration, social acceptability, correction frequency, accessibility benefit, and adaptation stability. By jointly optimizing utility, timing, modality, and reversibility, proactive intelligence can become a negotiable form of assistance rather than an always-active source of distraction or automation.
    
    \item \textbf{Interoperable display, spatial, and action ecosystems.} A mature smart-glasses platform should coordinate near-eye display, audio, gaze and hand input, persistent spatial state, personal memory, phones, cloud services, web tools, and embodied agents through explicit and inspectable interfaces. Current smart-glasses and egocentric-agent systems demonstrate the potential to connect first-person observations with external digital environments \cite{VisionClaw,Ego2Web,EgocentricCoPilot,IntentionAwareSemanticAgentCommunications}, but reliable coordination requires more than adding tool-calling capability to a multimodal model. Future architectures should introduce shared schemas for referents, locations, identities, temporal validity, permissions, and action status; capability negotiation across devices and services; and transactional execution that separates proposal, confirmation, execution, verification, and rollback. The display should function not only as an output channel, but also as a confirmation and provenance surface showing what the system believes, which evidence supports it, and which service will act. Spatial-state updates and external tool results should be reconciled before subsequent actions are issued. Open, policy-aware APIs and standardized failure traces would allow systems to be compared across ecosystems while reducing dependence on a single vendor, model, region, or service configuration.
    
    \item \textbf{Robot-validated transfer of human experience.} Smart glasses offer a uniquely scalable interface for capturing human demonstrations because they naturally observe hands, objects, gaze, language, movement, and task context from the actor's perspective. Recent work on imitation learning from egocentric video, robot learning from smart glasses, egocentric VLA training, embodiment alignment, and robot-data synthesis reflects growing interest in exploiting this source of experience \cite{Egomimic,Egozero,EgoVLA,EgoEngine,Ego2Robot,ACE-Ego-0}. The central research problem is to bridge differences between human and robot morphology, sensing, viewpoint, actuation, contact, and safety constraints. Future pipelines should transform glasses-captured streams into synchronized and quality-controlled representations of objects, hands, actions, contact, gaze, language, spatial state, and task outcomes, followed by embodiment-aware retargeting and robot-side adaptation. Active-vision and whole-body learning further suggest that head and body motion should be modeled as part of the demonstration rather than treated as nuisance variation \cite{Activeglasses,EgoMi}. Most importantly, claims of embodied transfer should be validated on the target robot through task success, sample efficiency, generalization, failure recovery, unsafe-execution rate, and comparison with robot-native data. Human-video alignment alone should not be interpreted as robot capability until the benefit has been demonstrated through downstream physical execution.
\end{itemize}

\subsection{Conclusion}
\label{sec:final-conclusion}

This survey presents a system-level framework for understanding smart glasses as first-person intelligence platforms, linking data-flow formulation, device profiles, foundational capabilities, application scenarios, deployment design, and claim-conditioned evaluation through a common perception-state-interaction-action loop. Across these layers, a consistent picture emerges: the value of smart glasses depends on their ability to maintain useful connections between ongoing first-person observations, spatial and temporal context, user intent, and downstream digital or physical actions. This shifts the research focus from isolated model performance toward the behavior of the complete system under wearable operating conditions. Reliable deployment requires sensing and inference that respect device resource limits, state and memory whose provenance and validity can be inspected and revised, feedback that is timely and appropriate to the user's activity, and external actions that are explicitly authorized, monitored, and recoverable. These requirements also make evaluation inherently conditional on the task, hardware and runtime configuration, system version, operating environment, affected stakeholders, and consequences of failure. Progress will therefore require closer integration of hardware-software co-design, longitudinal first-person data, persistent spatial and personal state, inclusive interaction, interoperable agent ecosystems, privacy and security mechanisms, and downstream validation for embodied transfer. The long-term promise of smart glasses will ultimately depend less on any individual sensor, display, model, or agent than on whether these components can work together to maintain a trustworthy, correctable, and auditable connection between first-person experience and real-world assistance or action.